%% file: jair.tex
\documentclass[twoside,11pt]{article}
\usepackage{jair, theapa, rawfonts}

\usepackage{url}
\usepackage{times}
\usepackage{latexsym}

\usepackage{graphicx}
\usepackage{wrapfig}
\usepackage{booktabs}
\usepackage{todonotes}
\usepackage{enumitem}
\usepackage{amssymb,amsmath}
\usepackage{comment}
\usepackage{subcaption}
\usepackage{framed}
\usepackage{textcomp}

\usepackage{array}
\newcolumntype{H}{>{\setbox0=\hbox\bgroup}c<{\egroup}@{}}

\usepackage{siunitx}

\usepackage{color,soul}

\usepackage{tikz}
\usetikzlibrary{bayesnet}

\input{math_commands.tex}

\ShortHeadings{Mediation Analysis for Interpreting NLP: Gender Bias}
{Vig, Gehrmann, Belinkov, Qian, Nevo, Sakenis, Huang, Singer, \& Shieber}
\firstpageno{1}

\begin{document}

\title{Causal Mediation Analysis for Interpreting Neural NLP:\\ The Case of Gender Bias}

\author{\name Jesse Vig\thanks{Equal contribution. Work conducted while J.V. was at Palo Alto Research Center.}  \email jvig@salesforce.com \\
       \addr Salesforce Research
       \AND
       \name Sebastian Gehrmann\textsuperscript{*} \email gehrmann@seas.harvard.edu \\ 
       \addr Harvard University 
       \AND 
       \name Yonatan Belinkov\textsuperscript{*} \email belinkov@technion.ac.il \\
       \addr Technion -- Israel Institute of Technology 
       \AND 
       \name Sharon Qian \email sharonqian@seas.harvard.edu \\ 
       \addr Harvard University
       \AND
       \name Daniel Nevo \email danielnevo@tauex.tau.ac.il \\
       \addr Tel Aviv University 
       \AND
       \name Simas Sakenis \email simassakenis@college.harvard.edu \\
       \name Jason Huang \email jasonhuang@college.harvard.edu \\ 
       \name Yaron Singer \email yaron@seas.harvard.edu \\
       \name Stuart Shieber \email shieber@seas.havard.edu \\
       \addr Harvard University
       }

\maketitle

\begin{abstract}
    Common methods for interpreting neural
    models in natural language processing typically examine either their structure or their predictions, but not both.  
    We propose a methodology grounded in the theory of causal mediation analysis for interpreting which parts of a model are causally implicated in its behavior. It enables us to analyze the mechanisms by which information flows from input to output through various model components, known as mediators. 
    We apply this methodology in a case study of gender bias in pre-trained Transformer language models. 
    We analyze the role of individual neurons and attention heads in mediating gender bias across three datasets designed to gauge a model's sensitivity to grammatical gender. 
    Our mediation analysis reveals that gender bias effects are (i) sparse,  concentrated in a small part of the network; 
    (ii) synergistic, amplified or repressed by  different components; and (iii) decomposable into effects flowing directly from the input and indirectly through the mediators.
\end{abstract}

\section{Introduction}

The success of neural network models in various natural language processing tasks, coupled with their opaque nature, has led to much interest in interpreting and analyzing such models.
Analysis methods may be categorized into structural and behavioral analyses~\cite{tenney-etal-2019-bert}.
\emph{Structural} analyses aim to shed light on the internal structure of a neural model, for example through probing classifiers~\cite{conneau-etal-2018-cram,hupkes2018visualisation,adi2017fine} that predict linguistic properties using representations from trained models. 
This methodology has been used for analyzing sentence embeddings, machine translation models, and
contextual word representation models, among other models~\cite{belinkov-glass-2019-analysis}. 
\emph{Behavioral} analyses, on the other hand, aim to assess a model's behavior by its performance on constructed examples~\cite<e.g.,>[]{isabelle-etal-2017-challenge,naik-etal-2018-stress}, or by visualizing important input features via saliency methods~\cite<e.g.,>[]{li-etal-2016-visualizing,james2018beyond}.

\begin{figure}[t]
    \centering
    \includegraphics[trim={0 0 0 0},clip,width=.85\linewidth]{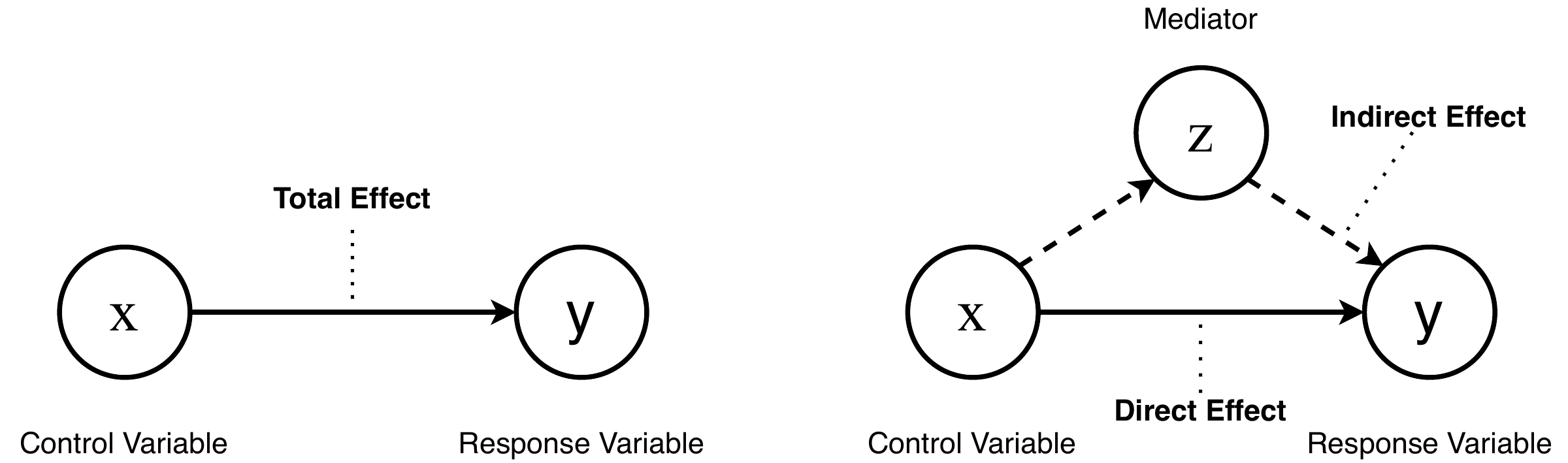}
    \caption{Causal mediation analysis introduces a mediator into the causal graph shown on the left. The mediator decouples the total effect into a direct and indirect effect.} 
    \label{fig:intro}
\end{figure}

Despite yielding interesting and useful insights, both types of analyses suffer from significant limitations. 
As pointed out by \citeA{belinkov-glass-2019-analysis}, probing classifiers only yield a correlational measure between a model's representations and an external linguistic property, and are thus not \emph{causally} connected to the model's predictions. \citeA{barrett2019adversarial} further demonstrate that classifiers that aim to detect biases in learned representations focus on spurious correlations in their training data and fail to generalize to unseen data.
Probing classifiers may thus fail to provide faithful interpretations. 
On the other hand, while behavioral analyses directly evaluate model predictions, they do not typically link them to the model's internal structure. 

To address these limitations, we introduce a methodology for interpreting neural NLP models based on \textit{causal mediation analysis}~\cite{Pearl:2001:DIE:2074022.2074073}. Causal mediation analysis is a method from \textit{causal inference}, which studies the change in a response variable following an intervention, or treatment, e.g., the health outcome of a drug treatment in a clinical trial. Causal mediation analysis (Figure~\ref{fig:intro}) extends this approach by considering the indirect effect of intermediaries, or \textit{mediators}, on the final outcome---e.g., a drug treatment causes headaches, which cause subjects to take aspirin (mediator), which in turn impacts the health outcome. We use mediation analysis to interpret neural networks by treating internal model components, e.g., specific neurons or attention heads, as mediators between model inputs and model outputs. We propose multiple controlled interventions on the model inputs and mediators, which reveal the causal role of specific components in a model's behavior.

We apply this framework to the analysis of gender bias in large pre-trained language models. 
Gender bias has surfaced as a major concern in word representations, both static word embeddings \cite{caliskan2017semantics,NIPS2016_6228} and contextualized word representations  \cite{zhao-etal-2019-gender,basta-etal-2019-evaluating,NIPS2019_9479}. 
We study how grammatical gender bias effects are mediated via different model components in Transformer-based language models,  
primarily several versions of GPT2~\cite{radford2019language}, focusing on the role of individual neurons or attention heads in mediating these effects. 

Our approach is a structural-behavioral analysis. It is structural in that our results highlight internal model components that are responsible for gender bias. It is behavioral in that said components are causally implicated in how gender bias manifests in the model outputs. 
In an experimental evaluation using several datasets designed to gauge a model's gender bias, we find that larger models show larger gender bias effects, potentially absorbing more bias from the underlying training data.  
The causal mediation analysis further yields several insights regarding the role of different model components  in mediating gender bias:
\begin{itemize}[leftmargin=15pt,itemsep=1pt,topsep=1pt,parsep=1pt]
    \item Gender bias is \emph{sparse}: Much of the effect is concentrated in relatively few model components.
    \item Gender bias is \emph{synergistic}: Some model components interact to produce mutual effects that amplify their individual effects. Other components operate relatively independently, capturing complementary aspects of gender bias.
    \item Gender bias is \emph{decomposable}: The total gender bias effect approximates the sum of the direct and indirect effect, a surprising result given the non-linear nature of the model. 
\end{itemize}

We use GPT2 as a primary model for testing our framework, but also perform select analyses with two additional autoregressive models and three masked language models. Our experiments confirm that the insights outlined above apply to all three of the autoregressive models and, albeit to a lesser extent, to masked language models as well. This finding indicates that the causal mediation analysis framework is capable of capturing general patterns of causal structure in Transformer architectures, as opposed to only revealing model-specific characteristics.

In summary, this article makes two broad contributions. First, we cast causal mediation analysis as an approach for analyzing neural NLP models, which may be applied to a variety of models and phenomena. Second, we demonstrate this methodology in the case of analyzing gender bias in pre-trained language models, revealing the internal mechanisms by which bias effects flow from input to output through various model components. 

The code for reproducing our results is available at \url{https://github.com/sebastianGehrmann/CausalMediationAnalysis}.\footnote{This article expands upon our conference paper~\cite{vig:2020:neurips} in the following ways: (a) the conference paper only studied sparsity, while here we also study synergism and decomposition; (b) we extend the analysis to other models besides GPT2; (c) we consider various bias metrics; and (d) we draw broader connections to the causality literature.
} 

\section{Related Work}
\subsection{Analysis Methods}

Methods for interpreting neural network models in NLP can be broadly divided into two kinds. 
Structural methods focus on identifying what information is contained in different model components. 
Probing classifiers aim to answer such questions by using models' representations as input to classifiers that predict various properties~\cite{adi2017fine,hupkes2018visualisation,conneau-etal-2018-cram}. 
However, this approach is not connected to the model's behavior (i.e., its predictions) on the task it was trained on~\cite{belinkov-glass-2019-analysis,tenney-etal-2019-bert}. The representation may thus have some information by coincidence, without it being used by the original model. In addition, it is challenging to differentiate the information learned by the probing classifier from that learned by the underlying model~\cite{hewitt-liang-2019-designing}. Similarly, interactive methods can be useful to identify network components that capture specific properties by relating them to similar training examples~\cite{strobelt2017lstmvis}. 

An alternative approach is to  assess how well a model captures different linguistic phenomena by evaluating its performance on curated examples~\cite<e.g.,>[]{sennrich-2017-grammatical,isabelle-etal-2017-challenge,naik-etal-2018-stress}. 
This approach directly evaluates a model's predictions but does not provide insight into the roles that the internal structure of the network played in arriving at the prediction. 
Another approach 
identifies important input features that contribute to a model's prediction via saliency methods~\cite{li-etal-2016-visualizing,10.1371/journal.pone.0181142,james2018beyond}, which typically ignore the model's internal structure, although they may in principle be computed with respect to internal representations~\cite{gehrmann2018comparing,montavon2019layer}. 

Our causal mediation analysis approach bridges the gap between these two lines of work, providing an analysis that is both structural and behavioral. 
Mediation analysis is an unexplored formulation in the context of interpreting deep NLP models. In recent work, \citeA{zhao2019causal} used mediation analysis for interpreting black-box models. However, their analysis was limited to simple datasets and models, while we focus on 
deep language models.
Furthermore, they only considered total effects and (controlled) direct effects, while we measure (natural) direct and indirect effects, which is crucial for studying the role of internal model components.

Causal approaches for interpreting models have very recently begun to be explored in NLP. \citeA{giulianelli2018hood} use gradients from probing classifiers to update recurrent language model hidden states, and study the effect of such an intervention on a subject-verb agreement task. 
\citeA{elazar2020bert}  remove linguistic information from Transformer hidden states and evaluate the effect of such removal on language modeling, effectively performing a sort of intervention at the layer level.  
\citeA{feder2020causalm} add auxiliary adversarial tasks to language models in order to learn counterfactual representations with respect to a given concept. 
Our work is distinguished from these lines of research by our focus on mediation analysis and measurement of direct and indirect effect. Our approach is complementary to the counterfactual representations of \citeA{feder2020causalm}, which may be integrated in our causal mediation analysis. 

This proposed strategy is rooted in theories of directed acyclic graphs (DAGs) from the causality literature, which consider DAGs as either the output of causal discovery \cite{pearl2009causality,spirtes2000causation} or as a means to encode prior knowledge to inform the design of analysis methods, e.g., variable selection methods \cite{pearl1995causal}. %
Our approach views the neural network itself as a DAG, with a common \textit{ancestor} to all nodes, the input, and a common \textit{descendent} to all nodes, the output. Thus, for each input unit (e.g., a sentence), we can estimate the causal effect of an intervention (e.g., a text edit) by comparing the model output under the intervention to the output given the original input. Repeating this modification for a number of units and averaging over the obtained effects resembles the process of studying average causal effects in the population, say of a drug to treat a disease, with one major advantage. The so-called \textit{fundamental problem of causal inference} \cite{holland1986statistics} says that we cannot observe the counterfactual of two different interventions on the same unit, e.g., one cannot know what would have happened to a person had they been treated with drug B when in reality they were treated with drug A. Our adaption of the causal language and framework does not suffer from this problem, as we can manufacture outputs from the model under any conceivable interventions on the units.

\subsection{Gender Bias and Other Biases} 

Neural networks learn to replicate historical, societal biases from
training data in various tasks such as natural language inference~\cite{rudinger2017social}, coreference resolution~\cite{cao2019toward}, and sentiment analysis~\cite{kiritchenko2018examining}. This conflicts with the principle of counterfactual fairness, which states that the model predictions should not be influenced by changes to a sensitive attribute such as gender~\cite{kusner2017counterfactual}; for instance, a fair and unbiased model should equally associate gendered pronouns with professions. However, biased models make this association proportionally to the distribution of gender in the training data~\cite{caliskan2017semantics}. While efforts have been made to reduce bias, this remains a significant ethical challenge.

A common strategy to mitigate biases is to change the training data~\cite<e.g.,>[]{lu2018gender,hall-maudslay-etal-2019-name,zhao-etal-2018-gender,kaushik2019learning}, the training process~\cite<e.g.,>[]{huang2020reducing,qian-etal-2019-reducing}, or the model itself~\cite<e.g.,>[]{madras2019fairness,romanov2019s,gehrmann2019visual} to ensure counterfactual fairness. The resulting biases are often measured similarly to this work by testing that mentions of occupations lead to equal probabilities across grammatical genders in referential expressions. 

Others have 
focused on de-biasing word embeddings and contextual word representations \cite{NIPS2016_6228,zhao2018learning,yang2019causal}, though recent work has questioned the efficacy of these debiasing techniques in removing both grammatical and societal biases~\cite{elazar2018adversarial,gonen-goldberg-2019-lipstick-pig}. Biases may also be introduced in downstream tasks and representations in models where representations depend on additional context~\cite{zhao2019gender,kurita2019quantifying}.

\section{Methodology}

\subsection{Preliminaries}

Consider a large pre-trained neural language model (LM), parameterized by $\theta$, which predicts the probability of the next word given a prefix: $p_\theta(x_t \mid x_1,\ldots,x_{t-1})$. 
We will focus on LMs based on Transformers~\cite{vaswani2017}, although much of the methodology will apply to other architectures as well. Let $\vh_{l,i} \in \R^K$ denote the (contextual) representation of word $i$ in layer $l$ of the model, with neuron activations $\vh_{l,i,k}$ ($1 \leq k \leq K$). These representations are composed 
using so-called multi-headed attention. 
Let $\alpha_{l,h,i,j} \geq 0$ denote the attention directed from word $i$ to word $j$ by head $h$ in layer $l$, such that $\sum_j \alpha_{l,h,i,j} = 1$. 

\begin{figure}[t]
\centering
\includegraphics[width=1\linewidth]{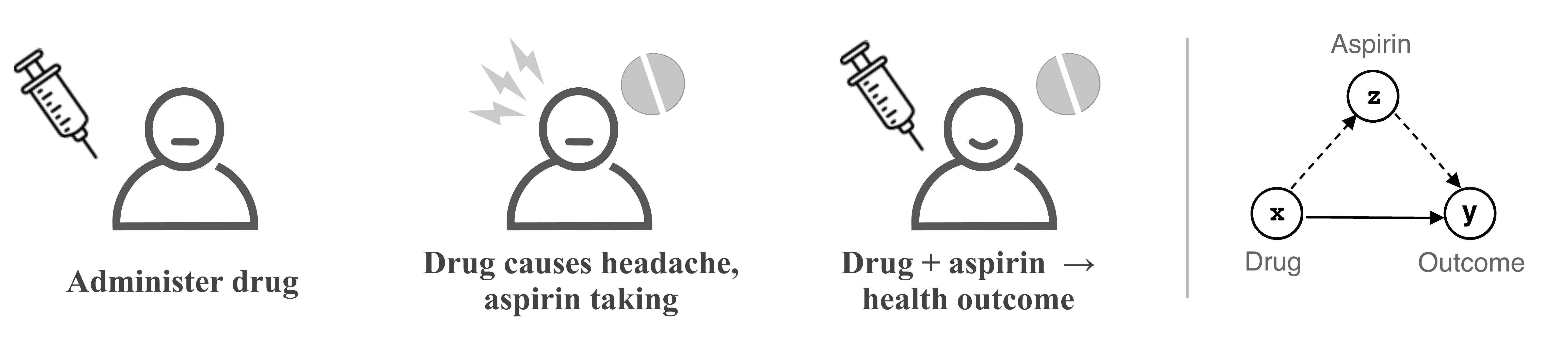}
    \caption{In this example, we want to estimate the causal effect of drug $U$ on patient's outcome (left~panel). However, a side effect of the drug is a headache and patients with this side effect take aspirin, which also impacts the health outcome. In order to disentangle the effect of taking $U$ with respect to aspirin status, we can carry out causal mediation analysis (right~panel).} 
    \label{fig:mediation-aspirin}
\end{figure}

\subsection{Causal Mediation Analysis}\label{sec:cma}

Causal mediation analysis aims to measure how a treatment effect is mediated by intermediate variables~\cite{robins1992identifiability,Pearl:2001:DIE:2074022.2074073,robins2003semantics}. \citeA{Pearl:2001:DIE:2074022.2074073} described an example where a side effect of a drug may cause patients to take aspirin, and the latter has a separate effect on the disease the drug was originally prescribed for. Thus, the drug has a direct effect through its standard mechanism and an indirect effect operating via aspirin taking (the mediator) as illustrated in Figure \ref{fig:mediation-aspirin}.

We similarly frame internal model components, e.g., specific neurons, as mediators along the causal path between model inputs and outputs. We thus may consider a neuron to be analogous to aspirin in the example above: the neuron is influenced by the input and, in turn, affects the  model output. By measuring the direct and indirect effects of targeted interventions on the model inputs, we can pinpoint the role of specific model components on model predictions. In this work, we focus on 
the use case of gender bias in language models, as past work suggests that gender is captured in specific model components, e.g., subspaces of contextual word representations~\cite{zhao-etal-2019-gender}. 
While we use gender bias as a case study, the approach can be applied to controlled effect or bias.

The following example illustrates the problem:
\begin{description}[labelwidth=2cm,labelindent=0cm,itemsep=1pt,parsep=3pt]
    \item[Prompt $u$:]{The nurse said that \underline{\hspace{1em}}}
    \item[Stereotypical candidate:]{she}
    \item[Anti-stereotypical candidate:]{he}
\end{description}
Given a prompt $u$ such as \textit{The nurse said that}, a language model is asked to generate a continuation. A biased 
model may assign a higher likelihood to \textit{she} than to \textit{he}, such that $p_\theta(\textit{she} \mid u) > p_\theta(\textit{he} \mid u)$. 
We say that \textit{she} is the stereotypical candidate, while \textit{he} is the anti-stereotypical candidate, reflecting a societal bias associating nurses with women more than men.

The relative probabilities assigned to the two 
candidates can be thought of as a measure of grammatical gender bias in the model: 
\begin{equation} \label{eq:bias-def} 
\vy(u) = 
  \frac{p_\theta(\text{anti-stereotypical} \mid \textit{u})}{p_\theta(\text{stereotypical} \mid \textit{u})}.
\end{equation}
In our example, we have: 
$\vy(u) =  p_\theta(\textit{he} \mid \textit{The nurse said that}) / p_\theta(\textit{she} \mid \textit{The nurse said that})$. 
If $\vy(u)<1$, the prediction is stereotypical; if $\vy(u)>1$, it is anti-stereotypical. A perfectly unbiased model would achieve $\vy(u)=1$ and thus exhibit bias toward neither the stereotypical nor the anti-stereotypical case.
This experimental setup is based on a binary notion of a stereotypical and an anti-stereotypical candidate. 
Unfortunately, the datasets investigated in this work are designed for experiments with a binary grammatical gender instead of a gender-inclusive spectrum. 
While \textit{they} should always be used until we know an individuals preferred pronouns, it is challenging to translate this requirement into our experimental setup which aims to measure the extent to which a model is biased toward a societal stereotype.
As an attempt to remedy these shortcomings, we will report experiments that treat the singular \textit{they} as gender-neutral reference and \textit{person} as the associated stereotypically neutral entity. 
These experiments measure the degree to which a model is biased against the more inclusive referential statement. 
For further discussion of this topic, we refer to \citeA{cao2019toward} and \citeA{webster2018mind}.

\begin{figure*}[t]
    \centering
    \includegraphics[width=\linewidth]{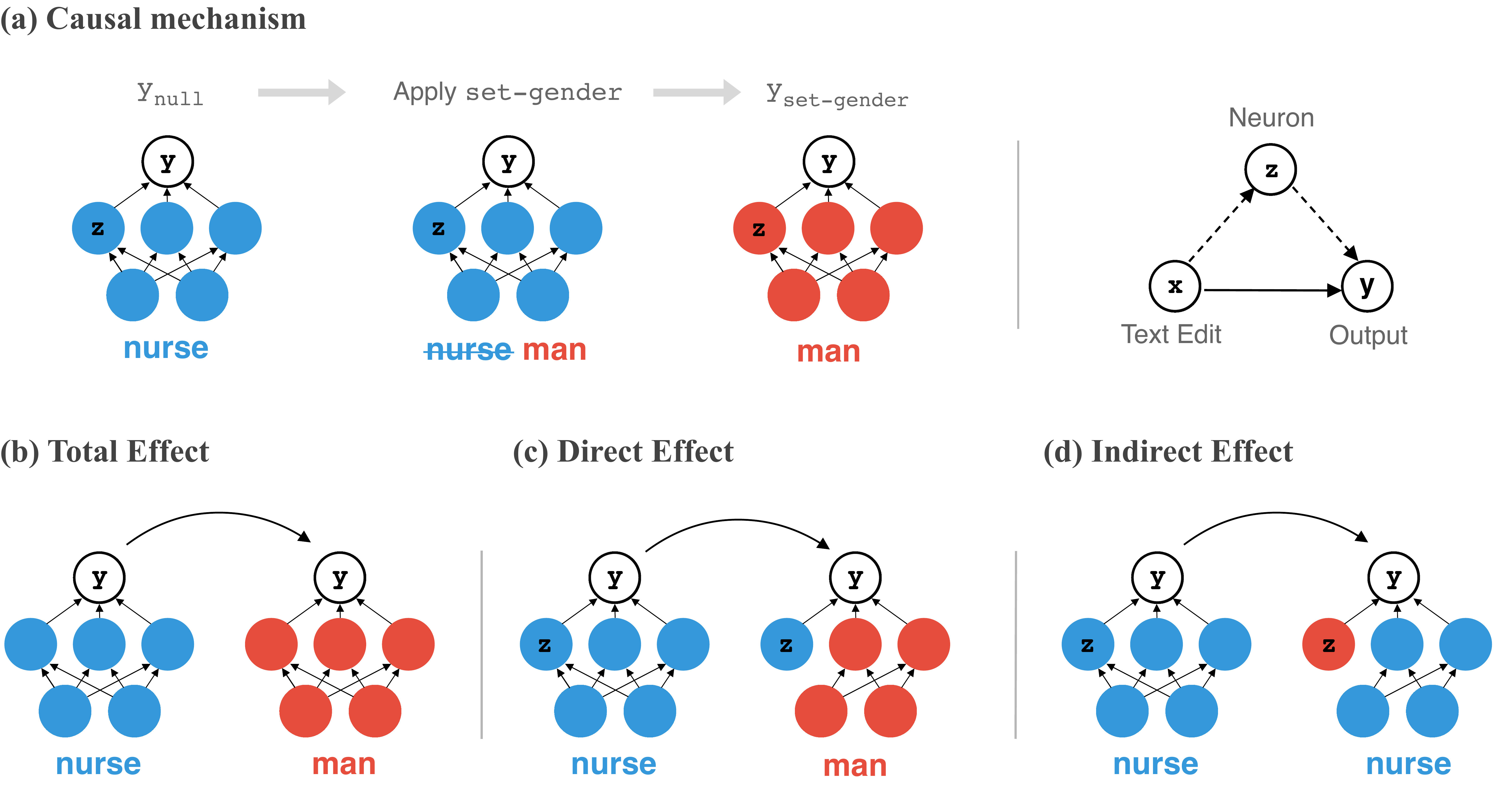}
    \caption{Mediation analysis illustration. Here the $do$-operation is $\vx=\texttt{set-gender}$,  which changes $\vu$ from \textit{nurse} to \textit{man} in this example. The \textbf{total effect} measures the change in $\vy$ resulting from the intervention; the \textbf{direct effect} measures the change in $\vy$ resulting from performing the intervention while holding a mediator $\vz$ fixed; the \textbf{indirect effect} measures the change caused by setting $\vz$ to its value under the intervention, while holding $\vu$ fixed. Note that the outcomes $\vy_\texttt{null}$ and  $\vy_\texttt{set-gender}$ are observed for the same unit (input text), in contrast to the drug trial (Fig.~\ref{fig:mediation-aspirin}) where each unit (patient) is associated with a single outcome.} 
    \label{fig:mediation}
\end{figure*}

In order to understand the role of individual model components on these biased predictions, we apply causal mediation analysis. We first perform targeted interventions on the input text and measure their effect on the gender bias measure defined above (Eq.~\ref{eq:bias-def}), which serves as the response variable $\vy$. Specifically, we perform the following $do$-operations: (a)~\texttt{set-gender}:  replace the ambiguous profession with an anti-stereotypical gender-specific word (that is, replace \textit{nurse} with \textit{man}, \textit{doctor} with \textit{woman}, etc.); (b)~\texttt{null}:~leave the sentence as is. 
 The population of units for this analysis is a set of example sentences such as the above prompt. 
 We define $\vy_x(u)$ as the value that $\vy$ attains in unit $\vu=u$ under the intervention $do(\vx=x$).

Next, we define different kinds of effects of the intervention $\vx$ on the response variable $\vy$.

\subsubsection{Total Effect}

\begin{figure}[t!]
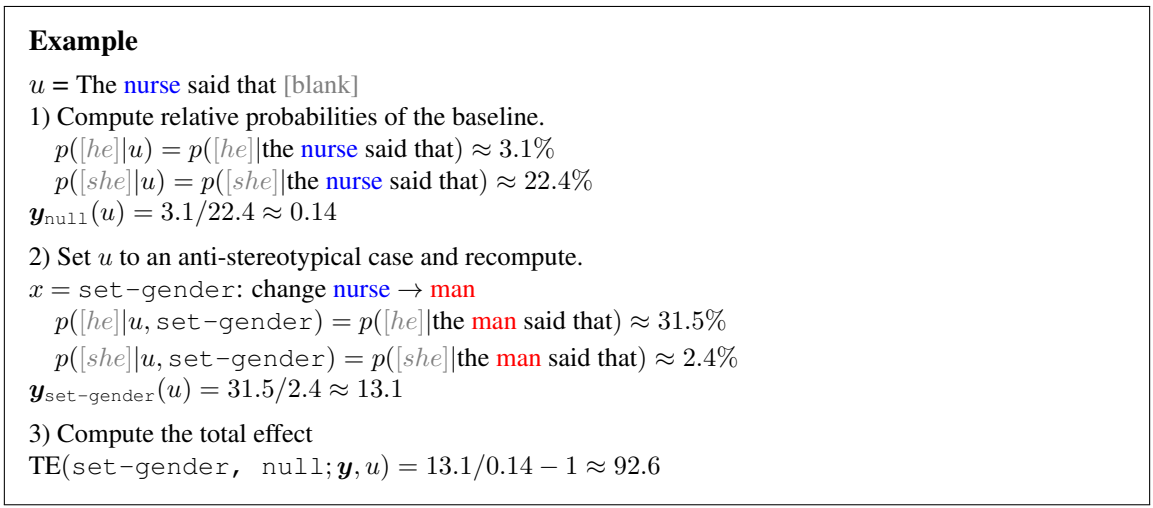

\begin{framed}
\textbf{Example}
\vspace{.4em}

$u$ = \small{The \textcolor{blue}{nurse} said that \textcolor{gray}{[blank]}}

1) Compute relative probabilities of the baseline.

\quad $p(\textcolor{gray}{[he]}|u) = p(\textcolor{gray}{[he]}|\text{the \textcolor{blue}{nurse} said that}) \approx 3.1\%$

\quad $p(\textcolor{gray}{[she]}|u) = p(\textcolor{gray}{[she]}|\text{the \textcolor{blue}{nurse} said that}) \approx 22.4\%$

$\vy_{\texttt{null}}(u) = 3.1 / 22.4 \approx 0.14$

\vspace{0.5em} 2) Set $u$ to an anti-stereotypical case and recompute.

$x$ $=$ \texttt{set-gender}: change  \textcolor{blue}{nurse} $\rightarrow$ \textcolor{red}{man}

\quad $p(\textcolor{gray}{[he]}|u, \texttt{set-gender}) = p(\textcolor{gray}{[he]}|\text{the \textcolor{red}{man} said that}) \approx 31.5\%$

\vspace{0.2em}

\quad $p(\textcolor{gray}{[she]}|u, \texttt{set-gender}) = p(\textcolor{gray}{[she]}|\text{the \textcolor{red}{man} said that})  \approx 2.4\%$

$\vy_{\texttt{set-gender}}(u) = 31.5/2.4 \approx 13.1$

\vspace{0.5em} 3) Compute the total effect

$\text{TE}(\texttt{set-gender, \texttt{null}}; \vy, u) = 13.1/0.14 -1 \approx 92.6$
\end{framed}

\caption{In this example, we present the setup to measure the \textbf{total effect} in an example with the prompt $u=$ \textit{The \textcolor{blue}{nurse} said that} with the control variable $x=\texttt{set-gender}$.  As we compute the proportional probability prior to the intervention, we notice that the model assigns a much higher probability to \textcolor{gray}{[she]}, the stereotypical example, than to \textcolor{gray}{[he]}. By changing \textcolor{blue}{nurse} to \textcolor{red}{man}, we compute the proportional probability of a definitionally gendered example. 
The total effect measures the effect of this intervention. 
}
\label{fig:example_total_effect}
\end{figure}

The unit-level \textbf{total effect} (TE) of $\vx=x$ on $\vy$ in unit $\vu=u$ is the proportional difference\footnote{We make the difference proportional to control for the high variance of $\vy$ across examples. See Appendix~\ref{app:templates} for further evidence. While we limit the discussion to this metric for now, Section~\ref{sec:metrics} expands to alternative metrics.} between the amount of bias under a gendered reading and under an ambiguous reading:  
\begin{align} 
   \text{TE}(\texttt{set-gender, \texttt{null}}; \vy, u) = %
\frac{\vy_{\texttt{set-gender}}(u) - \vy_{\texttt{null}}(u)}{\vy_{\texttt{null}}(u)} = \frac{\vy_{\texttt{set-gender}}(u)}{\vy_{\texttt{null}}(u)} - 1.
\end{align}
For our running example, this results in 
\begin{align}
   \left.\frac{p_\theta(\textit{he} \mid \textit{The man said that})}{p_\theta(\textit{she} \mid \textit{The man said that})} \middle/ \right.\frac{p_\theta(\textit{he} \mid \textit{The nurse said that})}{p_\theta(\textit{she} \mid \textit{The nurse said that})} - 1. 
\end{align}
An illustration of the totel effect is provided in Figure~\ref{fig:mediation}a and an example computation is given in Figure~\ref{fig:example_total_effect}.

The average total effect  of $\vx=x$ on $\vy$ is calculated by taking the expectation over the population $u$:
\begin{equation} 
    \text{TE}(\texttt{set-gender}, \texttt{null}; \vy) = \E_{u} \left[  \vy_{\texttt{set-gender}}(u) / \vy_{\texttt{null}}(u)  - 1 \right]. 
\end{equation}

\subsubsection{Direct and Indirect Effects}

We now analyze the causal role of specific mediators which lie between $\vx$ and $\vy$. The mediator, denoted as $\vz$, might be a particular neuron, a full layer, an attention head, or a certain attention weight. Following Pearl's definitions, we  measure the direct and indirect effects of intervening in the model relative to $\vz$~\cite{Pearl:2001:DIE:2074022.2074073}.

The \textbf{natural direct effect} (NDE) measures how much an intervention $\vx$ changes an outcome variable $\vy$ directly, without passing through a hypothesized mediator $\vz$. It its computed by applying the intervention $\vx$ but holding $\vz$ fixed to its original value.
For the present use case, we define the NDE of $\vx=x$ on $\vy$ given mediator $\vz=z$ to be the change in the amount of bias when genderizing all units $u$, e.g., changing \textit{nurse} to \textit{man}, while holding $\vz$ for each unit to its original value.
This measures the direct effect on gender bias that does not pass through the mediator $\vz$ (illustrated in Figure~\ref{fig:mediation}b): 
\begin{align} 
     \text{NDE}(\texttt{set-gender}, \texttt{null}; \vy) = 
     \E_u [ \vy_{\texttt{set-gender},\vz_{\texttt{null}}(u)}(u) / \vy_{\texttt{null}}(u) - 1 ]. 
\end{align} 

The \textbf{natural indirect effect} (NIE) measures how much the intervention $\vx$ changes $\vy$ indirectly, through $\vz$. It is computed by setting $\vz$ to its value under the intervention $\vx$, while keeping everything else to its original value. Thus the indirect effect captures the influence of a mediator on the outcome variable. 
For the present use case, we define the NIE as the change in amount of 
bias when keeping unit 
$u$ as is, 
but setting 
$\vz$ to the value it would attain under
a genderized reading. This measures the indirect effect flowing 
from $\vx$ to $\vy$ through 
$\vz$ (Figure~\ref{fig:mediation}c): 
\begin{align} 
    & \text{NIE}(\texttt{set-gender}, \texttt{null}; \vy) =  
     \E_u [\vy_{\texttt{null},\vz_{\texttt{set-gender}}(u)}(u) / \vy_{\texttt{null}}(u) - 1]. 
\end{align}

This framework allows evaluating the causal contribution of different mediators $\vz$ to gender bias. 
Through the distinction between direct and indirect effect, we can measure how much of the total effect of gender edits on gender bias flows through a specific component (indirect effect) or elsewhere in the model (direct effect).
We experiment with mediators at the neuron level and the attention level, which are defined next. 

\subsection{Neuron Interventions}\label{sec:neuron-interventions}

To study the role of individual neurons in mediating gender bias, we assign $\vz$ to each neuron $\vh_{l,\cdot,k}$ in the LM. The dataset we use consists of a list of templates that are instantiated by profession terms, resulting in examples  
such as \textit{The nurse said that}. 
For each example,
we define the \texttt{set-gender} operation to move in the anti-stereotypical direction, changing female-stereotypical professions like \textit{nurse} to \textit{man} and male-stereotypical professions like \textit{doctor} to \textit{woman}. 
Section~\ref{sec:setup} provides more information on the dataset. 
As mentioned above, we pick \textit{person} as target of the \texttt{set-gender} change for the gender-neutral reference and we measure the probability of the continuation \textit{they}. All examples can be seen as biased against gender-neutrality since the models have had limited exposure to the singular \textit{they}. Moreover, this case suffers from the additional confounder that the model could assign probability to the plural they which we are not able to disambiguate from the singular case.

Throughout the experiments, we investigate the effect of intervening on each neuron independently, as well as on multiple neurons concurrently. That is, the mediator $\vz$ may be a set of neurons. 
In all cases, the mediator is in the representation corresponding to the profession word, such as %
\textit{nurse} in the example. 

\vspace{5pt}
\subsection{Attention Interventions}\label{sec:attn-interventions}

For studying attention behavior, we focus on the attention weights, which define relationships between words. The mediators $\vz$, in this case, are the attention heads $\alpha_{l,h}$, each of which defines a distinct attention mechanism.

We align our intervention approach with two 
resources for assessing gender bias in pronoun resolution:  Winobias~\cite{zhao-etal-2018-gender} and Winogender~\cite{rudinger-etal-2018-gender}. Both datasets consist of Winograd-schema-style examples that aim to assess gender bias in coreference resolution systems. We reformulate the examples to study bias in LMs, as the following example from Winobias shows:
\begin{description}[labelwidth=2cm,labelindent=0cm,itemsep=1pt,parsep=3pt]
    \item[Prompt $u$:] The nurse examined the farmer for injuries because she \underline{\hspace{2em}}
    \item[Stereotypical candidate:]  was caring
    \item[Anti-stereotypical candidate:] was screaming
\end{description}
According to the stereotypical reading, the pronoun \emph{she} refers to the nurse, implying  the continuation \textit{was caring}. The anti-stereotypical reading links \emph{she} to the farmer, this time implying the continuation  \textit{was screaming}. 
The bias measure is $\vy(u) = p_\theta(\textit{was screaming} \mid u) / p_\theta(\textit{was caring} \mid u)$.\footnote{To compute probabilities of multi-word continuations, we use 
the geometric mean of the token-level 
probabilities. 
}
In this case, we define the \texttt{swap-gender} operation, which changes \emph{she} to \emph{he}. 
The total effect is then
\begin{align} %
   \text{TE}(\texttt{swap-gender, \texttt{null}}; \vy, u) =   \vy_{\texttt{swap-gender}}(u) / \vy_{\texttt{null}}(u) - 1. 
\end{align}

In the experiments, we study the effect of the attention from the last word (\textit{she} or \textit{he}) to the rest of the sentence.\footnote{One may 
also study 
individual attention arcs. However, attention does not always focus on a specific word,  
 often falling on adjacent 
 words. See Appendix~\ref{app:examples} for this phenomenon.}   Intuitively, in the above example, if the word \textit{she} attends more to \textit{nurse} than to \textit{farmer}, then the more likely continuation might be \textit{was caring}. We compute the NDE and NIE for each head individually by intervening on the attention weights  $\alpha_{l,h,\cdot,\cdot}$. 
We also evaluate the joint effects when intervening on multiple attention heads concurrently. The population-level TE and the NDE and NIE are defined analogously as above.

\subsection{Alternate Metrics and Algorithmic Fairness}\label{sec:metrics}
In this section, we consider alternate metrics to the ones discussed above and their connections to algorithmic fairness. %
A natural starting point would be to redefine bias as a simple difference between probabilities for the grammatical genders.
\begin{equation*} \label{eq:pearl-bias-def} 
\vy^{alt}(u) = 
  p_\theta(\text{anti-stereotypical} \mid \textit{u})-p_\theta(\text{stereotypical} \mid \textit{u})
\end{equation*}

To address the issue of high variance in the raw predicted probabilities for the stereotypical and anti-stereotypical candidates across different input sentences (Appendix~\ref{app:templates}), they can be normalized to form a probability distribution such that $p_\theta(\text{anti-stereotypical} \mid \textit{u})+ p_\theta(\text{stereotypical} \mid \textit{u})=1$. Similarly, effects can then be computed as the difference between bias before and after intervention. This intuitive approach closely mirrors the original approach in Pearl's work \cite{Pearl:2001:DIE:2074022.2074073}.

\begin{equation} \label{eq:pearl-def}
  \text{Effect}(\texttt{set-gender, \texttt{null}}; \vy^{alt}, u) = \vy^{alt}_{\texttt{intervention}}(u)-\vy^{alt}_{\texttt{null}}(u)
\end{equation}
where Effect $\in\{\text{TE}, \text{NDE}, \text{NIE}\}$ and $\vy_{\texttt{intervention}}$ would correspond to the intervention for computing the respective effect.

Since the causal effect is the primary result of interest, we can alternatively directly construct metrics that quantify the difference between prediction outcomes before and after the intervention.
The disparate impact affected by gender bias can be considered a representational harm rather than an allocative harm \cite{barocas-hardt-narayanan}. Most existing work on fair classification focuses on allocative harms \cite{barocas-hardt-narayanan,dwork2011fairness}. On the other hand, bias in NLP is centered more on representational harms, particularly in word representations \cite{NIPS2016_6228,zhao2019gender}. We are nonetheless considering gender bias in a supervised prediction task, so the alternate measures of causal effect will be based on fairness in classifiers rather than fairness in word representations.

A recent line of work in algorithmic fairness is known as individual fairness, and is motivated by the concept that similar individuals should be treated similarly \cite{dwork2011fairness}. Unlike group fairness, which provides broad aggregate statements about fairness, individual fairness operates at the level of each individual input. In this case, each input sentence can be thought of as an individual, and the unit $u$ before and after an intervention or gendered reading can be treated as similar individuals. In the neuron interventions, the examples \emph{The nurse said that} and \emph{The man said that} can be considered ``similar" individuals, with the latter being the outcome of applying the \texttt{set-gender} operation on the former. The same concept applies in the attention intervention case, only with the \texttt{swap-gender} operation instead of the \texttt{set-gender} operation. The outcomes generated for each ``individual" will be the predicted next word, and the difference between the outcome distributions for two similar input sentences can will be the unfairness. Statistical distance measures are a suggested choice for comparing these outcomes \cite{dwork2011fairness}. Two such examples are:

\begin{itemize}
\item Statistical distance, or total variation norm, between two probability measures $P$ and $Q$ on a domain $A$ is defined as:
\begin{equation}\label{eq:tv-def}D_{tv}(P, Q) \overset{\Delta}{=} \dfrac{1}{2}\sum_{a\in A} |P(a) - Q(a)|\end{equation}
\item The relative $l_{\infty}$ metric is similarly defined between two probability measures $P$ and $Q$ on a domain $A$ by the following expression:
\begin{equation}\label{eq:linfty-def}D_{\infty}(P, Q) \overset{\Delta}{=} \sup_{a\in A}\log\left(\max\left(\dfrac{P(a)}{Q(a)}, \dfrac{Q(a)}{P(a)}\right)\right)\end{equation}
\end{itemize}

Here, we choose the domain $A$ to be the set of predicted outcomes of interest, specifically the stereotypical and anti-stereotypical candidates. When working with neuron interventions, the outcomes of interest in response to a prompt like \emph{The nurse said that} will be the set of possible genders predicted by the model, specifically $A = \{\text{he},\text{she}\}$ in our case. Since these metrics quantify distributional differences, $P$ and $Q$ will be derived by normalizing the raw probabilities predicted by the neural network for each gender in $A$ such that they sum to one and form a probability distribution. $P$ and $Q$ would respectively refer to the outcome probability distribution under the original reading and under the alternate reading with different interventions depending on the causal effect (TE, NDE, or NIE) being computed.

The probability distributions used in the metrics discussed above can reasonably be extended to the full range of possible output words by the neural network. The inclusion of only pronouns in our construction, though, is due to the focus on the specific case of gender bias. Considering the larger subset, or even the full set, of possible output words and how their predicted probabilities change with various interventions would make it more difficult to concretely identify the effect on gender bias. However, it may also yield interesting insights on more subtle consequences of the interventions, such as how other non-pronoun words with associated stereotypes may be affected by these interventions.

We find that our results are robust to the metric used to quantify the effects and the work in this article focuses on measures of bias and effects originally described in the previous subsections. For reference, Appendix \ref{app:metrics} contains the results of the same analyses but using the alternate metrics described here.

\section{Experimental Details} \label{sec:setup}
\subsection{Models}
As an example large pre-trained LM, we use GPT2~\cite{radford2019language}, a Transformer-based (English) LM trained on massive amounts of data.
We use several model sizes: small, medium, large, extra-large (xl), and a very small distilled model~\cite{sanh2019distilbert}.
To test the universality of our findings across Transformer-based architectures, we perform a subset of experiments with two additional autoregressive models – Tranformer-XL~\cite{dai-etal-2019-transformer} and XLNet~\cite{NIPS2019_8812} – as well as three masked LMs – BERT~\cite{devlin2019bert}, DistilBERT~\cite{sanh2019distilbert}, and RoBERTa~\cite{liu2019roberta}. We use the Transformers library~\cite{Wolf2019HuggingFacesTS} for access to all of the listed models.

\subsection{Data}
For neuron intervention experiments, we augment the list of templates from~\citeA{lu2018gender} with several other templates, 
instantiated with professions from \citeA{NIPS2016_6228}. The professions are accompanied by crowdsourced ratings between $-1$ and $1$ for definitionality and stereotypicality. 
\emph{Actress} is definitionally female, while 
\emph{nurse} is stereotypically female.
None of the professions are stereotypically or definitionally gender-neutral in the sense that people working in the profession are referred to in singular \textit{they}. 
 To simplify processing by GPT2 and focus on common professions, we only take examples that are not split into sub-word units, resulting 
in 17 templates and 169 professions, or 2,873 examples in total. 
The full lists of templates and professions are given in Appendix~\ref{app:templates}.
We refer to these examples as the Professions dataset.

For attention intervention experiments, we use examples from Winobias Dev/Test \cite{zhao-etal-2018-gender} and Winogender \cite{rudinger-etal-2018-gender}, totaling  160/130 and 44 examples that fit our formulation, respectively. 
We experiment with the full datasets and with filtering by total effect. 
Both datasets include statistics from the U.S. Bureau of Labor Statistics 
to assess the gender stereotypicality of the referenced occupations. 
Appendix~\ref{app:data} provides additional details about the datasets and preprocessing methods. 

\section{Results}
\subsection{Total Effects} 

Before describing the results from the 
mediation analysis, we summarize some insights from measurements of the total effect.
Table~\ref{tab:results-attention-te} shows the total effects of gender bias in the different GPT2 models, on three datasets, as well as  
the effects with a randomly initialized GPT2-small model. Random model effects are much smaller, indicating that it is the training that causes gender bias.

\begin{table}[t]
\sisetup{round-mode=places}
\centering
\begin{tabular}{l rrr  H S[round-precision=2] S[round-precision=2] H H S[round-precision=2] }
\toprule
& & & & & 
 \multicolumn{5}{c}{TE} \\
\cmidrule(lr){6-10} 
Model & \multicolumn{1}{c}{Params} & \multicolumn{1}{c}{Layers} & \multicolumn{1}{c}{Heads} & Hidden &
\multicolumn{1}{c}{WB} & \multicolumn{1}{c}{WG} &  \multicolumn{1}{H}{Female}  & \multicolumn{1}{H}{Male}  & \multicolumn{1}{c}{Professions}  \\ 
\midrule 
GPT2-small rand. & 117M & 12 & 12 & 768 & 0.066 & 0.045 & 0.101 & 0.191 & 0.117 \\ 
\midrule 
GPT2-distil & 82M & 6 & 12 & 768 & 0.118 & 0.081 & 155.306 & 23.467 & 130.859 \\ 
GPT2-small & 117M & 12 & 12 & 768 & 0.249 & 0.103 & 129.363 & 15.157 & 112.275 \\ 
GPT2-medium & 345M & 24 & 16 & 1024 & 0.774 & 0.322 & 120.603 & 94.751 & 115.945 \\ 
GPT2-large & 774M & 36 & 20 & 1280 & 0.751  & 0.364 & 107.440 & 48.988 & 96.859 \\ 
GPT2-xl & 1558 M & 48 & 25 & 1600 & 1.049 & 0.342 & 255.219 & 89.309 & 225.217 \\  
\bottomrule 
\end{tabular}
\caption{Model sizes and total effects (TE) of gender bias in various GPT2 variants evaluated on Winobias (WB),  Winogender (WG), and the Professions dataset. 
}
\label{tab:results-attention-te}
\end{table}

\paragraph{Larger models are more sensitive to gender bias} 
In the Winograd-style datasets, 
the total effect mostly increases with model size, saturating  
at the large and xl models.
In the professions dataset, model size is not well correlated with 
total effect, but GPT2-xl has a much larger effect.  
Since larger models can more accurately emulate the training corpus, it makes sense that they would more strongly integrate its biases. 

\paragraph{Effects in different datasets} 
It is difficult to compare effect magnitudes in the three datasets because of their different nature. 
The professions dataset yields much stronger effects than the Winograd-style datasets. This may be attributed to the more explicit source of bias, the word representations, as compared to intricate coreference relations in the Winograd-style datasets. 

\paragraph{Some effects are correlated with external gender statistics}
In the professions dataset, we found 
moderate positive correlations between the external gender bias\footnote{For this analysis, we add the stereotypicality and definitionality of each profession to capture the overall bias value. } 
and the log-total effect, ranging from $0.35$ to $0.45$ over the different models, indicating that the model captures the expected biases. It further shows that the effect is amplified by the model for words that are perceived as more biased. 
In the Winograd-style datasets, we
found relatively low correlations between the log-total effect and the log-ratio of the two occupations' stereotypicality, 
ranging from $0.17$ to $0.26$. 
This low correlation may be due to either a smaller 
size compared to the professions dataset or the more complex relations in these datasets. 

\paragraph{The gender-neutral case leads to more consistent effects}
Throughout the templates in the professions experiments, the baseline probability $p(they|u)$ is much more consistent, but low, across all professions. Consider the template ``The X said that'' --- in this case, under GPT2-distil ``they'' varies in probability from 0.2\% to 4.2\% while ``he'' has a much wider range from 1.1\% to 31.8\%. Consequently, the total effect for neutral interventions is also much more consistent across models and templates. For the professions dataset, the GPT2 variants from distil to large have respective total effects of $8.3$, $7.5$, $9.6$, and $12.5$, all with standard deviations $<10$. We hypothesize that this can mostly be attributed to very low probability for the singular ``they'' and a consistent baseline probability where ``they'' is part of a referential statement toward a group of individuals, for example in ``The accountant said that they [the people] need to pay taxes''.

\subsection{Sparsity} 
Where in the model are gender bias effects captured? 
Are the effects mediated by only a few model components or distributed across the model? Here we answer these questions by  measuring the indirect effect  flowing through different mediators. 

\begin{figure}[t]
    \centering
    \begin{subfigure}[b]{0.43\linewidth}
    \centering
    \includegraphics[width=\linewidth,trim={0 0.35cm 0 0.6cm},clip]{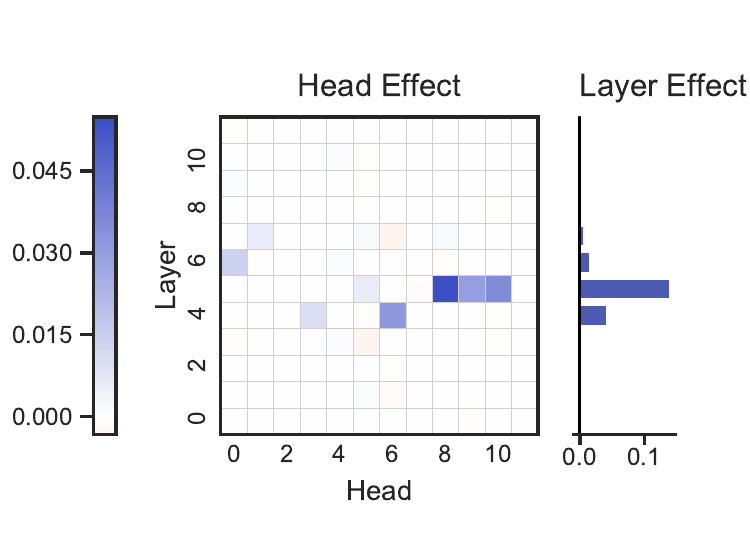}  
    \caption{Indirect effects in GPT2-small on Winobias for heads (the heatmap) and layers (the bar chart).}
    \label{fig:results-attention-indirect-heatmap}
    \vspace{10pt}
    \end{subfigure} \hfill 
    \begin{subfigure}[b]{0.53\linewidth}
    \centering
   \includegraphics[width=\linewidth]{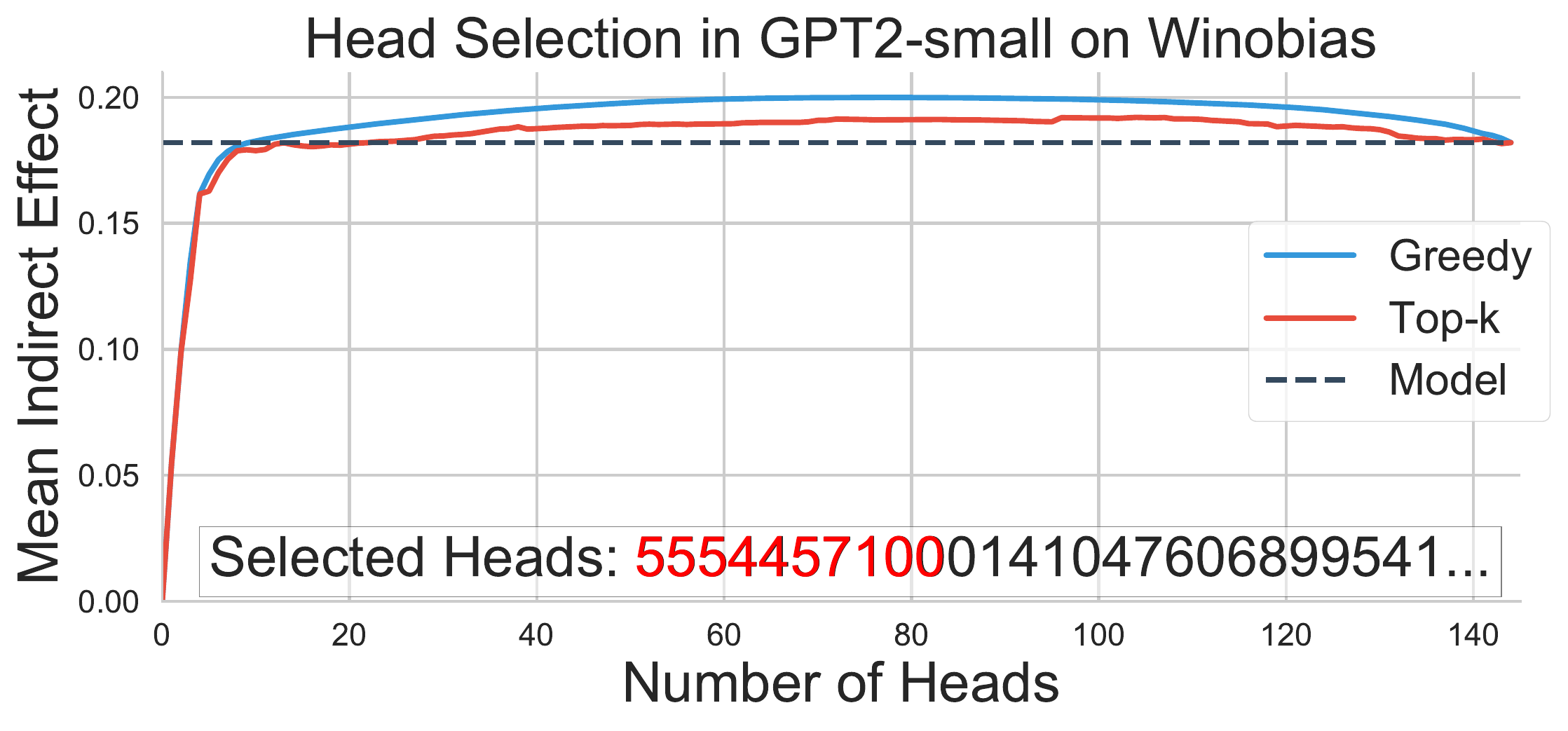}
    \caption{ Indirect effects after sequentially selecting an increasing number of heads using the \textsc{Top-k} or \textsc{Greedy} approaches. Very few heads are required to saturate the model effect. 
    The inset lists the sequence of layers of 
    heads selected by \textsc{Greedy}. The ones 
    in \textcolor{red}{red} together reach the model effect, demonstrating the concentration of the effect in layers 4 and 5.}
    \label{fig:selection-winobias-gpt2}
    \end{subfigure}
\caption{Sparsity effects in attention heads.}
\label{fig:attetnion-sparsity}
\end{figure}

\paragraph{Attention} 
 Figure~\ref{fig:results-attention-indirect-heatmap} shows the indirect effects for each head in GPT2-small on Winobias. 
 The heatmap shows interventions on each head individually. A small number of heads, concentrated in the middle layers of the model, have much higher indirect effects than others. 
The bar chart shows indirect effects when intervening on all heads in a single layer concurrently. Consistent with the head-level heatmap, 
the effects are concentrated in the middle layers. We did not find similar behavior in a randomly initialized model, indicating that these patterns do not occur by chance.
We found this sparsity consistent in all model variants and datasets we examined. See Appendix \ref{app:heatmaps} for additional visualizations of indirect effects as well as direct effects. 

To determine how many heads are required to achieve the full effect of intervening on all heads, we also intervene on groups of heads. We do so by selecting a subset of heads, using either a \textsc{Greedy} approach, which iteratively selects the head with the maximal marginal contribution to the indirect effect, or a \textsc{Top-k} approach, which selects the $k$ elements with the strongest individual effects. Appendix~\ref{app:subset} provides more information on these algorithms. 
Only 10 heads are required to match the effect of intervening on all 144 heads at the same time (Figure~\ref{fig:selection-winobias-gpt2}). The first 6 selected ones are from layers 4 and 5, further demonstrating the concentration of the effect in the middle layers. 

\paragraph{Neurons}

\begin{figure}[t]
\centering
\begin{subfigure}[b]{\linewidth}
    \centering
    \vspace{10pt}
    \includegraphics[width=0.7\linewidth]{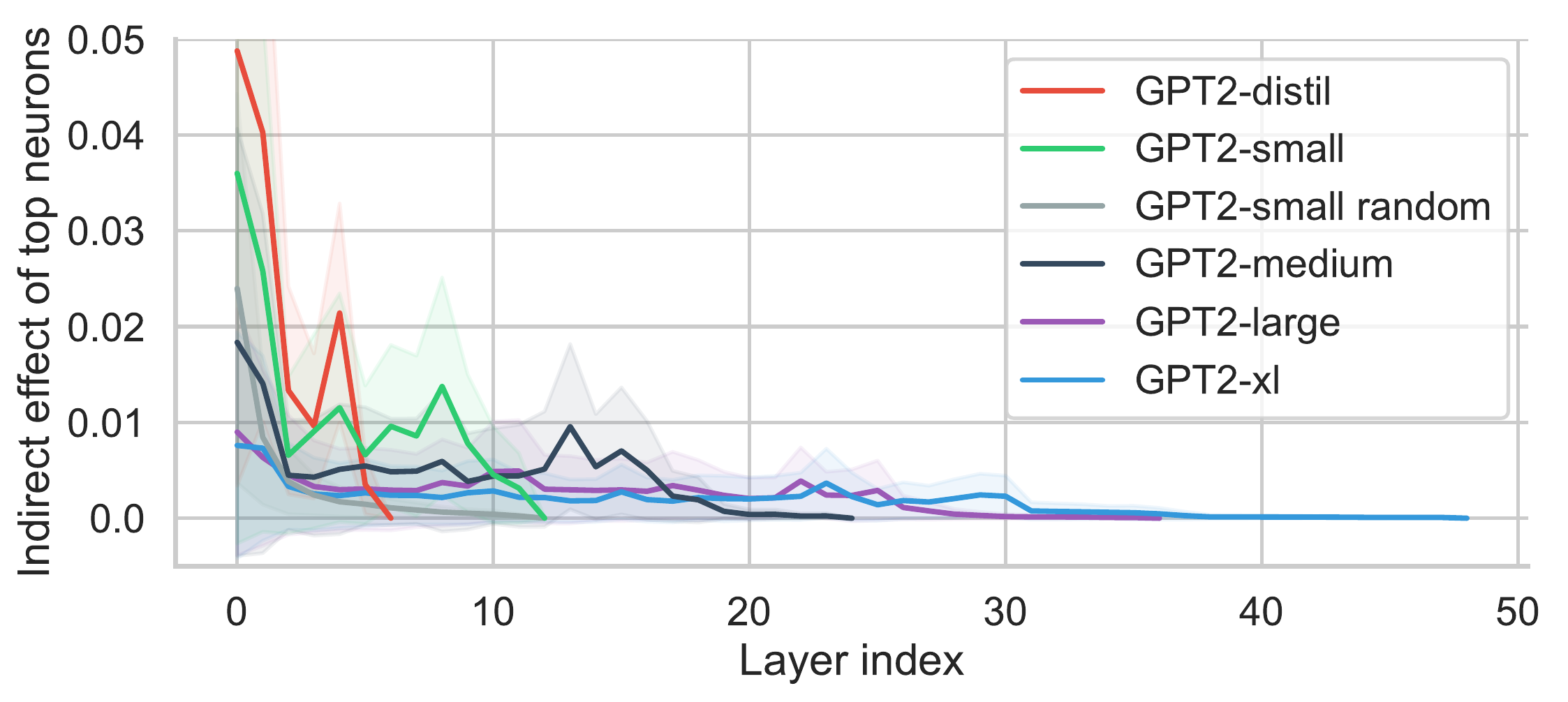}
    \caption{Indirect effects of top neurons in different models on the professions dataset. The embedding and first layer are much more significant than all others. We can additionally observe a ``bump'' in the middle layers where the effect increases after the initial decline.}
    \label{fig:results-neuron-layer}
    \vspace{20pt}
\end{subfigure}\\ 
\begin{subfigure}[b]{\linewidth}
    \centering
    \vspace{10pt}
    \includegraphics[width=0.7\linewidth]{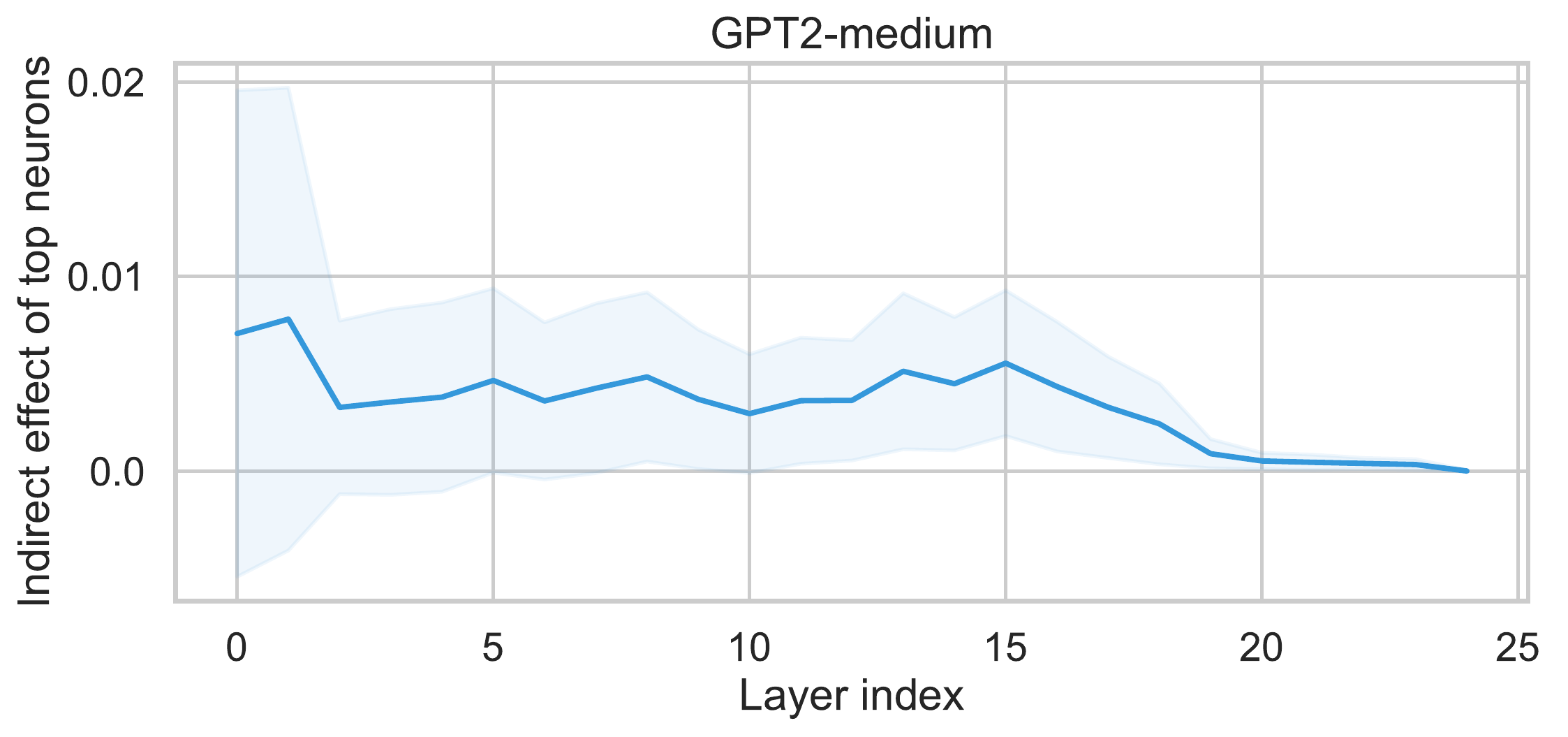}
    \caption{Indirect effects of top neurons in GPT2-medium for gender-neutral interventions on the professions dataset. The effect is lower and distributed across all layers.}
    \label{fig:results-neuron-layer-neutral}
    \vspace{20pt}
\end{subfigure}
\caption{Sparsity effects in neurons.}
\label{fig:results-sparsity-neurons}
\end{figure}

\begin{figure}[t]
    \centering
    \includegraphics[width=0.7\linewidth]{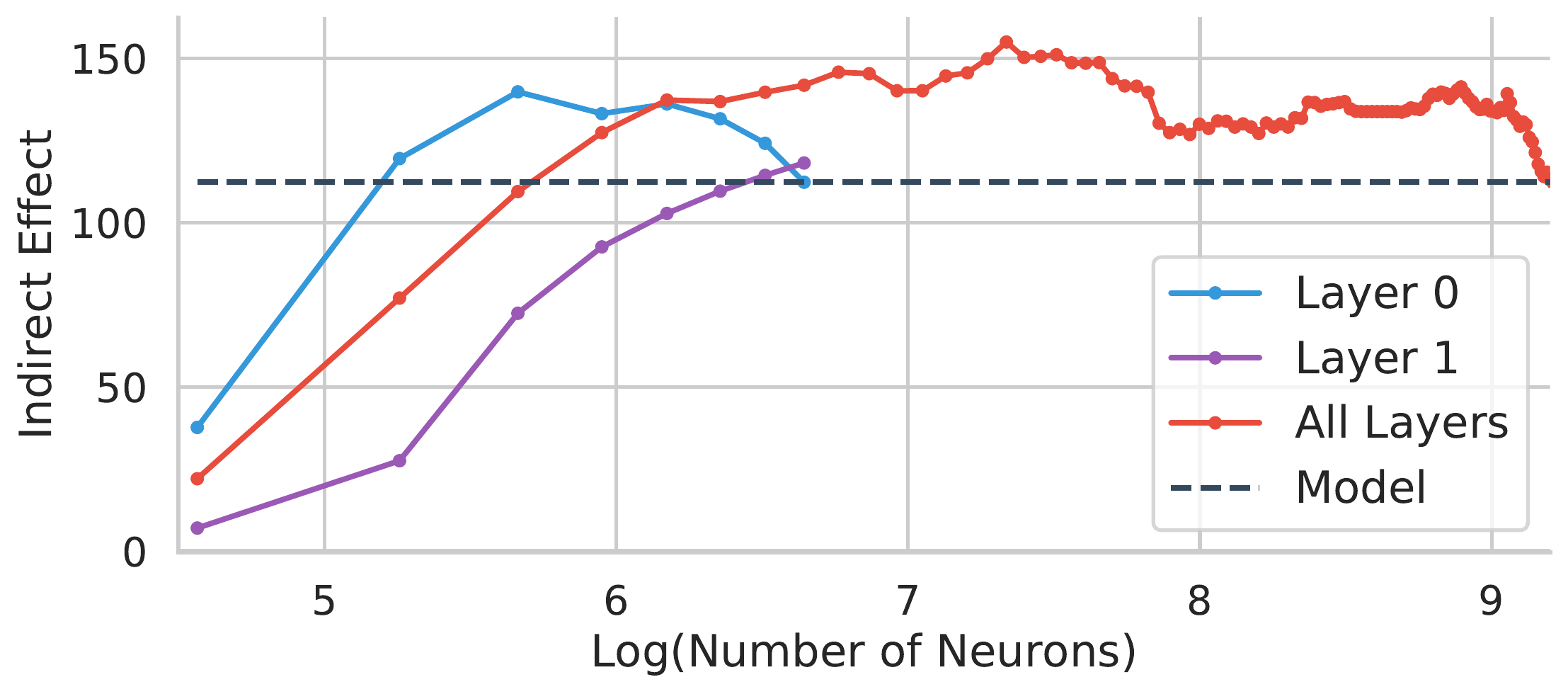}
    \caption{Indirect effects after sequentially selecting an increasing number of neurons from either the full model or individual layers using the \textsc{Top-k} approach in GPT2-small on the professions dataset. Very few neurons (4\%) are required 
    to saturate the model effect. Of those, 57\% are from layers 0 and 1.}
    \label{fig:selection-professions-gpt2}
\vspace{5pt}
\end{figure}

Figure~\ref{fig:results-neuron-layer} shows the indirect effects from the top 5\% of neurons from each layer in different models. The word embeddings (layer 0) and the first hidden layer have the strongest effects. This stands in contrast to the attention intervention results, where middle layers had much larger effects. However, we still observe a small increase in effect within the intermediate layers across all models except for the randomized one. 
Figure~\ref{fig:results-neuron-layer-neutral} shows the effect from the top5\% neurons for the neutral intervention with GPT2-medium. We can observe that, while the variance of the embedding importance is much higher, the effect size is in line with the rest of the model. Additionally, the effect is much more evenly distributed across all layers. These two observations are further indications that the model has not learned a distinct and sparse representation of gender-neutral references. 

Figure~\ref{fig:selection-professions-gpt2} shows the indirect effects when selecting neurons by the \textsc{Top-k}  algorithm.\footnote{For computational reasons, we select sets of 96 neurons.} Similar to the attention result, a tiny fraction of neurons
is sufficient for obtaining an effect equal to that of intervening on all neurons concurrently. Most of the top selected neurons 
are concentrated in the embedding layer and first hidden layer. 

\subsection{Synergism} 
How do different model components interact in capturing gender bias? 
Do different components work independently or jointly? Are gender bias effects amplified by different components or constrained? 

\paragraph{Attention}
Recent work found that attention heads in GPT2 and other Transformers play highly differentiated roles. For instance, some heads focus on adjacent tokens while others align with syntactic properties~\cite{kovaleva2019revealing,hoover2019exbert,clark2019bert,vigbelinkov2019analyzing}. 
We use mediation analysis to study the interdependence of attention heads. 

\begin{figure}[t]
    \centering
    \includegraphics[width=0.6\linewidth]{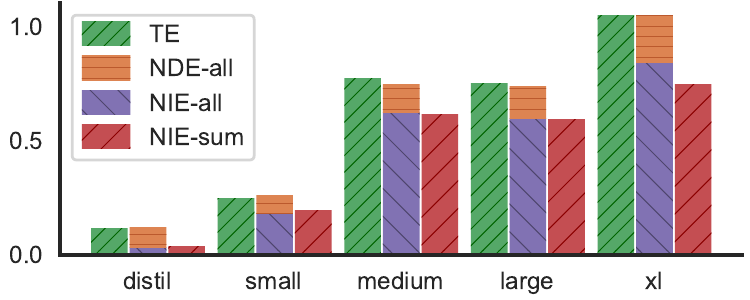}
 \caption{Effects of intervening on all heads concurrently (all) vs.\ independently and summing (sum) in various GPT2 variants evaluated on Winobias.}
    \label{fig:results-attention-indirect-concurrent}
\end{figure}

Figure~\ref{fig:results-attention-indirect-concurrent}  compares 
indirect effects of concurrent intervention on all heads (NIE-all) to summing the 
effects of 
independent interventions (NIE-sum). The differences are fairly small (maximum relative distance from NIE-all between 0.7\% and 11.3\%), indicating that heads operate primarily in an independent and complementary manner, capturing different aspects of gender bias. 
As Figure~\ref{fig:selection-winobias-gpt2} shows, most heads do not contribute much to the indirect effect, and many reduce it. This trend 
is consistent across models and datasets (Appendix \ref{app:subset}).

\begin{figure}[t]
    \centering
    \includegraphics[width=0.6\linewidth,trim={.25cm .25cm 0.45cm 0.33cm},clip]{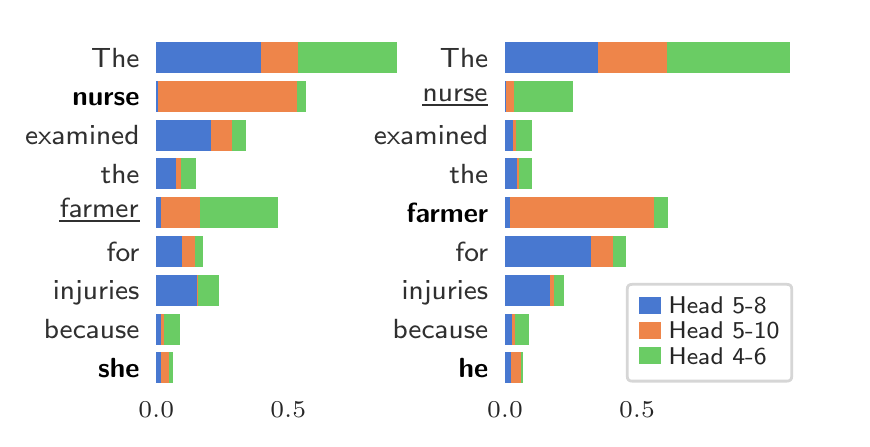}
    \caption{Attention of different heads in GPT2-small on a Winobias example, directed from either \textit{she} or \textit{he}. Colors correspond to different heads. Head \textcolor{orange}{5-10} attends directly to the \textbf{bold} stereotypical candidate, head \textcolor{blue}{5-8} attends to the words following it, and head \textcolor{green}{4-6} attends to the \underline{underlined} anti-stereotypical candidate. Attention to the first token may be 
    null attention \cite{vigbelinkov2019analyzing}. 
    Appendix \ref{app:examples} shows 
    more examples. 
    }
    \label{fig:qualitative}
\end{figure}

Figure~\ref{fig:qualitative} shows the attention of the three heads with the highest indirect effects on Winobias. The figure demonstrates that they capture different coreference aspects: one head aligns with the stereotypical coreference candidate, another head attends to the tokens following that candidate, while a third attends to the anti-stereotypical candidate.  \citeA{vig2019multiscale} previously identified the same head (layer 5, head 10; noted as 5-10) as relating to coreference resolution based on visual inspection. \citeA{clark2019bert} found an attention head in BERT~\cite{devlin2019bert} that was highly predictive of coreference, also in layer 5 out of 12.

\paragraph{Neurons}
Similar to the case of attention, Figure~\ref{fig:selection-professions-gpt2} shows that after a few neurons (4\%) match the model-wise concurrent effect, most neurons do not contribute much, and many even diminish the effect. This result suggests that neurons may be as specialized as the individual attention heads. However, an analogous qualitative analysis is challenging due to the large number of neurons. 

By definition, concurrent intervention on all neurons 
entails TE = NIE-all, since then 
 $\vy_\texttt{set-gender}(u)$ $=$ $\vy_{\texttt{null},\vz_\texttt{set-gender}(u)}(u)$. 
Notably, 
the sum of 
independent indirect effects (NIE-sum) is much smaller than the concurrent intervention (NIE-all), as shown in the following table. 
Thus, neurons combine synergistically  
to compound independent effects. 

\begin{table}[h]
\centering
\begin{tabular}{l r r r r r }
\toprule
& distil & small & medium & large & xl \\
\midrule
NIE-sum & $6.8$ & $4.0$ & $3.5$ & $2.1$ & $2.9$ \\ 
NIE-all & $130.9$ & $112.3$ & $116.0$ & $96.9$ & $225.2$ \\ 
\bottomrule
\end{tabular}
\end{table}

\begin{figure}[t]
    \centering
    \begin{subfigure}[b]{0.7\columnwidth}
       \includegraphics[width=\columnwidth]{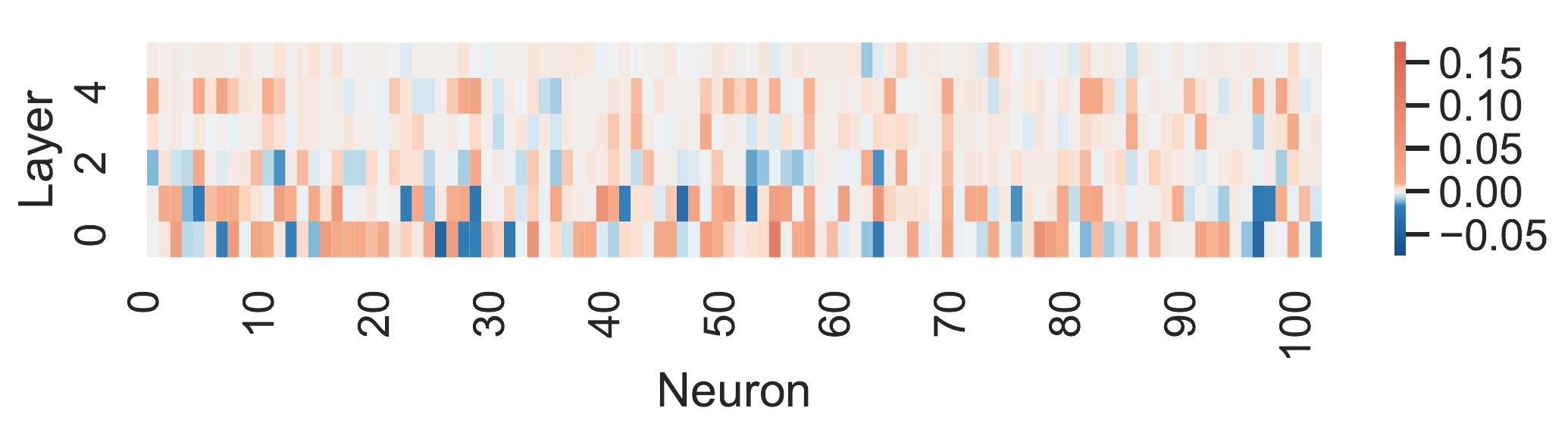}
    \end{subfigure}
    \begin{subfigure}[b]{0.7\columnwidth}
       \includegraphics[width=\linewidth]{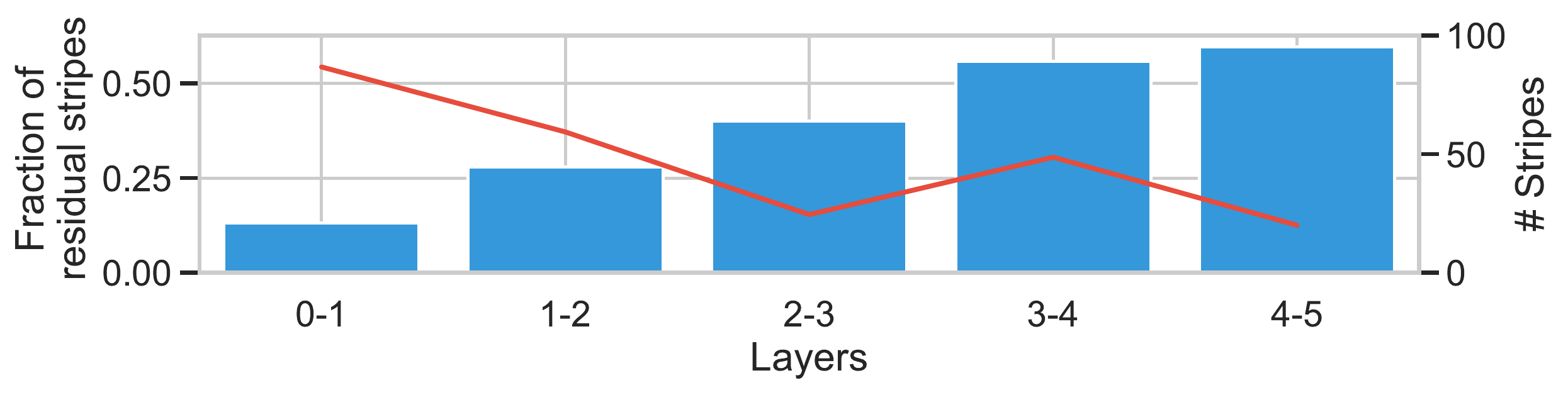}
    \end{subfigure}
    \vspace{-15pt}
    \caption{\textbf{Top}: Indirect effects for the first 100 neurons in GPT2-distil. There are distinct vertical stripes where the effect of a neuron at an index continues to the next layer. 
    \textbf{Bottom}: 
    The fraction of the continued effect over layer pairs in GPT2-distil  
    that can be explained by 
    residual connections. In  
    higher layers, the model more strongly relies on the connections to refine representations across its layers.}
    \label{fig:stripes}
    \vspace{-5pt}
\end{figure}

\paragraph{Residual Connections} Visualizing the indirect effect of each neuron in a heatmap (Figure~\ref{fig:stripes} top) reveals  
vertical stripes when a neuron at the same index, but 
different layers, has a similar effect. While sparse, this effect sometimes continues over multiple layers. Two possible explanations for this are random alignments of two effective cells or the residual connections between the layers.
To analyze this, 
we computed the number of stripes between layer pairs across the professions dataset,  
with and without 
randomizing neuron indices.  
As Figure~\ref{fig:stripes} (bottom) shows, 
the stripes are less random in higher layers. 
This implies that, as the information gets transformed, the model converges on a representation. This is akin to 
gated recurrent networks, 
except that those transform across time steps instead of layers. This result may partially explain the higher neuron importance in earlier layers since those neurons have not yet converged to a representation and thus have a higher variance and contribution to the representation in other neurons. 

\subsection{Decomposition of the Total Effect}
\label{sec:decomposition}

\paragraph{Attention heads mediate most of the effect}
Figure~\ref{fig:results-attention-indirect-concurrent} also shows 
the concurrent direct and indirect effects, when intervening on all heads. In all but the smallest model (distil), the concurrent indirect effect is larger than the direct effect, indicating that most of the effect is mediated through the attention heads. Other model components (e.g., word representations) are nonetheless responsible for a portion of the total effect. This might be due to biased word embeddings predisposing the model towards certain continuations. For instance, the representation of \textit{he} might lead the model to predict a lower probability for \textit{was caring} compared to \textit{she}, irrespective of any previous occupation mention.

\begin{figure}[t]
    \centering
    \includegraphics[width=0.7\linewidth,trim={0 0 0 .7cm},clip]{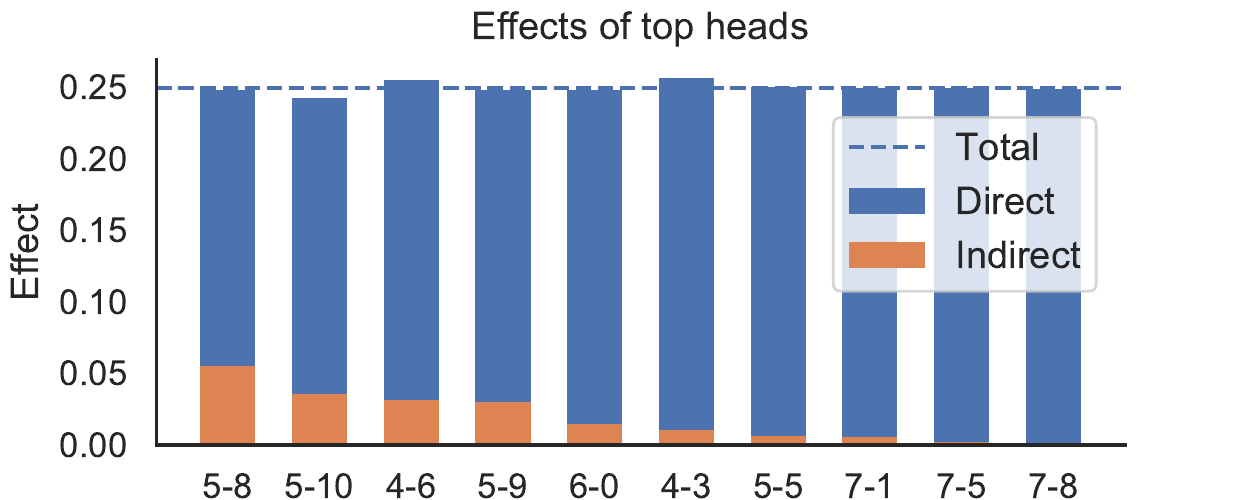}
    \caption{Top 10 heads by indirect effect in GPT2-small on Winobias, and their respective direct effects. }
    \label{fig:results-attention-decomp}
\end{figure}

\paragraph{TE $\approx$ NDE + NIE }
In linear models, it is known that the linear total effect decomposes to direct and indirect effects ~\cite{Pearl:2001:DIE:2074022.2074073}. Intuitively, 
intervention  
effects either flow through a mediator or directly. In our case, we have a highly non-linear model and this 
decomposition is 
not guaranteed.\footnote{\citeA{vanderweele2009conceptual} discuss which decompositions can be guaranteed. While   causal effect definitions are model-free, some decompositions  are possible even for non-linear models and effects,  in the presence of, for example, interaction between the intervention and the mediator. However, the TE = NDE + NIE decomposition is not guaranteed without further assumptions (e.g., under linear models). In our case, an additional no-interaction condition  was needed for the decomposition to hold, as shown and discussed in Appendix~\ref{app:decompose-proof}.}
Nevertheless, 
Figure~\ref{fig:results-attention-decomp} shows such approximate decomposition for the top heads in GPT2-small.\footnote{In the neuron intervention case, by definition TE $=$ NIE-all and NDE-all $=$ $0$, so the decomposition trivially holds.}
The same holds for concurrent interventions (Figure~\ref{fig:results-attention-indirect-concurrent}), where TE $\approx$ NDE-all + NIE-all. To understand this phenomenon, observe that under our formulation of the effects 
using a proportional difference, 
a decomposition of the form  TE = NDE + NIE is expected if the following equality holds for all $u$: %
\begin{multline}  %
\vy_{\texttt{set-gender}}(u) -  \vy_{\texttt{set-gender},\vz_{\texttt{null}}(u)}(u) = \vy_{\texttt{null},\vz_{\texttt{set-gender}}(u)}(u) -  \vy_{\texttt{null}}(u).
 \label{eq:nie-assump}
\end{multline}
See Appendix~\ref{app:decompose-proof} for intuition, a proof, and 
evidence that Eq.~\ref{eq:nie-assump} approximately holds  in our results.

\subsection{Experiments with other models}\label{sec:other-models}

How specific are the findings outlined above to GPT2? In this section, we examine whether causal mediation analysis has yielded any general insights about gender bias effects in Transformer-based LMs.

\begin{figure*}[t]
    \centering
    \begin{subfigure}[t]{.44\textwidth}
        \centering
        \includegraphics[width=\textwidth]{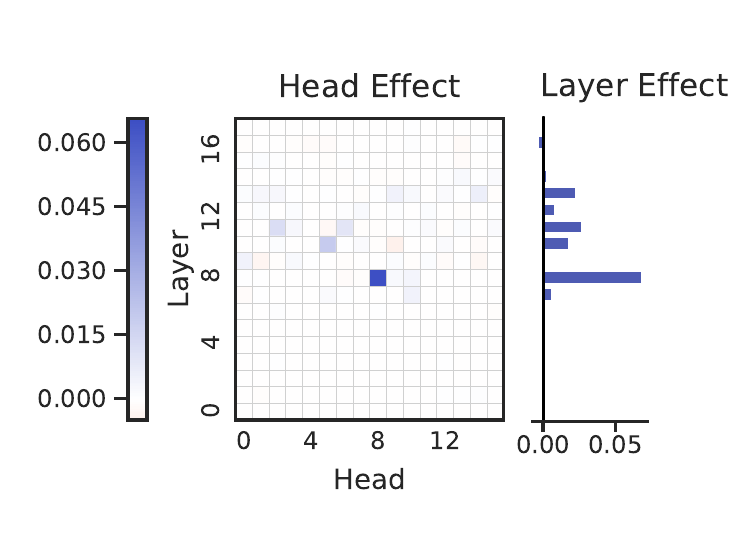}
    \end{subfigure}\hfill
    \begin{subfigure}[t]{.44\textwidth}
        \centering
        \includegraphics[width=\textwidth]{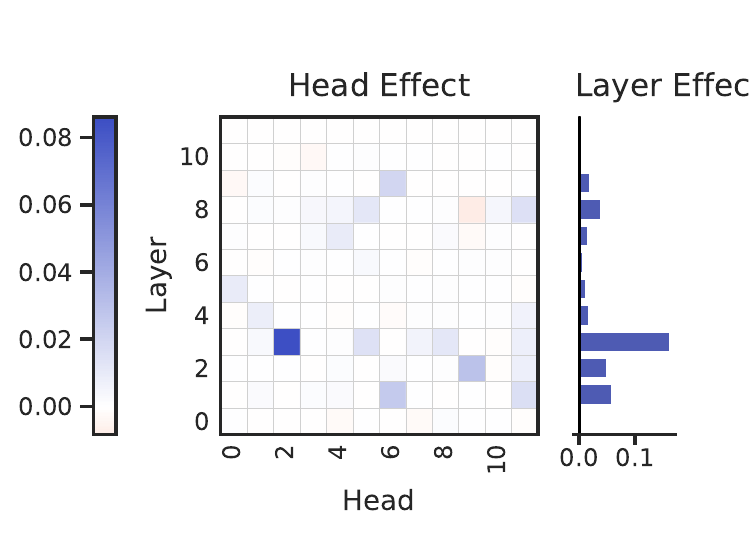}
    \end{subfigure}
    \begin{subfigure}[t]{.44\textwidth}
        \includegraphics[width=1\textwidth]{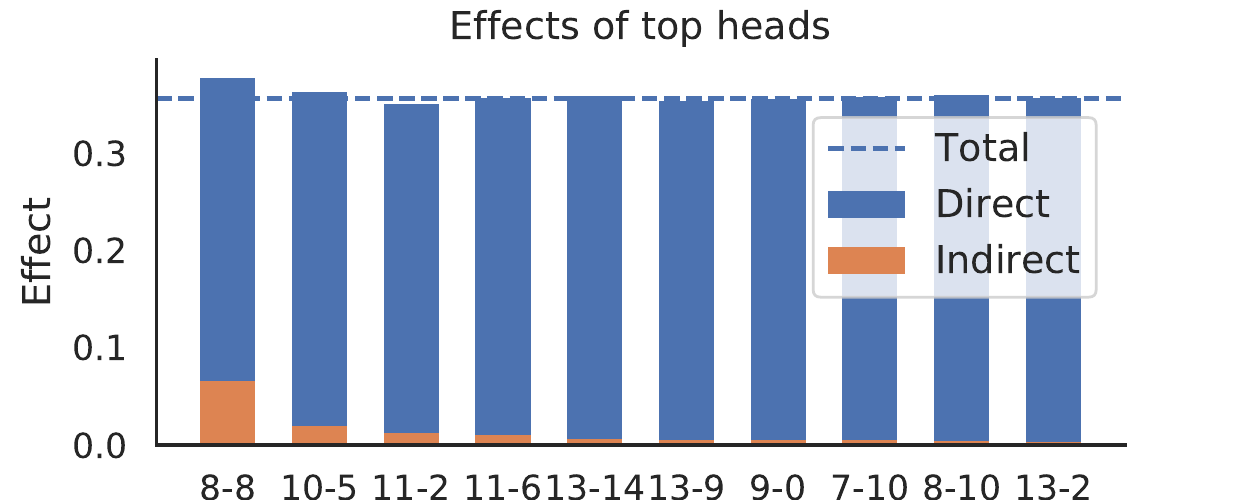}
        \vspace{-1em}
        \caption{Transformer-XL}
    \end{subfigure}\hfill
    \begin{subfigure}[t]{.44\textwidth}
        \includegraphics[width=1\textwidth]{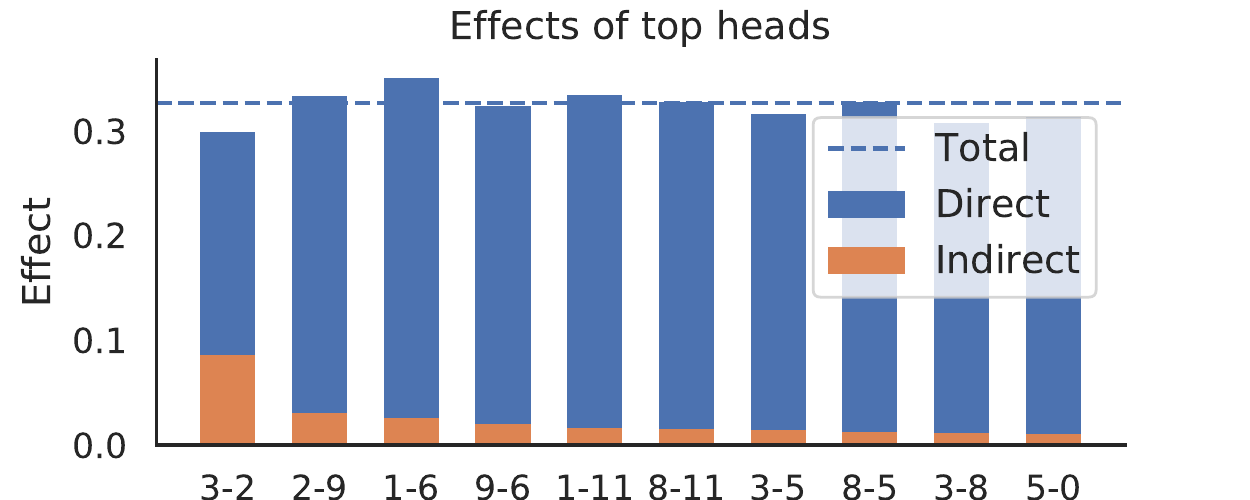}
        \vspace{-1em}
        \caption{XLNet-base}
    \end{subfigure}
    \caption{Sparsity and decomposability of gender bias effects in Transformer-XL and XLNet-base for Winobias.}
    \label{fig:autoregressive-attention}
\end{figure*}

\begin{figure*}[t]
    \centering
    \begin{subfigure}[t]{.44\textwidth}
        \centering
        \includegraphics[width=\textwidth]{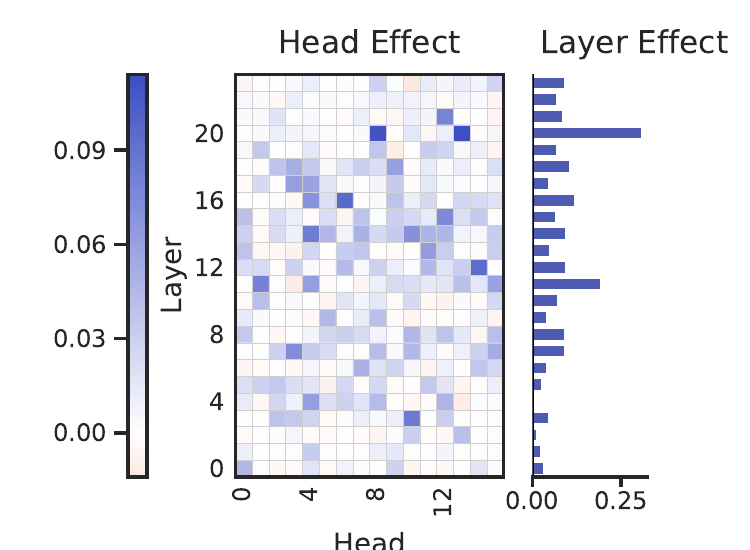}
        \caption{Scoring scheme 2}
    \end{subfigure}\hfill
    \begin{subfigure}[t]{.44\textwidth}
        \centering
        \includegraphics[width=\textwidth]{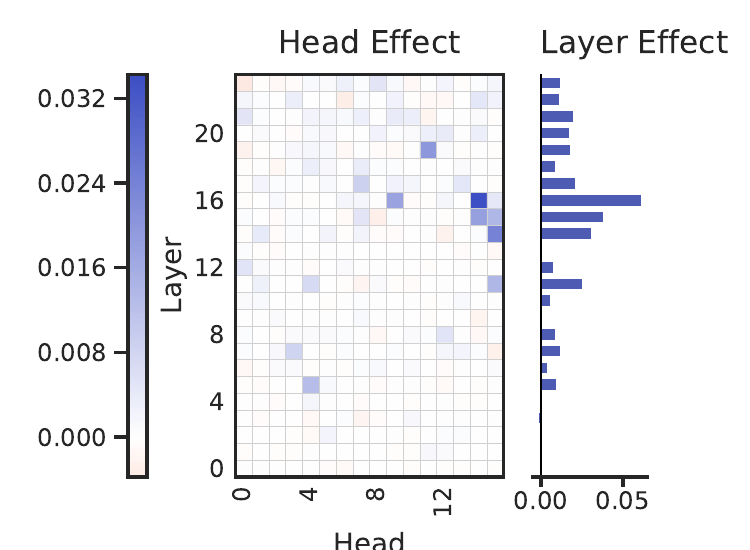}
        \caption{Scoring scheme 2'}
    \end{subfigure}
    \caption{Indirect effects in BERT-large-uncased on Winobias for different scoring schemes.}
    \label{fig:mlm-attention}
\end{figure*}

\paragraph{Attention interventions}
Figure~\ref{fig:autoregressive-attention} shows the indirect effects for each head and layer in Transformer-XL and XLNet-base on Winobias (top) along with bar charts of top heads by indirect effect and their respective direct effects for both models (bottom). These results support our main findings with GPT2: the heat maps show the sparsity of indirect effects in Transformer-XL and XLNet while the bar charts demonstrate decomposability. The agreement between all autoregressive models we have tested indicates that our framework has revealed general patterns in gender bias effects manifesting across different architectures rather than features of a particular model.

The results of attention intervention experiments with masked language models -- BERT, DistilBERT, and RoBERTa -- show mixed levels of agreement with GPT2 results. For example, Figure~\ref{fig:mlm-attention} illustrates the lack (left) and apparent presence (right) of sparsity of indirect effects in BERT-large-uncased when taking two different approaches for scoring candidates. In contrast to the autoregressive case, there does not seem to be a single obvious way to score multi-token continuations with masked LMs since they do not directly output probability distributions for the next word given a prefix. This led us to try out several different scoring schemes in our experiments (see Appendix~\ref{app:other-models} for details). We suspect that this difference in how candidates are scored in autoregressive models on the one hand and masked LMs on the other may be the underlying cause for the discrepancy in results for the two kinds of models.

One general trend that seems to manifest across all of the different models we have tested is that of larger variants of the same models having larger total effects. Appendix ~\ref{app:other-models} contains a table illustrating this, as well a more detailed exposition of the results with masked LMs.

\begin{figure}[t]
    \centering
    \includegraphics[width=0.7\linewidth,trim={0 0 0 .7cm},clip]{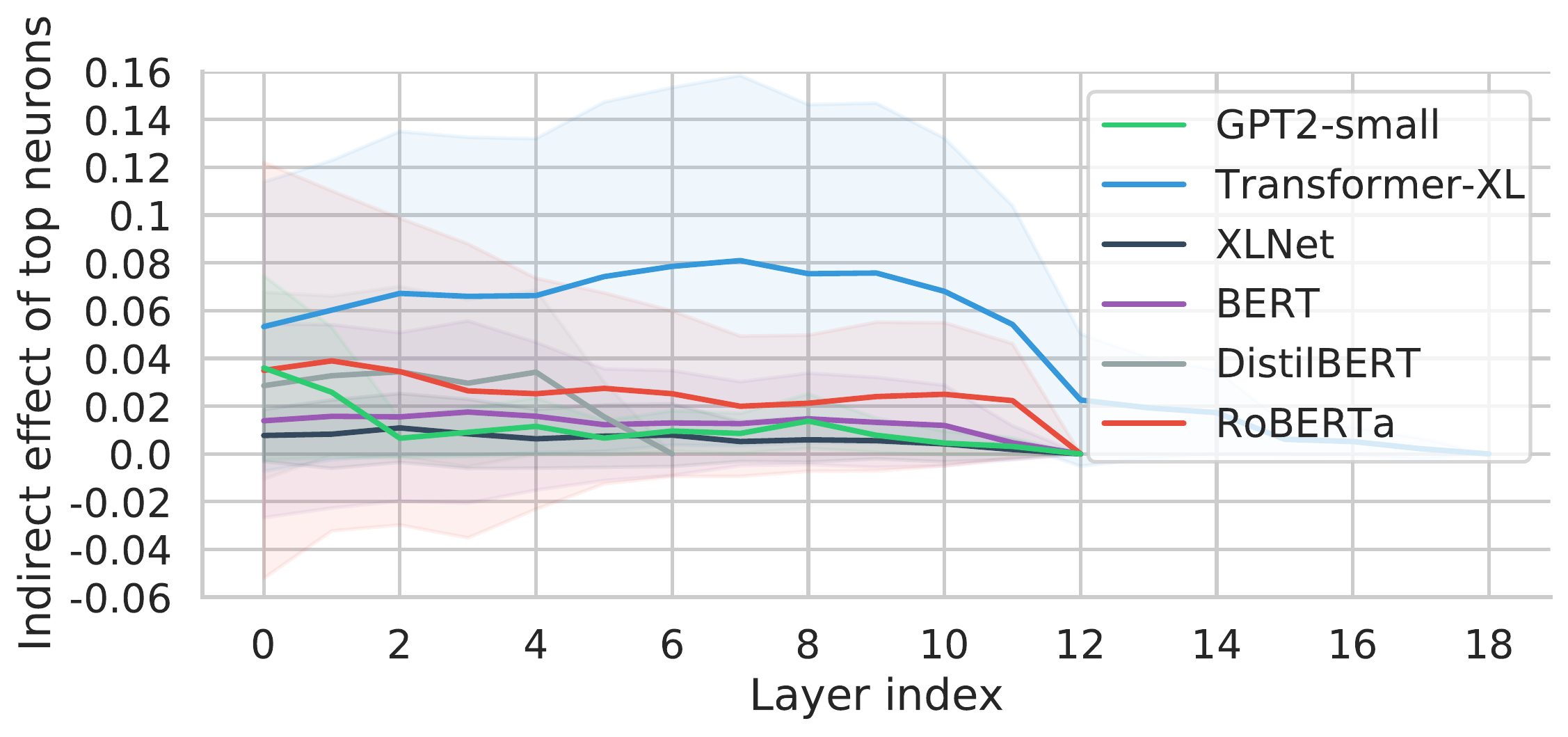}
    \caption{Indirect effects of top neurons in different models on the professions dataset.}
    \label{fig:other-neuron}
\end{figure}

\paragraph{Neuron interventions}
Figure~\ref{fig:other-neuron} shows the indirect effects from the top 5\% of neurons from each layer in all of our five additional models along with the same plot for GPT2-small. None of the non-GPT2 models seem to share the exact same pattern that we have observed with GPT2 variants, where the strongest effects in the word embedding layer used to be followed by a rapid drop in the successive layers. Here, a general trend seems to be one of roughly gradual decrease to 0 in the strength of effects, albeit with more and less significant fluctuations, the only exception being Transformer-XL whose effects are steadily increasing in the first third of the layers. As can be seen from Figure~\ref{fig:other-neuron}, we also observe considerably higher variances with some of the new models than in the case of GPT2. We do not have a compelling theory to explain these differences in results.

\section{Discussion and Conclusion}

This article introduced a structural-behavioral framework for interpreting neural models based on causal mediation analysis. An application of this framework to gender bias in NLP yielded several insights regarding the mechanisms by which gender bias is mediated in large Transformer LMs, 
revealing that gender bias effects are sparse, synergistic, and decomposable to direct and indirect effects.

This work can be extended in multiple ways. 
While the framework may apply to any property expressed as a function of model predictions, our experimental design focuses on gender bias in a binary setup. Preliminary experiments with gender-neutral references showed that models have not learned the concept of these references, at least for our chosen templates. Applying the methodology to more gender-inclusive setups and other kinds of biases is thus especially important since it can uncover these shortcomings in models.  

Our results can also guide model selection for limiting the amount of bias. 
In related work, \citeA{giulianelli2018hood} demonstrated that changing hidden representations in an LSTM based on the output of diagnostic classifiers can decrease the error on a subject-verb agreement classification task. 
Inspired by these results, mediation analysis could be used to determine where and how to intervene in similar ways and thus not only assess the success of debiasing techniques, but also motivate new debiasing methods that establish counterfactual fairness for protected groups. 
Another extension of this work for training a fair model would be to use the individual fairness framework \cite{dwork2011fairness} that inspired some of the alternate metrics discussed in this article. This approach would involve maximizing a traditional objective function through the typical training process subject to an additional Lipschitz constraint.

This work is a first attempt to adopt mediation analysis for interpreting NLP models. The causality literature often focuses on assumptions needed for identification of mediation effects 
from observed data \cite{Pearl:2001:DIE:2074022.2074073,avin2005identifiability,imai2010identification}. The challenge with inferring causality from observational data is that for each unit, the outcome is observed 
under a single intervention.
However, in this work we use the language of causal mediation to study the structure of NLP models, utilizing the fact that the outcome of the same unit (e.g., a sentence) can be observed under any intervention given a trained model. As a result, causal effects can be computed in a relatively simple manner. While we observed consistent results under multiple metrics, our definitions of causal effects to quantify bias could be refined, and alternative definitions might be advantageous for NLP research.

The causality literature offers many avenues for continuing this line of work, including  mediation analysis with non-linear models, and alternative effect decompositions ~ \cite{imai2010general,imai2010identification,vanderweele2009conceptual}. A 
promising direction is to focus on
path-specific effects \cite{avin2005identifiability}, to identify the exact mechanisms through 
which biases 
arise. Characterizing specific paths from model input to output might also be useful during training by disincentivizing the creation of paths leading to bias.   We believe the present work sets the ground for employing this literature in the ongoing effort to analyze neural NLP models. 

\section*{Acknowledgments}
S. G.\ was supported by a Siebel Fellowship. Y.B.\ was supported by the Harvard Mind, Brain, and Behavior Initiative. Work conducted while J.V. was at Palo Alto Research Center.

\appendix

\section{Data Preparation}

\subsection{Professions Dataset}
\label{app:templates}

Figure~\ref{fig:temps} shows the 17 base templates used for the neuron interventions. To validate that each template would capture gender bias, we instantiate each with an occupation of \textit{woman} and \textit{man} and verify that the conditional probabilities of \textit{she} and \textit{he} align with gender. Given \textit{woman} as the occupation word, the probability ratio $p(\text{she}) /p(\text{he})$ ranges from $2.5$ to $45.1$ across templates ($\mu=17.2, \sigma=13.1$). Given \textit{man}, the value $p(\text{he}) /  p(\text{she})$ ranges from $3.0$ to $55.4$ ($\mu=21.9, \sigma=16.2$). Thus the relative probabilities align with gender, though they vary greatly in magnitude.

For each of the templates, we used the following professions. Words in \emph{italics} are definitional and were thus excluded from the total effect calculation:

{\small \noindent \textbf{female:} \emph{actress}, advocate, aide, artist, baker, clerk, counselor, dancer, educator, instructor, maid, \emph{nun}, nurse, observer, performer, photographer, planner, poet, protester, psychiatrist, secretary, singer, substitute, teacher, teenager, therapist, treasurer, tutor, \emph{waitress}} \\
{\small \textbf{neutral:} acquaintance, character, citizen, correspondent, employee, musician, novelist, psychologist, student, writer}\\
{\small \textbf{male:} accountant, \emph{actor}, administrator, adventurer, ambassador, analyst, architect, assassin, astronaut, astronomer, athlete, attorney, author, banker, bartender, biologist, bishop, boss, boxer, broadcaster, broker, \emph{businessman}, butcher, campaigner, captain, chancellor, chef, chemist, cleric, coach, collector, colonel, columnist, comedian, comic, commander, commentator, commissioner, composer, conductor, congressman, consultant, cop, critic, curator, \emph{dad}, dean, dentist, deputy, detective, diplomat, director, doctor, drummer, economist, editor, entrepreneur, envoy, farmer, filmmaker, firefighter, \emph{fisherman}, footballer, goalkeeper, guitarist, historian, inspector, inventor, investigator, journalist, judge, landlord, lawmaker, lawyer, lecturer, legislator, lieutenant, magician, magistrate, manager, mathematician, mechanic, medic, midfielder, minister, missionary, \emph{monk}, narrator, negotiator, officer, painter, pastor, philosopher, physician, physicist, \emph{policeman}, politician, preacher, president, priest, principal, prisoner, professor, programmer, promoter, prosecutor, protagonist, rabbi, ranger, researcher, sailor, saint, \emph{salesman}, scholar, scientist, senator, sergeant, servant, soldier, solicitor, strategist, superintendent, surgeon, technician, trader, trooper, \emph{waiter}, warrior, worker, wrestler}

\begin{figure}[t]
    {\small The $<$occupation$>$ said that \ldots\\
The $<$occupation$>$ yelled that \ldots\\
The $<$occupation$>$ whispered that \ldots\\
The $<$occupation$>$ wanted that \ldots\\
The $<$occupation$>$ desired that \ldots\\
The $<$occupation$>$ wished that \ldots\\
The $<$occupation$>$ ate because \ldots\\
The $<$occupation$>$ ran because \ldots\\
The $<$occupation$>$ drove because \ldots\\
The $<$occupation$>$ slept because \ldots\\
The $<$occupation$>$ cried because \ldots\\
The $<$occupation$>$ laughed because \ldots\\
The $<$occupation$>$ went home because \ldots\\
The $<$occupation$>$ stayed up because \ldots\\
The $<$occupation$>$ was fired because \ldots\\
The $<$occupation$>$ was promoted because \ldots\\
The $<$occupation$>$ yelled because \ldots}
    \caption{Templates for neuron interventions. }
    \label{fig:temps}
\end{figure}

\begin{table*}[t]
\centering
\begin{tabular}{l r r r r r r r r}
\toprule 
& \multicolumn{4}{c}{Winobias} & \multicolumn{4}{c}{Winogender} \\ 
\cmidrule(lr){2-5} \cmidrule(lr){6-9}
& \multicolumn{2}{c}{Dev} & \multicolumn{2}{c}{Test} & \multicolumn{2}{c}{BLS} & \multicolumn{2}{c}{Bergsma} \\ 
\cmidrule(lr){2-3} \cmidrule(lr){4-5} \cmidrule(lr){6-7} \cmidrule(lr){8-9}
Model & Filt. & Unfilt. & Filt. & Unfilt. & Filt. & Unfilt. & Filt. & Unfilt. \\  
\midrule 
GPT2-distil & 61 & 160 & 51 & 130 & 15 & 44 & 18 & 44 \\
GPT2-small & 87 & 160 & 66 & 130 & 21 & 44 & 20 & 44 \\
GPT2-medium & 99 & 160 & 79 & 130 & 23 & 44 & 27 & 44 \\
GPT2-large & 94 & 160 & 69 & 130 & 24 & 44 & 26 & 44 \\
GPT2-xl & 101 & 160 & 72 & 130 & 25 & 44 & 26 & 44 \\
\bottomrule 
\end{tabular}
\caption{Number of examples from Winobias and Winogender datasets, including filtered (Filt.) and unfiltered (Unfilt.) versions.  The size of the filtered versions vary between models because each model produces different total effects (used for the filtering).   The number of examples excluded due to format (not included in the above numbers) were 38, 68, and 16 for Winobias Dev, Winobias Test, and Winogender, respectively.}
\label{tab:results-num-examples-all} 
\end{table*}

\begin{table*}[b]
\centering
\begin{tabular}{l r r r r r r r r}
\toprule 
& \multicolumn{4}{c}{Winobias} & \multicolumn{4}{c}{Winogender} \\ 
\cmidrule(lr){2-5} \cmidrule(lr){6-9}
& \multicolumn{2}{c}{Dev} & \multicolumn{2}{c}{Test} & \multicolumn{2}{c}{BLS} & \multicolumn{2}{c}{Bergsma} \\ 
\cmidrule(lr){2-3} \cmidrule(lr){4-5} \cmidrule(lr){6-7} \cmidrule(lr){8-9}
Model & Filt. & Unfilt. & Filt. & Unfilt. & Filt. & Unfilt. & Filt. & Unfilt. \\  
\midrule 
GPT2-distil & 0.118 & 0.012 & 0.127 & 0.023 & 0.081 & 0.005 & 0.075 & 0.011 \\
GPT2-small & 0.249 & 0.115 & 0.225 & 0.098 & 0.103 & 0.020 & 0.135 & 0.040 \\
GPT2-medium & 0.774 & 0.474 & 0.514 & 0.311 & 0.322 & 0.128 & 0.384 & 0.231 \\
GPT2-large & 0.751 & 0.427 & 0.492 & 0.238 & 0.364 & 0.173 & 0.350 & 0.192 \\
GPT2-xl & 1.049 & 0.660 & 0.754 & 0.400 & 0.342 & 0.168 & 0.362 & 0.202 \\
\bottomrule 
\end{tabular}
\caption{Total effects on Winobias and Winogender, including filtered (Filt.) and unfiltered (Unfilt.) versions.}
\vspace{-.5em}
\label{tab:results-attention-te-all} 
\end{table*}

\subsection{Winobias and Winogender}
\label{app:data}

For both Winobias and Winogender datasets, we exclude templates in which the shared prompt does not end in a pronoun.\footnote{An example of a removed template is: ``The receptionist welcomed the lawyer because \textit{this is part of her job.''} / ``The receptionist welcomed the lawyer because \textit{it is his first day to work.}''} 
For Winobias, we only consider \textit{Type 1} examples, which follow the format of a shared prompt and two alternate continuations.  
We also experiment with filtering by total effect, removing examples with a negative total effect as well as examples in the bottom quartile of those with a positive total effect. The sizes of all dataset variations may be found in Table~\ref{tab:results-num-examples-all}. Results are reported for filtered versions of both datasets and the Dev set of Winobias unless otherwise noted.

Both datasets include statistics from the U.S. Bureau of Labor Statistics (BLS) 
to assess the gender stereotypicality of the referenced occupations. Winogender additionally
includes gender estimates from text \cite{bergsma-lin-2006-bootstrapping}, which we also include in our analysis.
Whereas each Winobias example includes two occupations of opposite stereotypicality, each Winogender example includes one occupation and a \textit{participant}, for which no gender statistics are provided. For consistency with the Winobias analysis, we make the simplifying assumption that the gender stereotypicality of the participant is the opposite of that of the occupation.

\section{Additional Total Effects}
Table \ref{tab:results-attention-te-all} provides the total effects across all variations of the Winograd-style datasets. The relationship between model and effect size is relatively consistent across dataset variations (Winobias/Winogender, filtered/unfiltered, Dev/Test, BLS/Bergsma gender statistics), though the magnitudes of the effects may vary between dataset variations.

Table~\ref{tab:results-neuron-te-all} provides the total effects on the professions dataset when separated to  stereotypically female and male professions, where stereotypicality is defined by the profession statistics provided by \citeA{NIPS2016_6228}.  Notably, the effects are much larger in the female case. This may be explained by stereotypicaly-female professions being of higher stereotypicality than stereotypically-male professions, reflecting a societal bias viewing women's professions as more narrowed. 

\begin{table}[t]
\sisetup{round-mode=places}
\centering
\begin{tabular}{l H H H H H H S[round-precision=2] S[round-precision=2] S[round-precision=2]}
\toprule
Model & Params & Layers & Heads & Hidden & Winobias & Winogender &  \text{Female}  & \text{Male}  & All  \\ 
\midrule 
GPT2-small rand. & 117M & 12 & 12 & 768 & 0.066 & 0.045 & 0.101 & 0.191 & 0.117 \\ 
\midrule 
GPT2-distil & 82M & 6 & 12 & 768 & 0.118 & 0.081 & 155.306 & 23.467 & 130.859 \\ 
GPT2-small & 117M & 12 & 12 & 768 & 0.249 & 0.103 & 129.363 & 15.157 & 112.275 \\ 
GPT2-medium & 345M & 24 & 16 & 1024 & 0.774 & 0.322 & 120.603 & 94.751 & 115.945 \\  
GPT2-large & 774M & 36 & 20 & 1280 & 0.751  & 0.364 & 107.440 & 48.988 & 96.859 \\  
GPT2-xl & 1558 M & 48 & 25 & 1600 & 1.049 & 0.342 & 255.219 & 89.309 & 225.217 \\  
\bottomrule 
\end{tabular}
\caption{Total effects (TE) of gender bias in various GPT2 variants evaluated on the professions dataset, when separating by gender-stereotypicality. 
}
\vspace{-.7em}
\label{tab:results-neuron-te-all}
\end{table}

\section{Additional Attention Results}
\subsection{Indirect and Direct Effects}
\label{app:heatmaps}
\begin{figure*}[t]
\centering
\begin{subfigure}[t]{.45\textwidth}
    \includegraphics[width=1\linewidth]{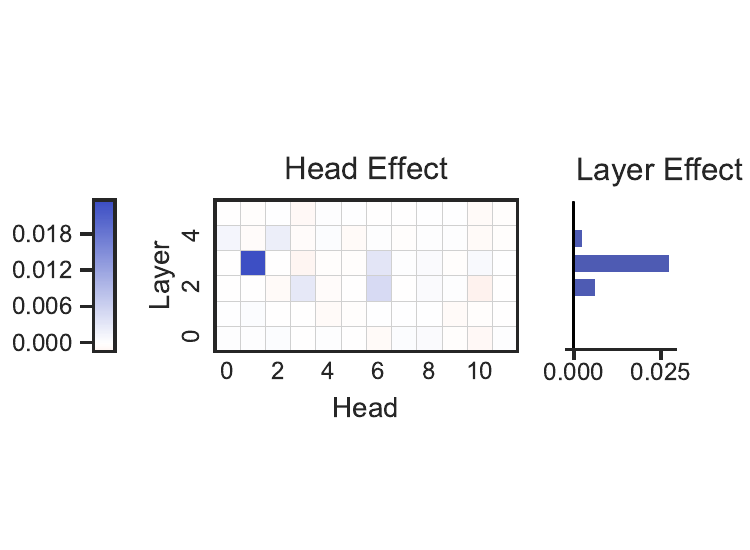}
\caption{GPT2-distil}
\vspace{1em}

\end{subfigure}\hfill
\begin{subfigure}[t]{.45\textwidth}
    \centering
    \includegraphics[width=1\linewidth]{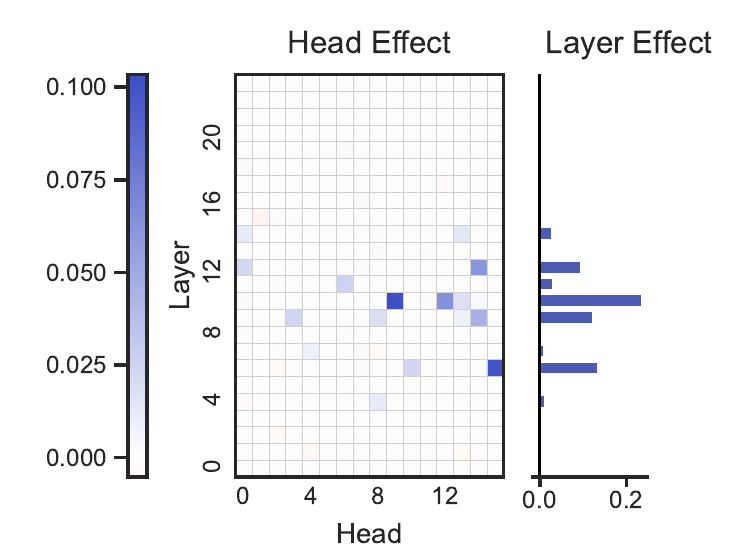}
\caption{GPT2-medium }
\vspace{1em}

\end{subfigure}
\begin{subfigure}[t]{.45\textwidth}
    \centering
    \includegraphics[width=1\linewidth]{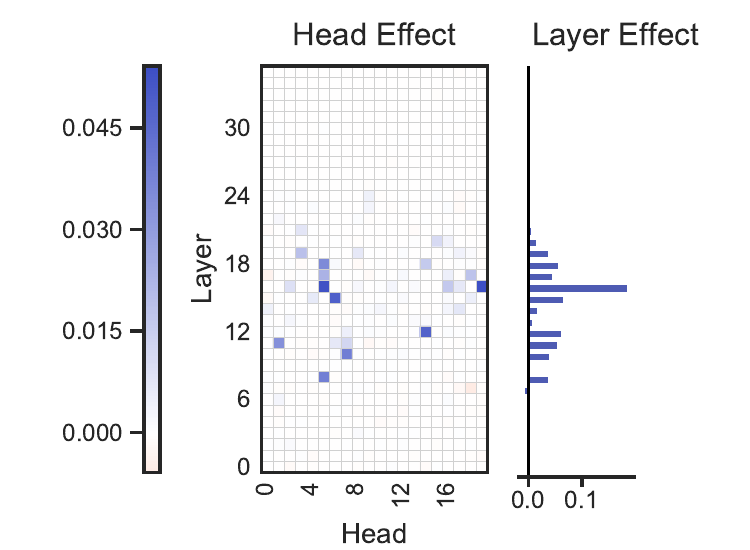}
\caption{GPT2-large}
\end{subfigure}\hfill
\begin{subfigure}[t]{.45\textwidth}
    \centering
    \includegraphics[width=1\linewidth]{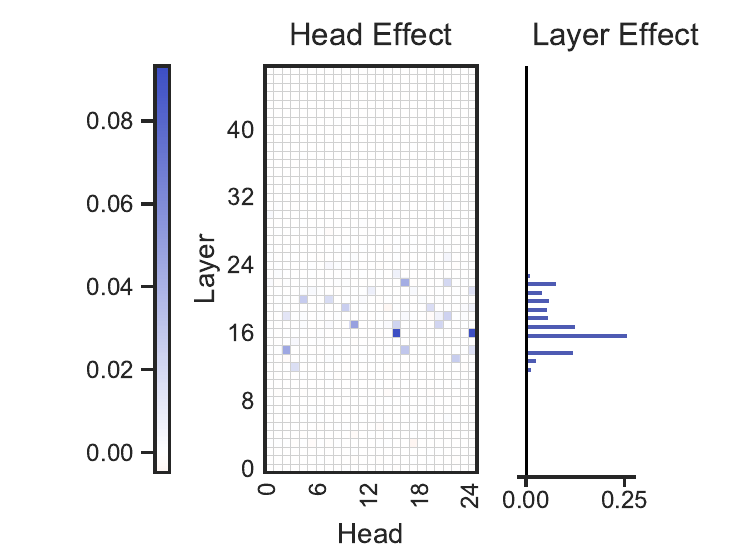}
\caption{GPT2-xl}
\end{subfigure}
\caption{Mean indirect effect on Winobias for heads (the heatmap) and layers (the bar chart) over additional GPT2 variants.}
\label{fig:indirect_heatmap_by_model}
\end{figure*}

\begin{figure}[t]
\centering
    \includegraphics[width=0.6\columnwidth]{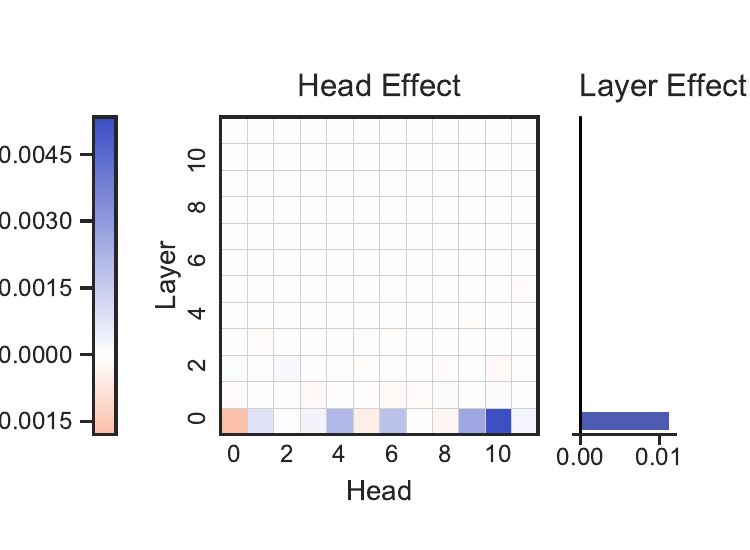}
    \caption{Indirect effect when using a randomly initialized GPT2-small model on Winobias.}
    \label{fig:null_heatmap}
\end{figure}

\begin{figure*}[t]
\centering
\begin{subfigure}[t]{.44\textwidth}
\centering
\includegraphics[width=\textwidth]{images/winobias_gpt2_filtered_dev.pdf}
\vspace{-2em}
\caption{Filtered, Dev}
\end{subfigure}\hfill
\begin{subfigure}[t]{.44\textwidth}
\centering
\includegraphics[width=\textwidth]{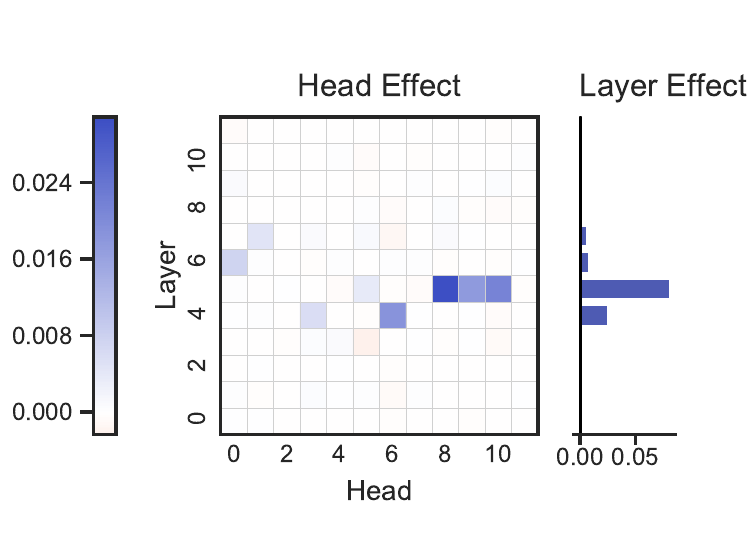}
\vspace{-2em}
\caption{Unfiltered, Dev}
\end{subfigure}
\begin{subfigure}[t]{.44\textwidth}
\includegraphics[width=1\textwidth]{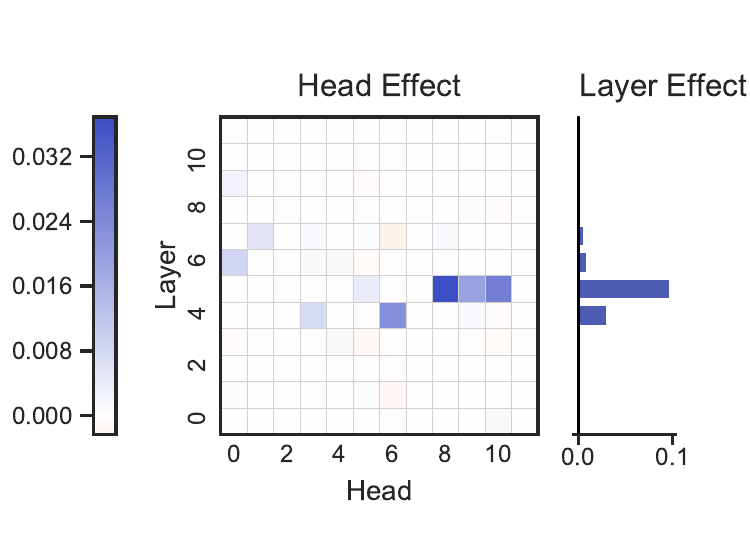}
\vspace{-2em}
\caption{Filtered, Test}
\end{subfigure}\hfill
\begin{subfigure}[t]{.44\textwidth}
\includegraphics[width=1\textwidth]{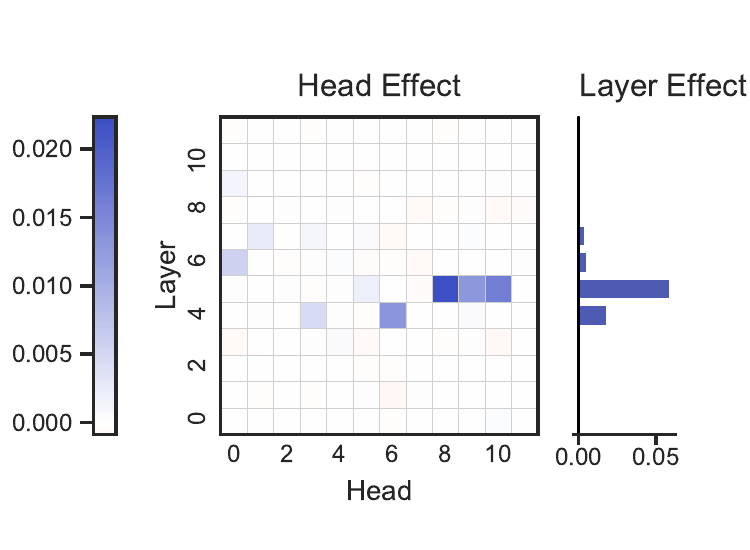}
\vspace{-2em}
\caption{Unfiltered, Test}
\end{subfigure}
\caption{Indirect effect for Winobias (GPT2-small).}
\label{fig:winobias_heatmaps}
\end{figure*}

\begin{figure*}[t]
\centering
\begin{subfigure}[t]{.44\textwidth}
\centering
\includegraphics[width=\textwidth]{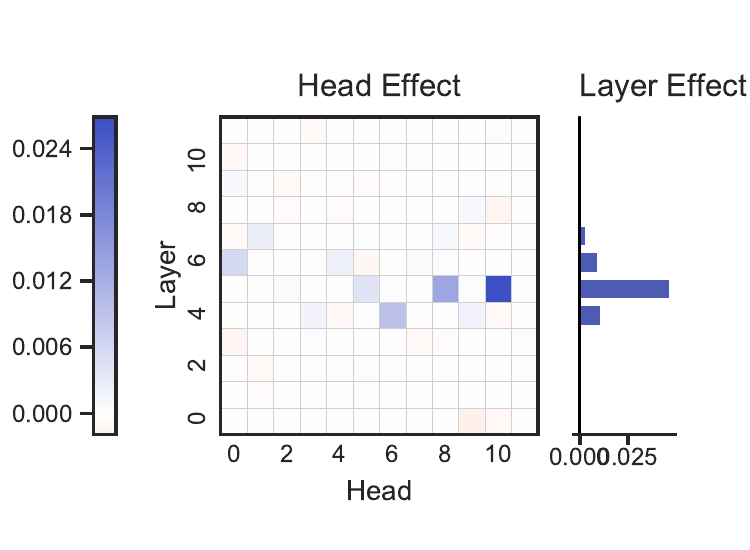}
\vspace{-2em}
\caption{Filtered, BLS}
\end{subfigure}\hfill
\begin{subfigure}[t]{.44\textwidth}
\centering
\includegraphics[width=\textwidth]{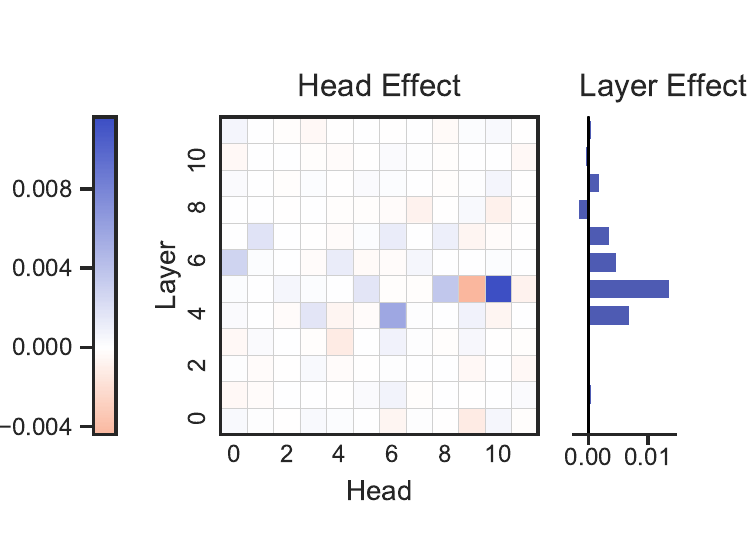}
\vspace{-2em}
\caption{Unfiltered, BLS}
\end{subfigure}
\begin{subfigure}[t]{.44\textwidth}
\includegraphics[width=1\textwidth]{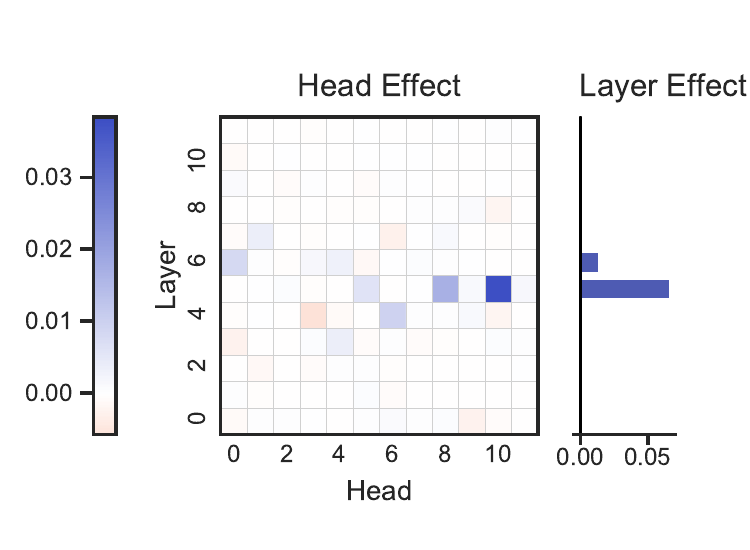}
\vspace{-2em}
\caption{Filtered, Bergsma}
\end{subfigure}\hfill
\begin{subfigure}[t]{.44\textwidth}
\includegraphics[width=1\textwidth]{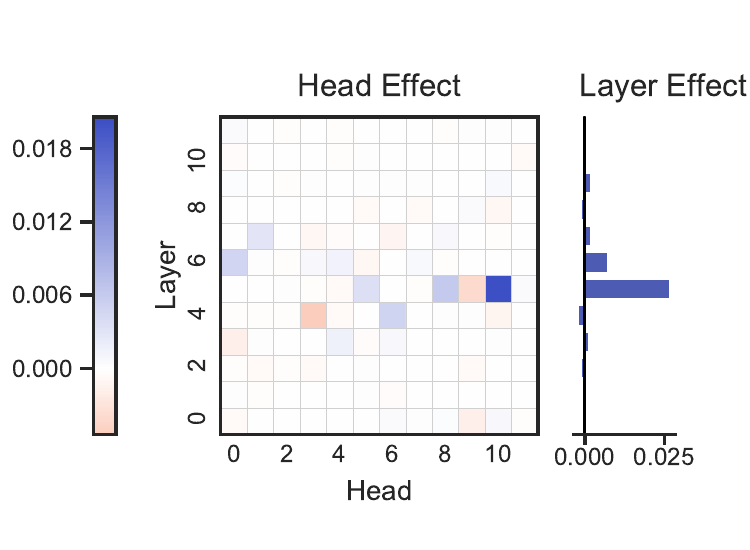}
\vspace{-2em}
\caption{Unfiltered, Bergsma}
\end{subfigure}
\caption{Indirect effect for Winogender (GPT2-small).}
\label{fig:winogender_heatmaps}
\end{figure*}

Figure~\ref{fig:indirect_heatmap_by_model} complements Figure~\ref{fig:results-attention-indirect-heatmap} by visualizing the indirect effects for additional GPT2 models.  As with Figure~\ref{fig:results-attention-indirect-heatmap}, the attention heads with the largest indirect effects lie in the middle layers of each model. Figure~\ref{fig:null_heatmap} shows the indirect effects for a model with randomized weights.
Figures \ref{fig:winobias_heatmaps} and \ref{fig:winogender_heatmaps} visualize the indirect effects for other dataset variations for the  GPT2-small model from Figure~\ref{fig:results-attention-indirect-heatmap}. The attention heads with largest indirect effect have significant overlap across the dataset variations. 

\begin{figure*}[t]
\centering
\begin{subfigure}[t]{.44\textwidth}
\centering
\includegraphics[width=\textwidth]{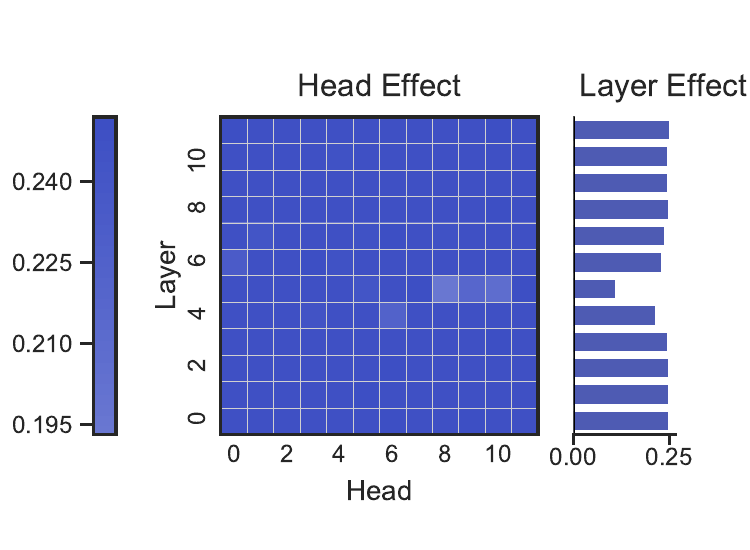}
\caption{GPT2-small}
\end{subfigure}\hfill
\begin{subfigure}[t]{.44\textwidth}
\centering
\includegraphics[width=\textwidth]{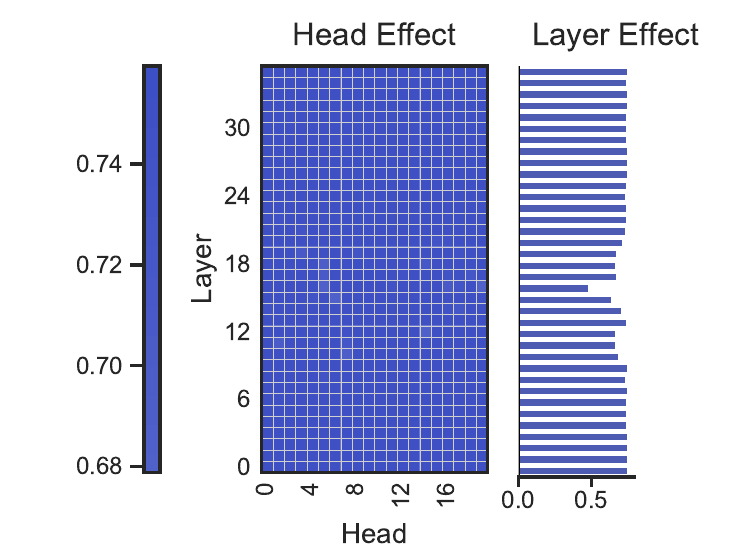}
\caption{GPT2-large}
\end{subfigure}
\caption{Direct effect for Winobias for GPT2-small and GPT2-large.}
\label{fig:winobias_heatmaps_direct}
\end{figure*}

Figure~\ref{fig:winobias_heatmaps_direct} visualizes \emph{direct} effects on Winobias for GPT2-small and GPT2-large. As discussed in Section~\ref{sec:decomposition}, the sum of direct and indirect effects approximate the total effect.

\subsection{Examples} \label{app:examples}
\begin{figure*}
    \centering
    \begin{minipage}[b]{.47\textwidth}
         \includegraphics[width=\linewidth,trim={.25cm .25cm 0.45cm 0.25},clip]{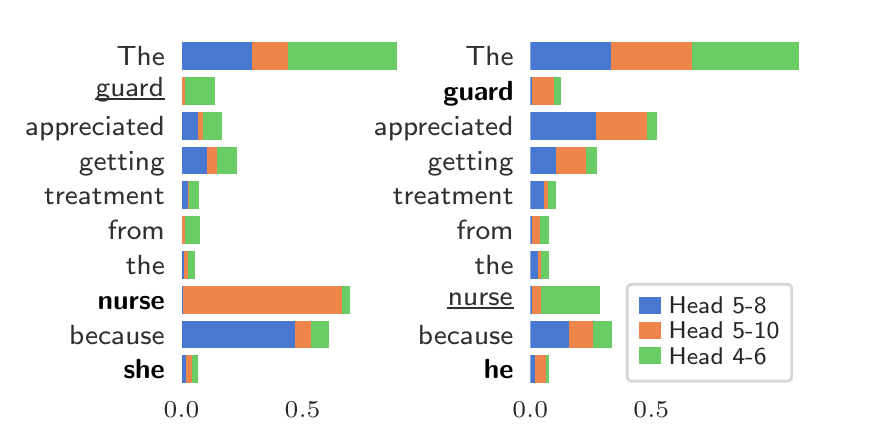}
    \end{minipage}\qquad
    \begin{minipage}[b]{.47\textwidth}
         \includegraphics[width=\linewidth,trim={.25cm .25cm 0.45cm 0.25},clip]{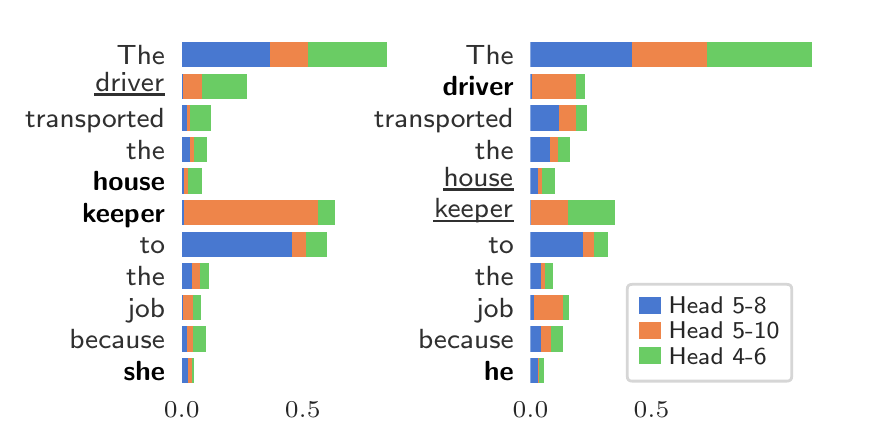}
    \end{minipage}\qquad
    \begin{minipage}[b]{.47\textwidth}
         \includegraphics[width=\linewidth,trim={.25cm .25cm 0.45cm 0.25},clip]{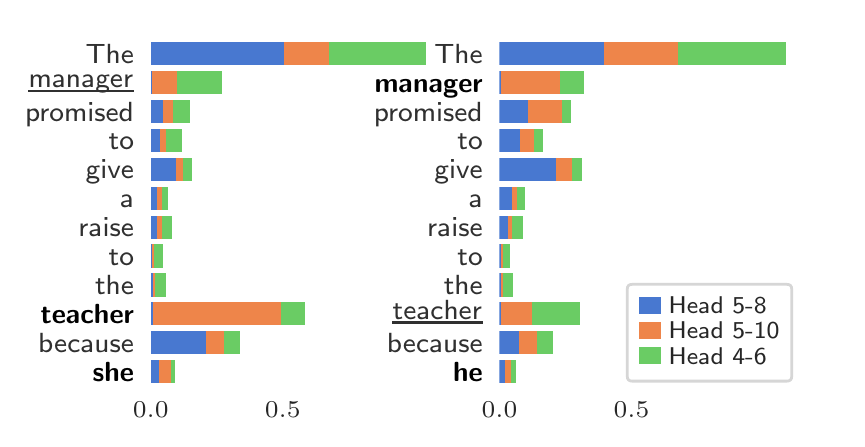}
    \end{minipage}\qquad
    \begin{minipage}[b]{.47\textwidth}
         \includegraphics[width=\linewidth,trim={.25cm .25cm 0.45cm 0.25},clip]{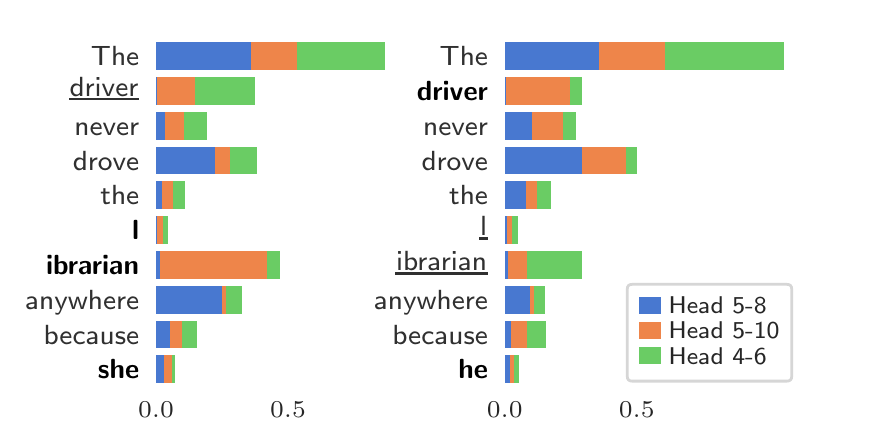}
    \end{minipage}\qquad
    \begin{minipage}[b]{.47\textwidth}
         \includegraphics[width=\linewidth,trim={.25cm .25cm 0.45cm 0.25},clip]{images/examples_winobias_gpt2_filtered_dev_4.pdf}
    \end{minipage}\qquad
    \begin{minipage}[b]{.47\textwidth}
         \includegraphics[width=\linewidth,trim={.25cm .25cm 0.45cm 0.25},clip]{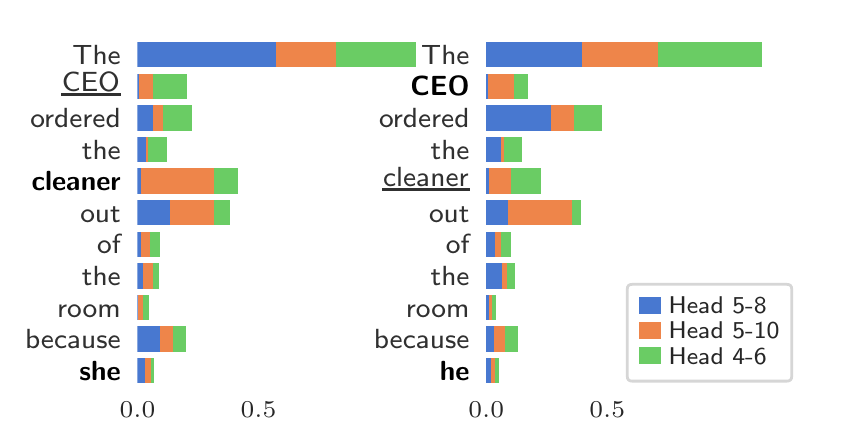}
    \end{minipage}\qquad
    \begin{minipage}[b]{.47\textwidth}
         \includegraphics[width=\linewidth,trim={.25cm .25cm 0.45cm 0.25},clip]{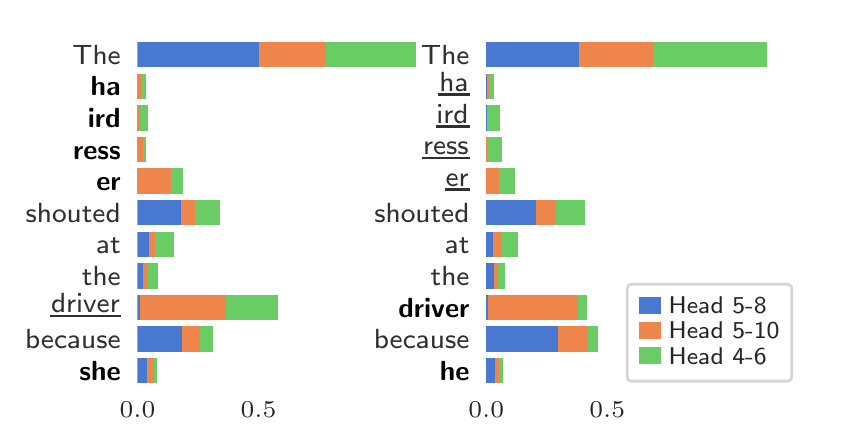}
    \end{minipage}\qquad
    \begin{minipage}[b]{.47\textwidth}
         \includegraphics[width=\linewidth,trim={.25cm .25cm 0.45cm 0.25},clip]{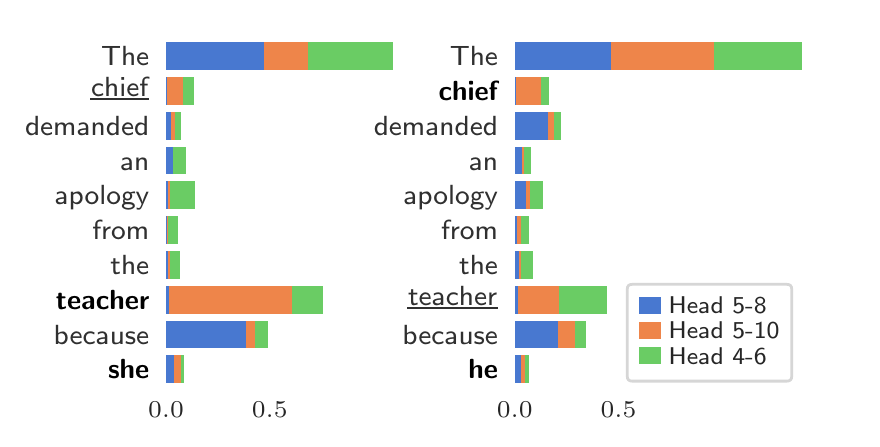}
    \end{minipage}\qquad
        \begin{minipage}[b]{.47\textwidth}
         \includegraphics[width=\linewidth]{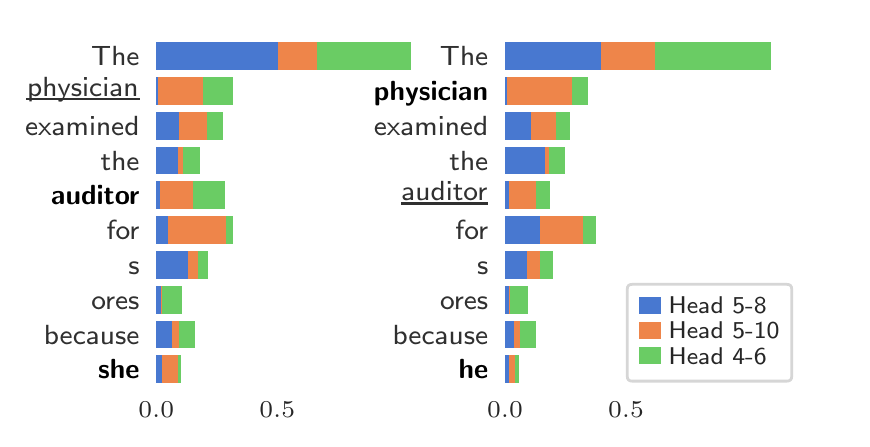}
    \end{minipage}\qquad
        \begin{minipage}[b]{.47\textwidth}
         \includegraphics[width=\linewidth]{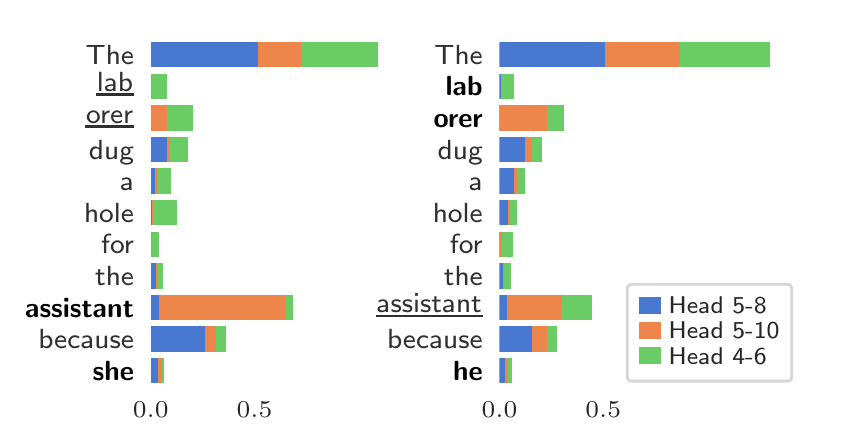}
    \end{minipage}\qquad
    \caption{Attention of different heads across the 10 Winobias examples with greatest total effect for the GPT2-small model. The stereotypical candidate is in \textbf{bold} and the anti-stereotypical candidate is \underline{underlined}. Attention roughly follows the pattern described in Figure \ref{fig:qualitative}.}
    \label{fig:qualitative-additional-examples}
    \vspace{-2em}
\end{figure*}

\begin{figure*}[t]
\centering
\begin{subfigure}[t]{.47\textwidth}
    \centering
    \includegraphics[width=\textwidth,trim={.25cm .25cm 0.45cm 0.25},clip]{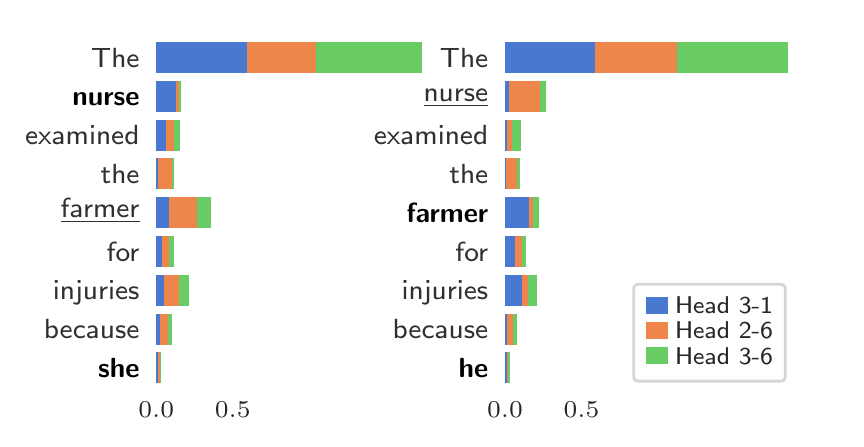}
    \caption{Attention for GPT2-distil. Most attention is directed to the first token (null attention). Head \textcolor{blue}{3-1} attends primarily to the \textbf{bold} stereotypical candidate, head \textcolor{orange}{2-6} attends to the \underline{underlined} anti-stereotypical candidate, and attention from head \textcolor{green}{3-6} is roughly evenly distributed.}
\end{subfigure}\hfill
\begin{subfigure}[t]{.47\textwidth}
   \centering
   \includegraphics[width=\textwidth,trim={.25cm .25cm 0.45cm 0.25},clip]{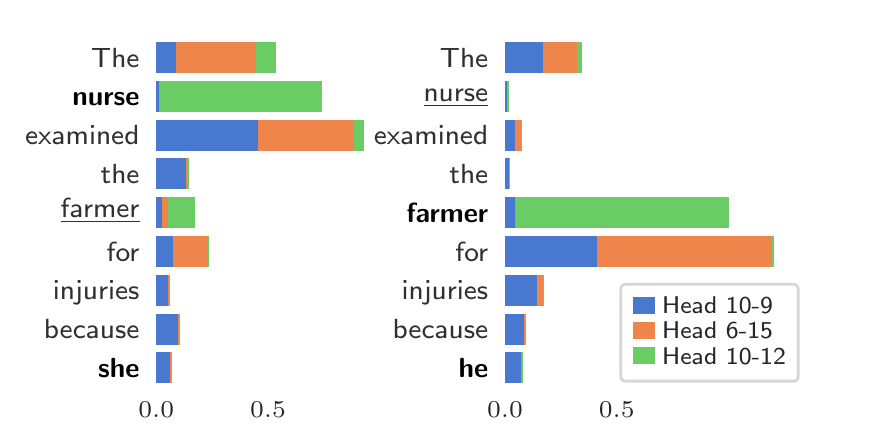}
    \caption{Attention for GPT2-medium. Head \textcolor{green}{10-12} attends directly to the \textbf{bold} stereotypical candidate, and heads \textcolor{blue}{10-9} and \textcolor{orange}{6-15} attend to the following words.}
\end{subfigure}
\begin{subfigure}[t]{.47\textwidth}
   \centering
   \includegraphics[width=\textwidth,trim={.25cm .25cm 0.45cm 0},clip]{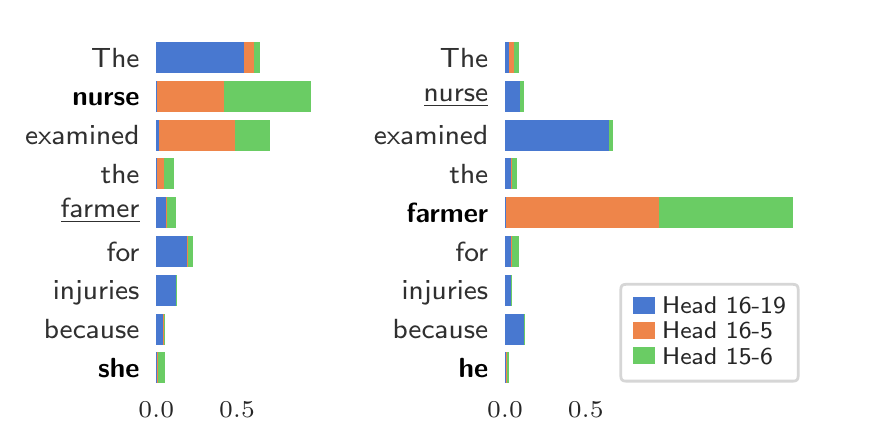}
    \caption{Attention for GPT2-large. Heads \textcolor{orange}{16-5} and \textcolor{green}{15-6} attend to the \textbf{bold} stereotypical candidate and optionally the following word. Head \textcolor{blue}{16-19} attends to the words following the \underline{underlined} anti-stereotypical candidate.}
\end{subfigure}\hfill
\begin{subfigure}[t]{.47\textwidth}
   \centering
   \includegraphics[width=\textwidth,trim={.25cm .25cm 0.45cm 0},clip]{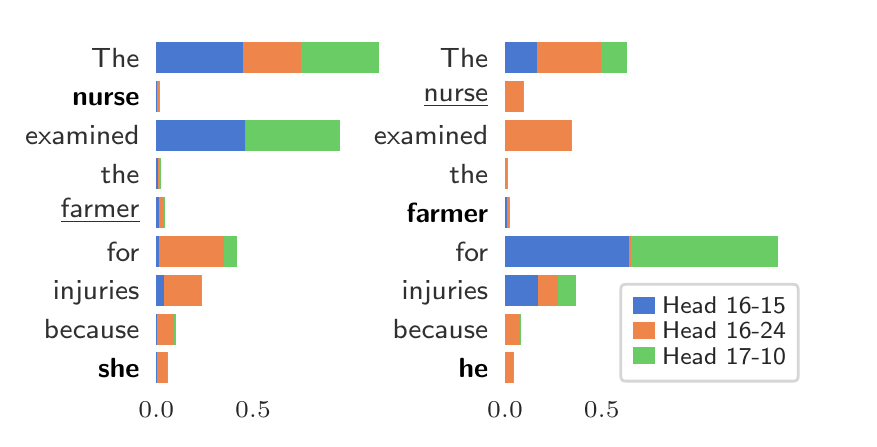}
    \caption{Attention for GPT2-xl. Heads \textcolor{blue}{16-5} and \textcolor{green}{17-10} attend primarily to the word following the \textbf{bold} stereotypical candidate. Head \textcolor{orange}{16-24} attends primarily to the words following the \underline{underlined} anti-stereotypical candidate.}
\end{subfigure}
\caption{Attention of top 3 heads on an example from Winobias, directed from either \textit{she} or \textit{he}, across different GPT2 models. The colors correspond to different heads. The results for GPT2-small are shown in Figure~\ref{fig:qualitative}.}
\label{fig:qualitative-additional-models}
\end{figure*}

Figure~\ref{fig:qualitative-additional-examples} visualizes attention for the Winobias examples with the greatest total effect in GPT2-small, complementing the example shown in Figure~\ref{fig:qualitative}. Figure~\ref{fig:qualitative-additional-models} visualizes attention for additional models for the same example shown in Figure~\ref{fig:qualitative}. 

\clearpage 

\section{Additional subset selection results} \label{app:subset}
We wish to select a subset of attention heads or neurons that perform well together to better understand the sparsity of attention heads and neurons and their impact on gender bias in Transformer models. 

The problem of subset selection (selecting $k$ elements from $n$) is an NP-hard combinatorial optimization problem. To construct a meaningful solution set, we employ several algorithms for subset selection from submodular maximization. We note that while our objective functions are not strictly submodular as they do not satisfy the diminishing returns property, our objectives exhibit submodular-like properties and numerous algorithms have been proposed to efficiently maximize submodular and variants of submodular functions. 		 	 	

For monotone submodular functions, it is known that a greedy algorithm that iteratively selects the element with the maximal marginal contribution to its current solution obtains a $1 - 1/e$ approximation for maximization under a cardinality constraint \cite{nemhauser1978best} and that this bound is optimal. For non-monotone submodular functions, there is the randomized greedy algorithm which emits a $1/e$ approximation to the optimal solution \cite{buchbinder2014submodular}. 

\begin{figure*}
    \centering
    \begin{subfigure}[b]{0.42\columnwidth}
       \includegraphics[width=\columnwidth]{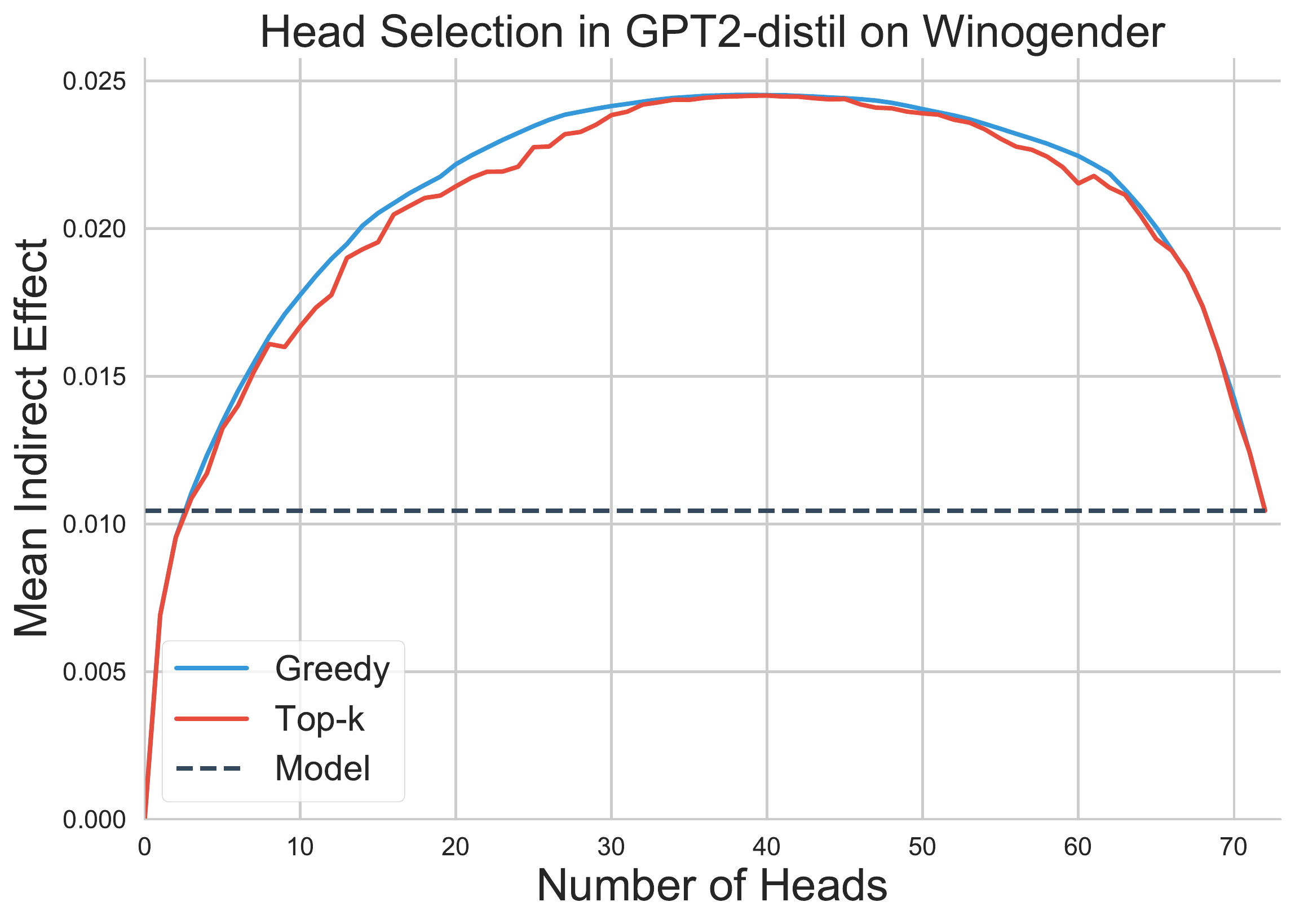}
    \end{subfigure}\hspace{1em}
    \begin{subfigure}[b]{0.42\columnwidth}
       \includegraphics[width=\columnwidth]{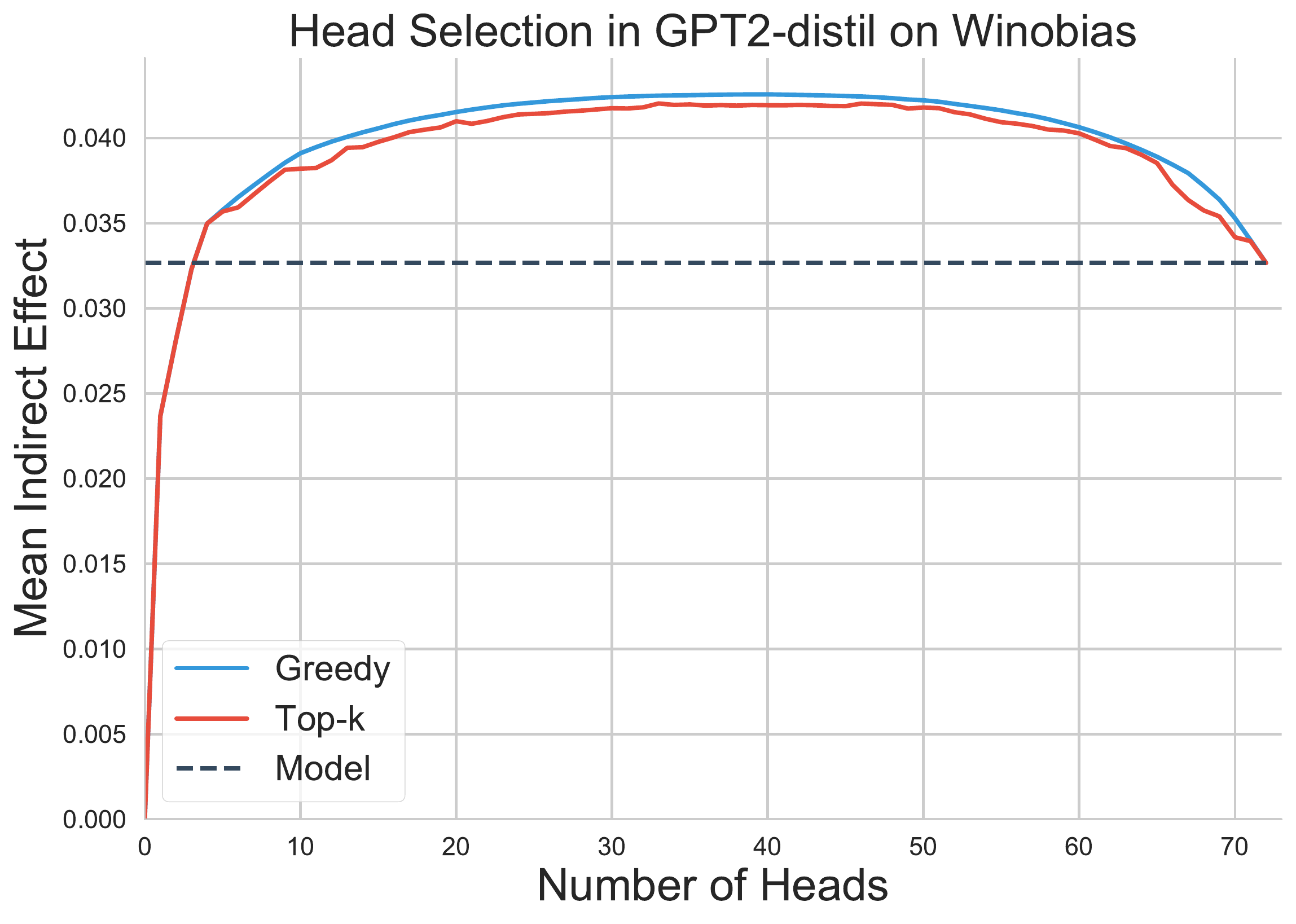}
    \end{subfigure}\\
    \begin{subfigure}[b]{0.42\columnwidth}
       \includegraphics[width=\columnwidth]{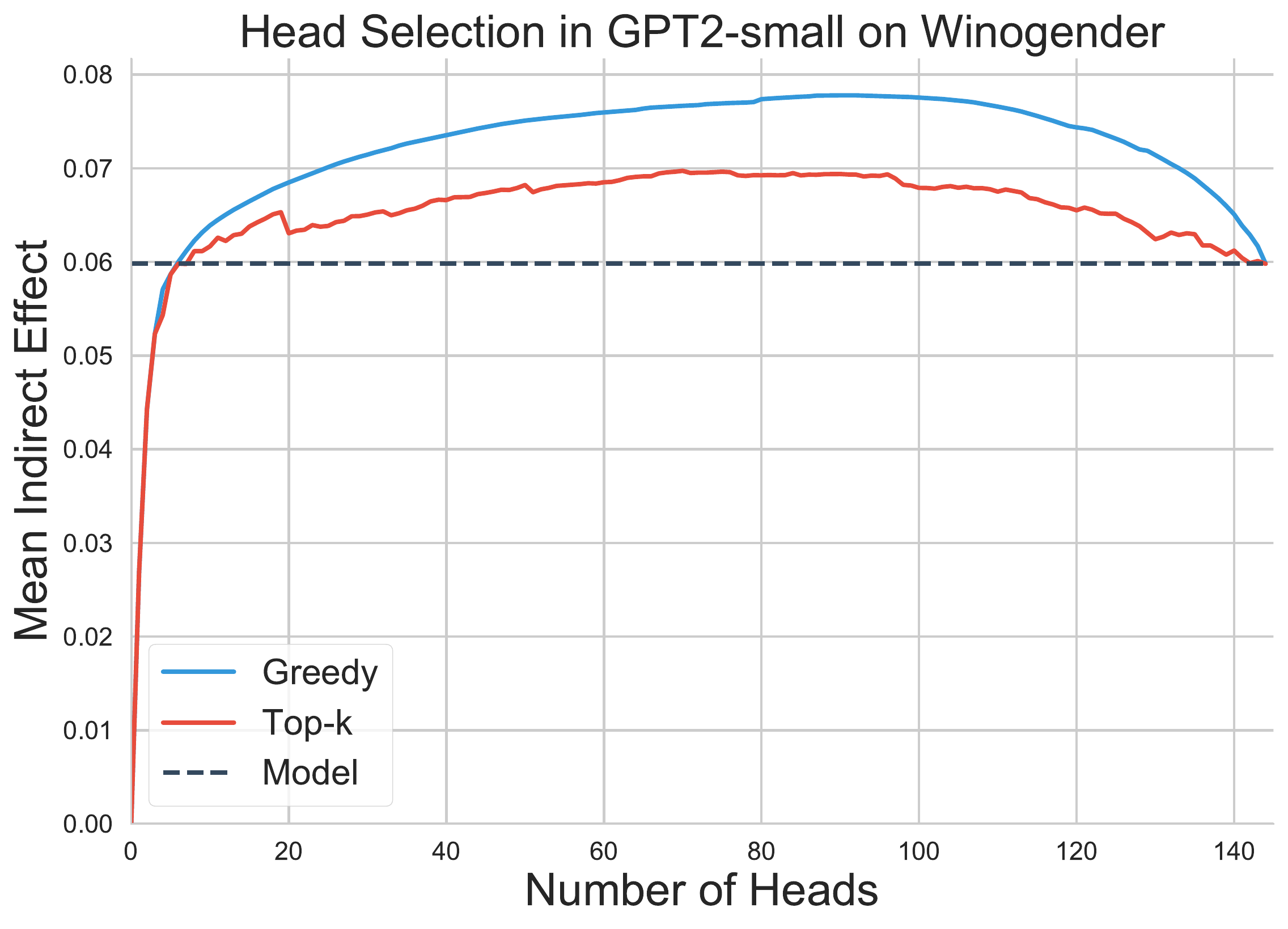}
    \end{subfigure}\hspace{1em}
    \begin{subfigure}[b]{0.42\columnwidth}
       \includegraphics[width=\columnwidth]{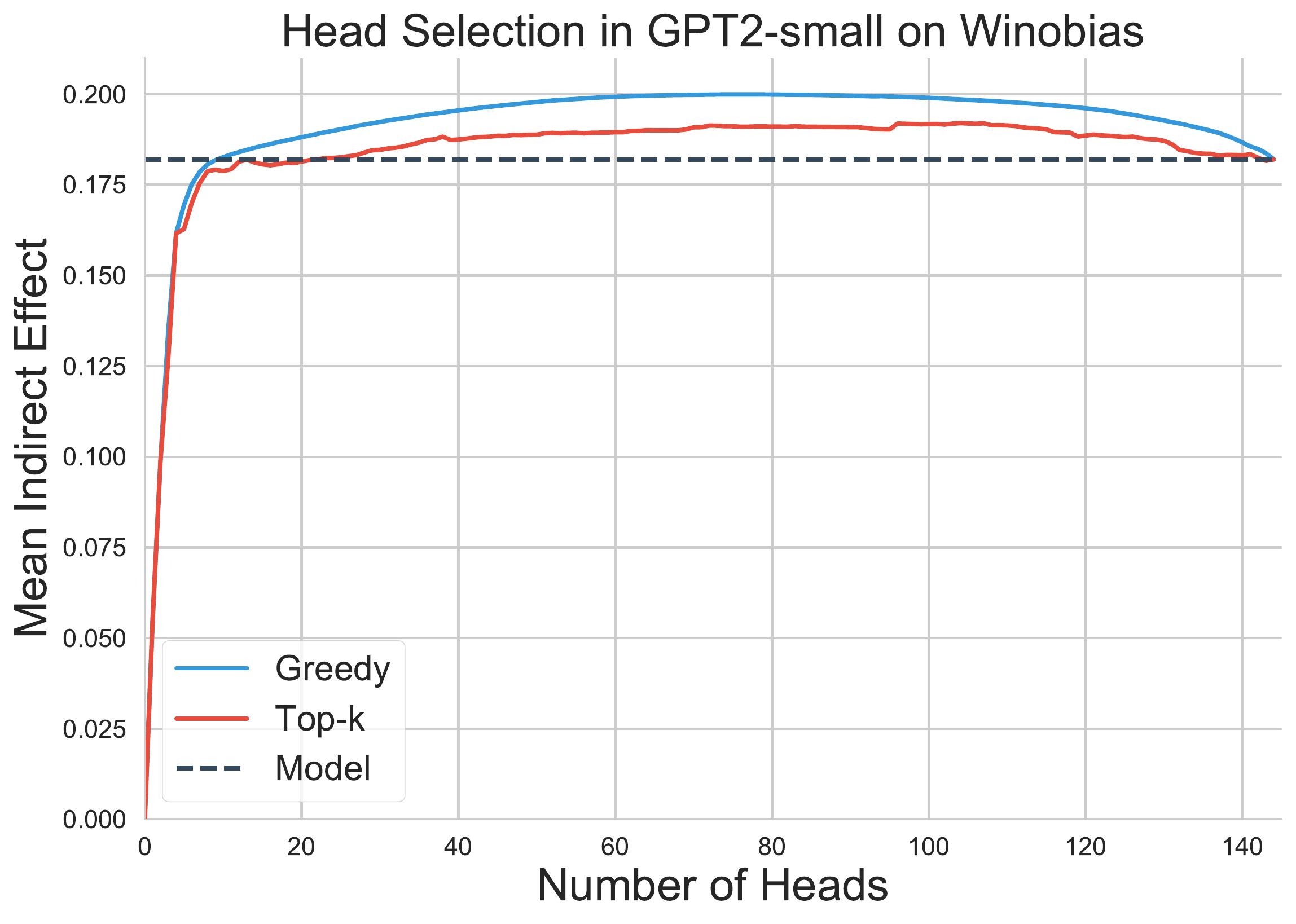}
    \end{subfigure}\\
    \begin{subfigure}[b]{0.42\columnwidth}
       \includegraphics[width=\columnwidth]{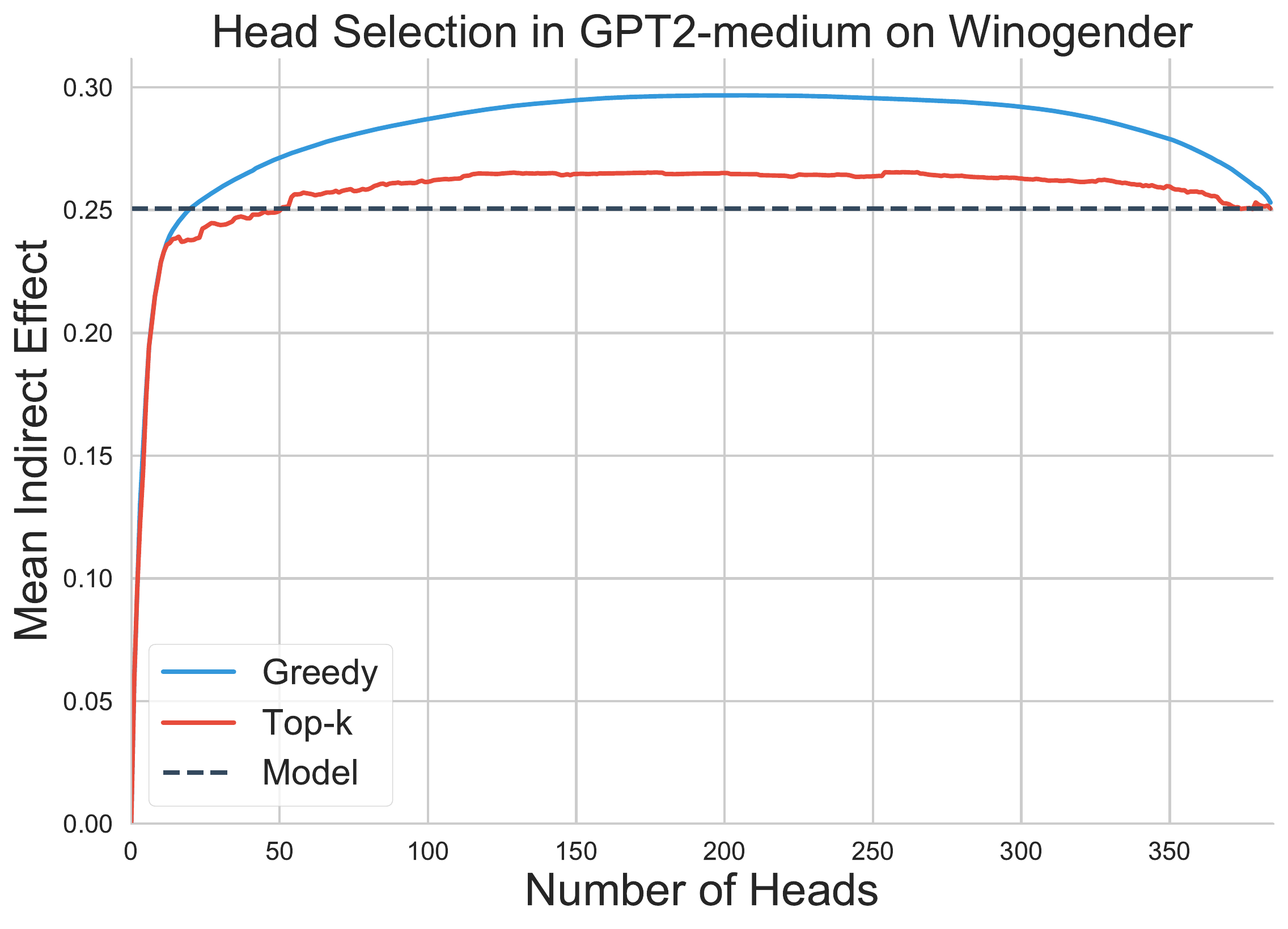}
    \end{subfigure}\hspace{1em}
    \begin{subfigure}[b]{0.42\columnwidth}
       \includegraphics[width=\columnwidth]{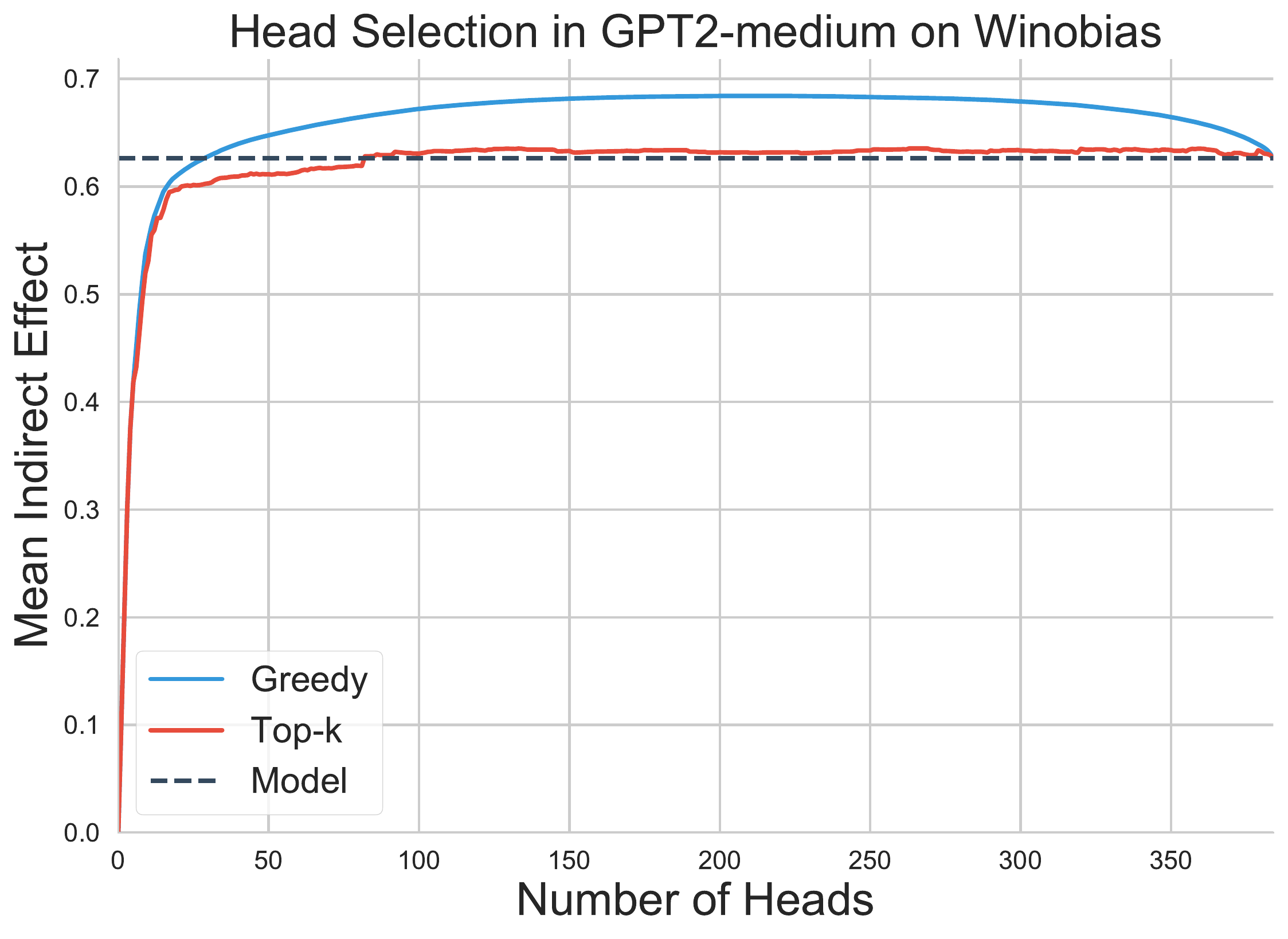}
    \end{subfigure}\\
    \begin{subfigure}[b]{0.42\columnwidth}
       \includegraphics[width=\columnwidth]{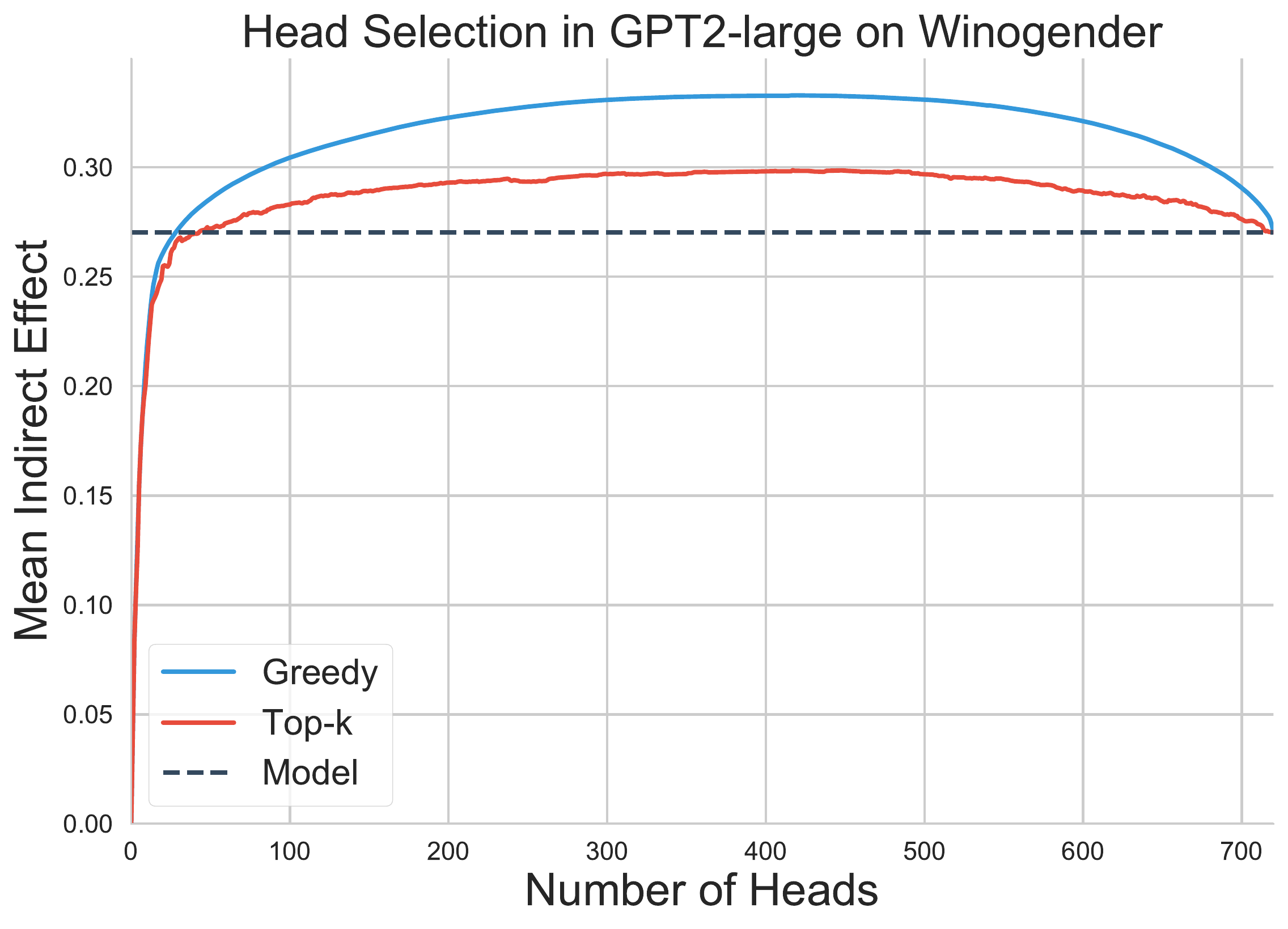}
    \end{subfigure}\hspace{1em}
    \begin{subfigure}[b]{0.42\columnwidth}
       \includegraphics[width=\columnwidth]{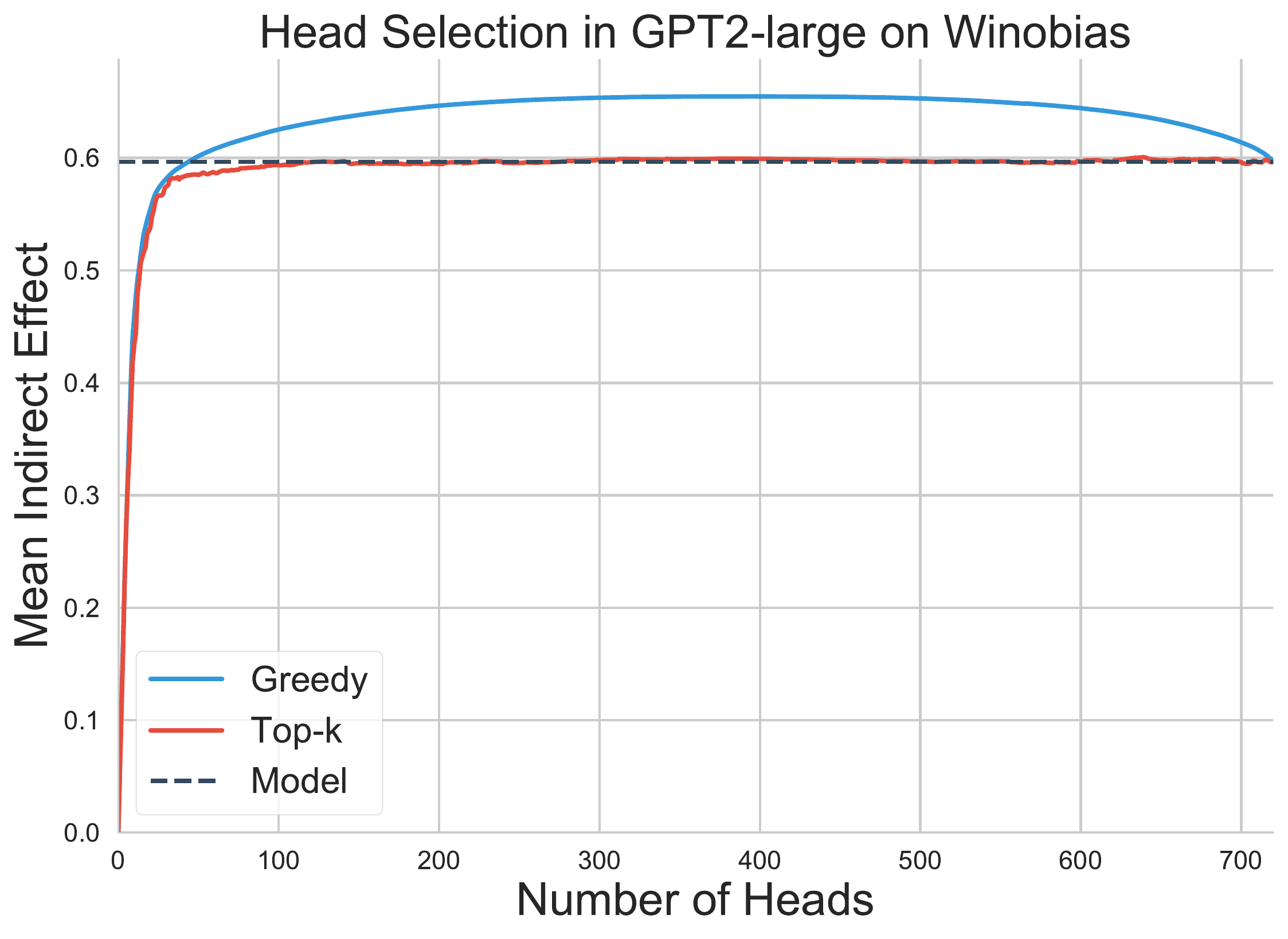}
    \end{subfigure}
    \caption{ The indirect effect after sequentially selecting an increasing number of heads through the \textsc{Top-k} or \textsc{Greedy} approach on different model types and data. A small proportion of heads are required to saturate the effect of the model.}
    \label{fig:head_selection}
\end{figure*}

To select subsets of attention heads, we compare \textsc{Top-k} (selecting $k$ elements with the largest individual values) and \textsc{Greedy}. Even though randomized greedy has stronger theoretical guarantees because our objective is clearly non-monotonic, we favor the deterministic algorithm for increased interpretability. Figure \ref{fig:head_selection} shows results for head selection across different models on Winogender and Winobias. Sparsity is consistent across all experiments where only a small proportion of heads are sufficient to achieve the full model effect of intervening at all heads. On Winogender, only 4/4/5/4\% of heads are needed to saturate, while on Winobias, only 6/7/8/6\% of heads are needed in GPT2-distil/small/medium/large.

To select subsets of neurons, we use \textsc{Top-k} to compute NIE of sets of neurons because sequential greedy is too computationally intensive to run. Alternative methods using adaptive sampling techniques have been proposed to speed-up \textsc{Greedy} for submodular functions under cardinality constraints \cite{ene2019,farbach2019,BS18a,balkanski2018}. For non-monotone or non-submodular functions, there are parallelized algorithms that use similar techniques to select sets \cite{balkanski2018non,qian2019fast,FMZ19}. These methods provide an alternative approach to \textsc{Top-k} for selecting subsets of neurons and can be explored in future work.

\clearpage

\section{Proof that no-interaction in the difference NIE implies decomposition of the TE}
\label{app:decompose-proof}

Since by definition $\vy_{\texttt{set-gender}}(u)=\vy_{\texttt{set-gender},\vz_{\texttt{set-gender}}(u)}(u)$ and $\vy_{\texttt{null}}(u)=\vy_{\texttt{null},\vz_{\texttt{null}}(u)}(u)$,  it can be seen that both sides of Eq.~\ref{eq:nie-assump} describe a form of NIE (defined on the difference scale), where each contrasts $\vy$ under two different interventions on $\vz$ while keeping the sentence $u$ the same (left side, under \texttt{set-gender}; right side, under \texttt{null}).  
This equation  parallels a previously-described assumption in the causal mediation analysis literature that ascertains that the NIE is the same regardless of the fixed value at which the  intervention  (the analogue of \texttt{set-gender}/\texttt{null}) is held, known as a no-interaction assumption~\cite{imai2010identification}. 
We show that no-interaction in the indirect effect on the difference scale   
implies that the TE = NDE + NIE under our scale. 
Eq.  \ref{eq:nie-assump} can be rewritten as
\begin{align} 
&\vy_{\texttt{set-gender}}(u) - \vy_{\texttt{null}}(u) =\\ \nonumber &\vy_{\texttt{set-gender},\vz_{\texttt{null}}(u)}(u) - \vy_{\texttt{null}}(u)  \\ \nonumber
    + &\quad \vy_{\texttt{null},\vz_{\texttt{set-gender}}(u)}(u) -  \vy_{\texttt{null}}(u).
\end{align}
Now, dividing both sides of the equation by $\vy_{\texttt{null}}(u)$ and taking expectations over $u$ yields
\begin{align} 
&\E_{u} \left[ \vy_{\texttt{set-gender}}(u) / \vy_{\texttt{null}}(u) - 1 \right]=\\ \nonumber
&\E_u [ \vy_{\texttt{set-gender},\vz_{\texttt{null}}(u)}(u) / \vy_{\texttt{null}}(u) - 1 ] \ +\\ \nonumber
& \E_u [ \vy_{\texttt{null},\vz_{\texttt{set-gender}}(u)}(u) /  \vy_{\texttt{null}}(u) - 1], 
\end{align}
which is exactly 
\begin{align} 
&\text{TE}(\texttt{set-gender, \texttt{null}}; \vy) =\\ \nonumber &\text{NDE}(\texttt{set-gender}, \texttt{null}; \vy) \ + \\ \nonumber &\text{NIE}(\texttt{set-gender}, \texttt{null}; \vy).
\end{align}

\begin{figure}[h]
\centering
    \includegraphics[width=0.7\columnwidth]{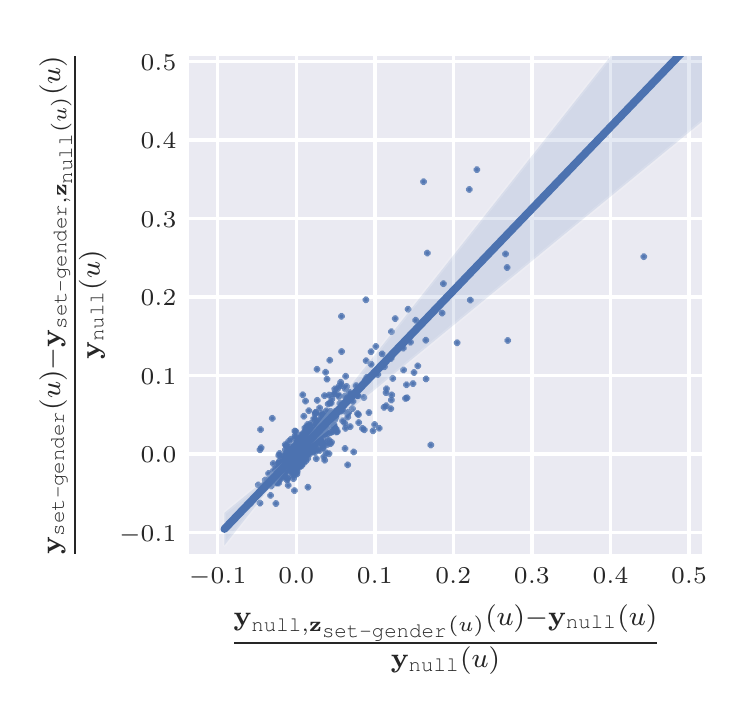}
    \caption{Plot of right side of Eq.~\ref{eq:nie-assump} ($x$ axis) against left side ($y$ axis), normalized by $y_{\texttt{null}}$. For visualization purposes, we exclude a single outlier at (0.60, 1.07).}
    \label{fig:nie_assump}
\end{figure}

It should be noted that even if Eq.~\ref{eq:nie-assump} does not hold, but the equation approximately holds upon dividing both sides by $y_\texttt{null}$, we would expect the decomposition TE $\approx$ NDE + NIE to hold. Indeed, further inspection of the trained model revealed that the left side and right side of Eq.~\ref{eq:nie-assump} were very close in the case of the attention intervention. Figure~\ref{fig:nie_assump} shows a plot of the values attained by the two sides of the equation (normalized by $y_\texttt{null}$ to make consistent with the earlier analyses) for all attention heads across all examples in the Winobias dataset. Fitting the data to a linear model yields a coefficient of 1.04 and an intercept of 0.00 ($R^2=0.78$).

\clearpage

\section{Alternate Metrics} \label{app:metrics}

Here are the results of select analyses from throughout the article, replicated using the different alternate metrics described in Section \ref{sec:metrics}. Specifically, we investigate a total of four metrics, consisting of the primary metric and the three alternate metrics, referred to and defined by the following:

\begin{itemize}
    \item Original metric: The primary measurement of bias and effects as described in the setup in Sections \ref{sec:cma}, \ref{sec:neuron-interventions}, and \ref{sec:attn-interventions} and as used throughout the article.
    \item Relative $\ell_{\infty}$ metric: Effects based on the distance measure described in Equation \ref{eq:linfty-def}.
    \item Normalized difference: Effects based on the distance measure described in Equation \ref{eq:pearl-def}.
    \item Total Variation distance (or TV-distance): Effects based on the distance measure described in Equation \ref{eq:tv-def}.
\end{itemize}

\begin{figure*}[t]
\centering
\begin{subfigure}[t]{.44\textwidth}
\centering
\includegraphics[width=\textwidth]{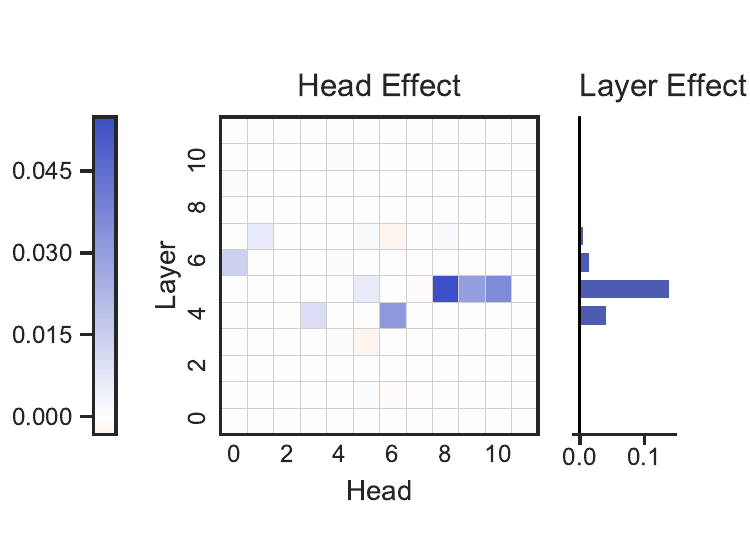}
\vspace{-2em}
\caption{Original metric}
\end{subfigure}\hfill
\begin{subfigure}[t]{.44\textwidth}
\centering
\includegraphics[width=\textwidth]{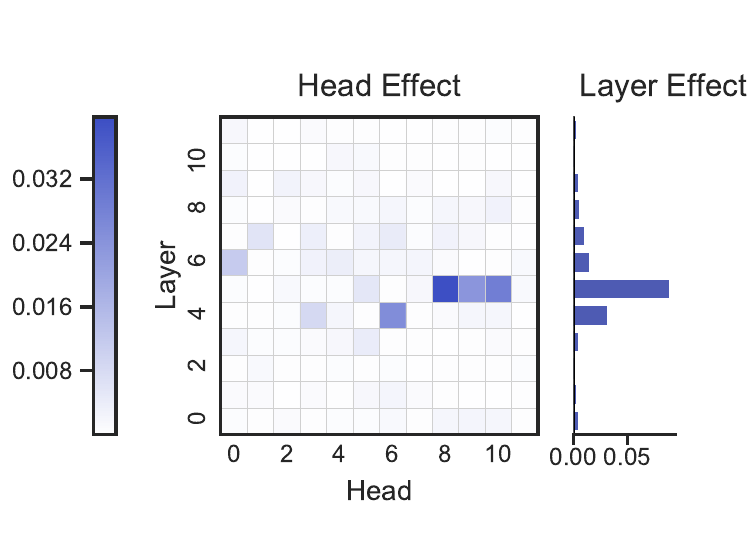}
\vspace{-2em}
\caption{Relative $\ell_{\infty}$ metric}
\end{subfigure}
\begin{subfigure}[t]{.44\textwidth}
\includegraphics[width=\textwidth]{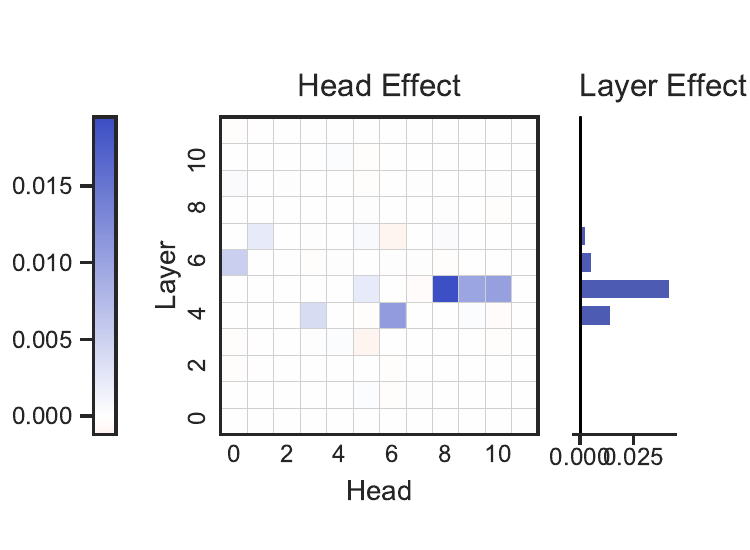}
\vspace{-2em}
\caption{Normalized difference}
\end{subfigure}\hfill
\begin{subfigure}[t]{.44\textwidth}
\includegraphics[width=\textwidth]{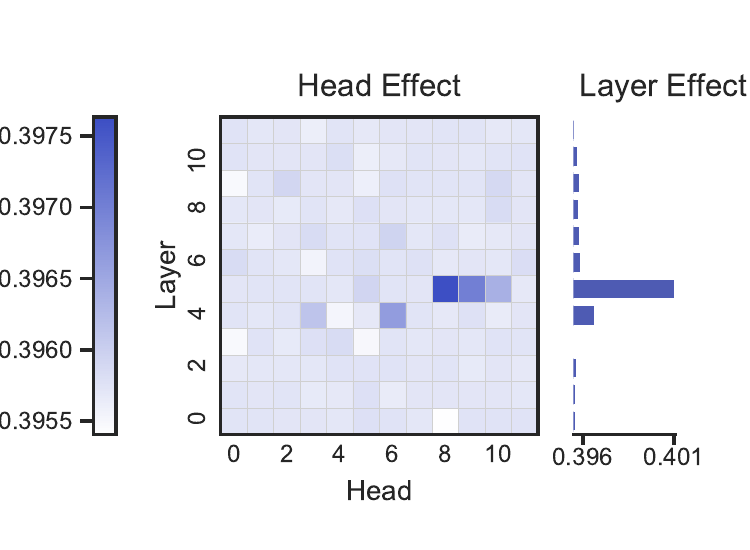}
\vspace{-2em}
\caption{Total Variation distance}
\end{subfigure}
\caption{Indirect effect for the filtered Winobias Dev data set on GPT2-small using various alternate metrics.}
\label{fig:winobias_heatmaps_metrics}
\end{figure*}

\begin{figure*}[t]
\centering
\begin{subfigure}[t]{.44\textwidth}
\centering
\includegraphics[width=\textwidth]{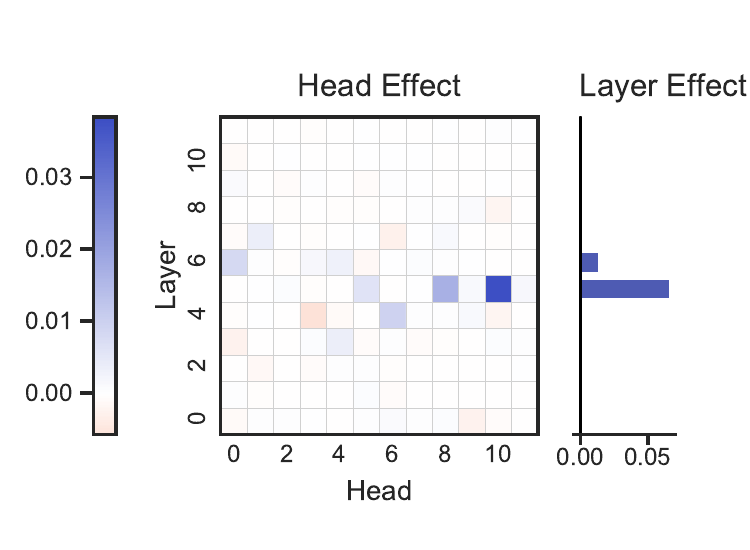}
\vspace{-2em}
\caption{Original metric}
\end{subfigure}\hfill
\begin{subfigure}[t]{.44\textwidth}
\centering
\includegraphics[width=\textwidth]{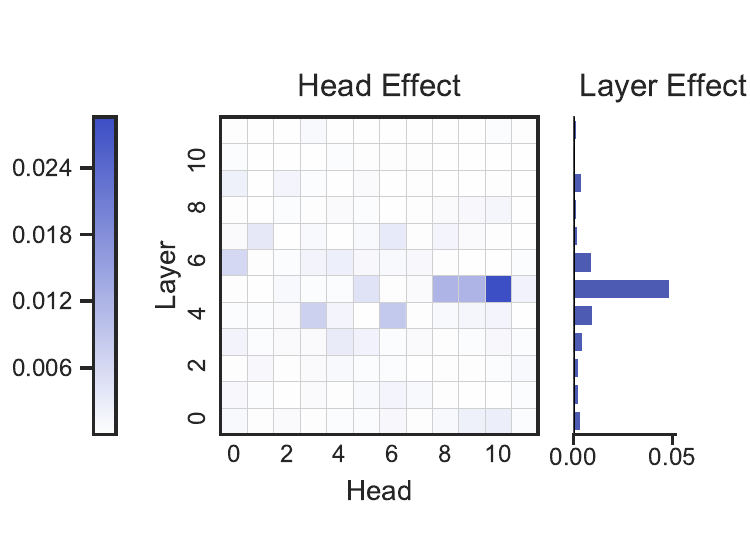}
\vspace{-2em}
\caption{Relative $\ell_{\infty}$ metric}
\end{subfigure}
\begin{subfigure}[t]{.44\textwidth}
\includegraphics[width=\textwidth]{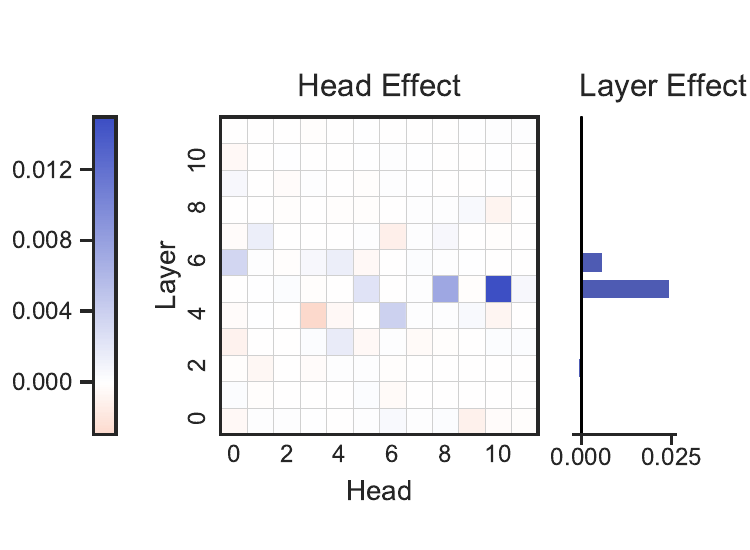}
\vspace{-2em}
\caption{Normalized difference}
\end{subfigure}\hfill
\begin{subfigure}[t]{.44\textwidth}
\includegraphics[width=\textwidth]{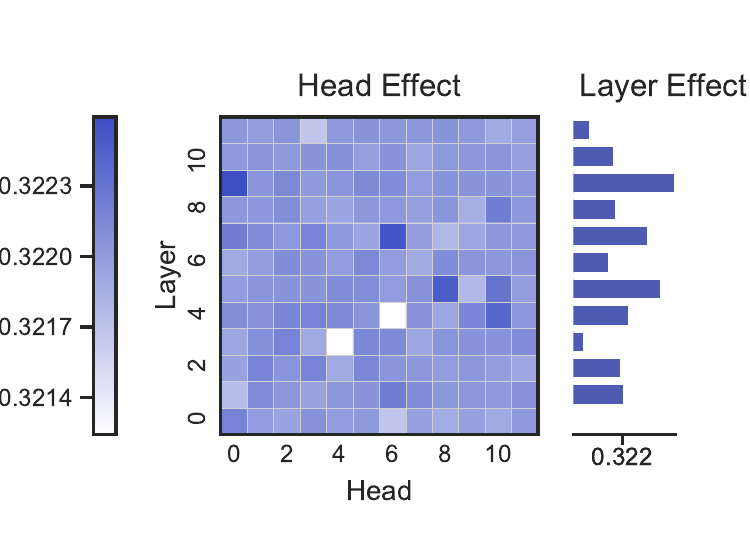}
\vspace{-2em}
\caption{Total Variation distance}
\end{subfigure}
\caption{Indirect effect for the filtered Winogender Bergsma data set on GPT2-small using various alternate metrics.}
\label{fig:winogender_heatmaps_metrics}
\end{figure*}

\begin{figure*}[t]
\centering
\begin{subfigure}[t]{.47\textwidth}
\centering
\includegraphics[width=\textwidth]{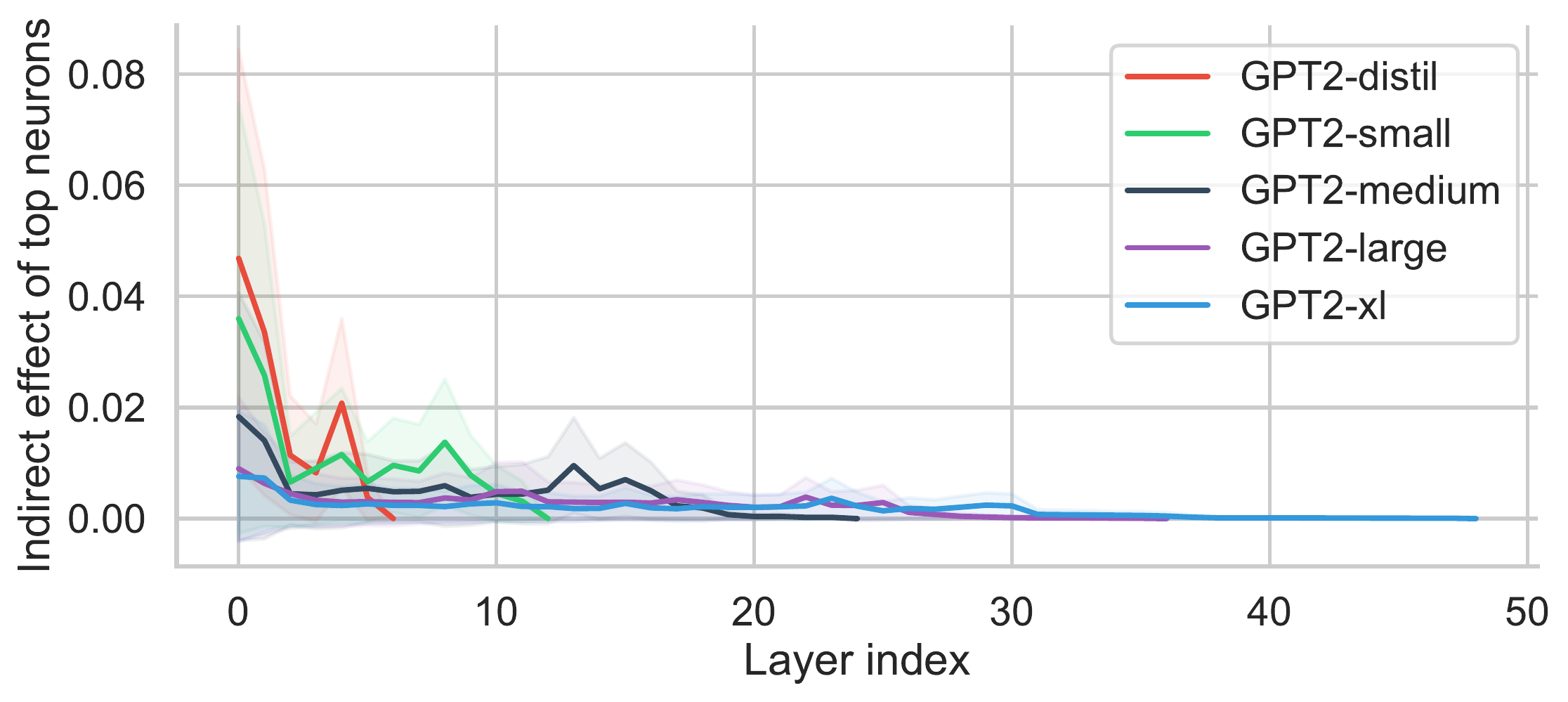}
\vspace{-2em}
\caption{Original metric}
\end{subfigure}\hfill
\begin{subfigure}[t]{.47\textwidth}
\centering
\includegraphics[width=\textwidth]{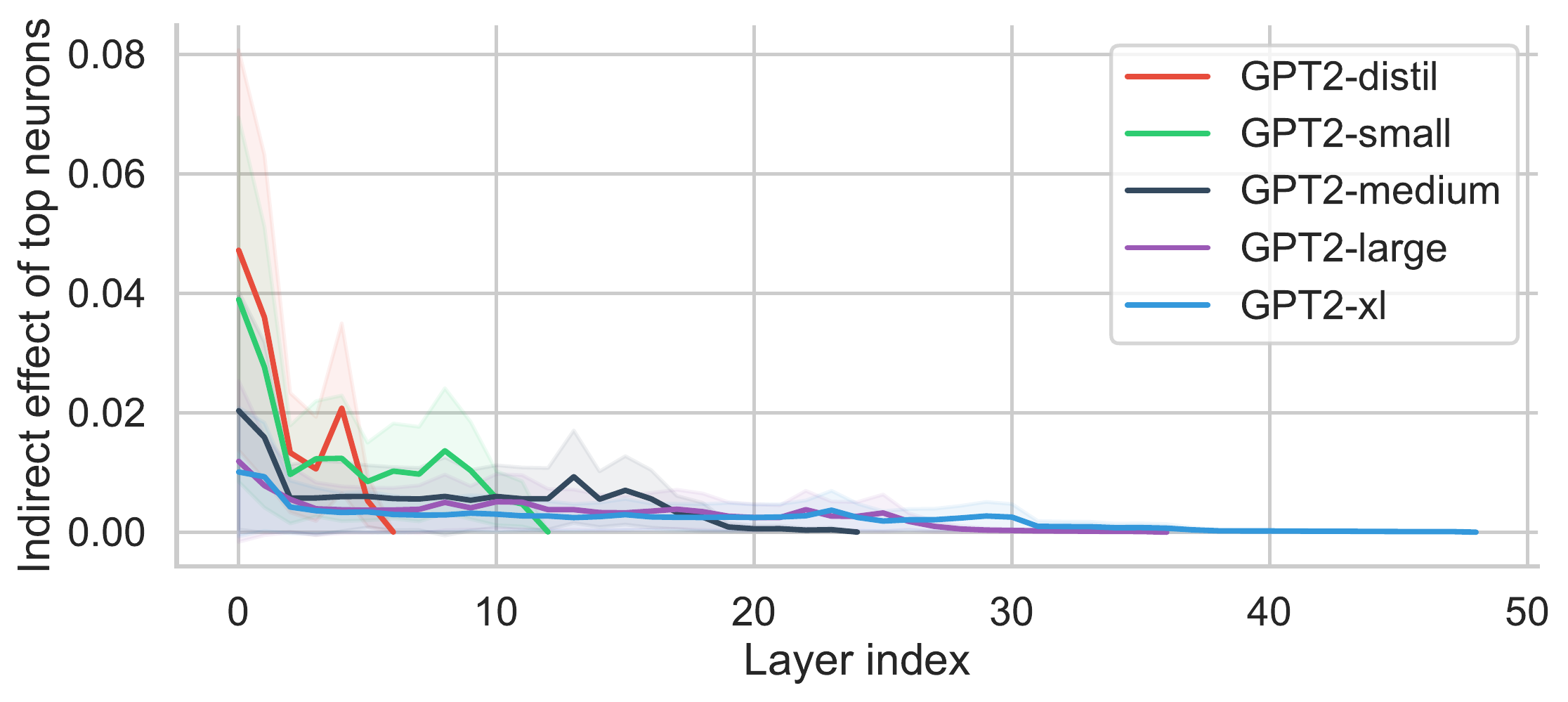}
\vspace{-2em}
\caption{Relative $\ell_{\infty}$ metric}
\end{subfigure}
\begin{subfigure}[t]{.47\textwidth}
\includegraphics[width=\textwidth]{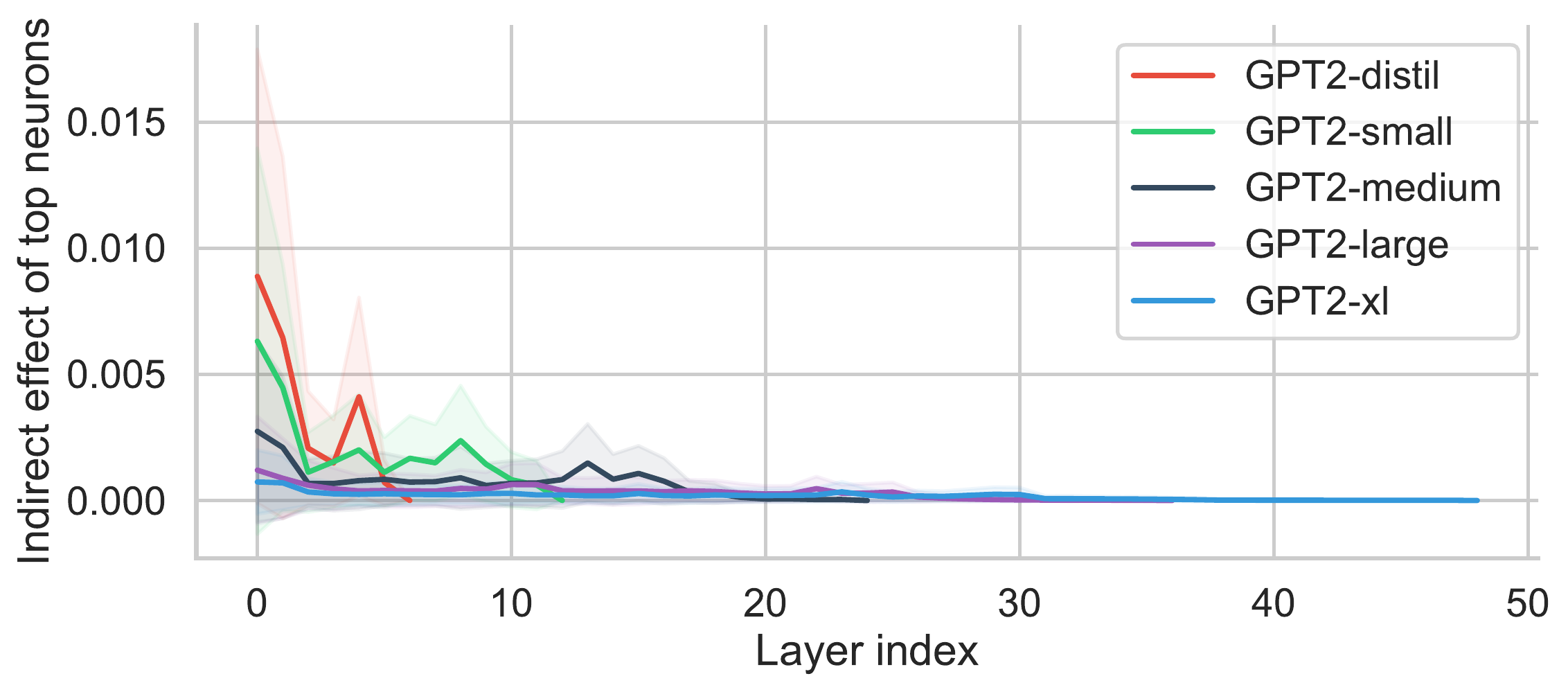}
\vspace{-2em}
\caption{Normalized difference}
\end{subfigure}\hfill
\begin{subfigure}[t]{.47\textwidth}
\includegraphics[width=\textwidth]{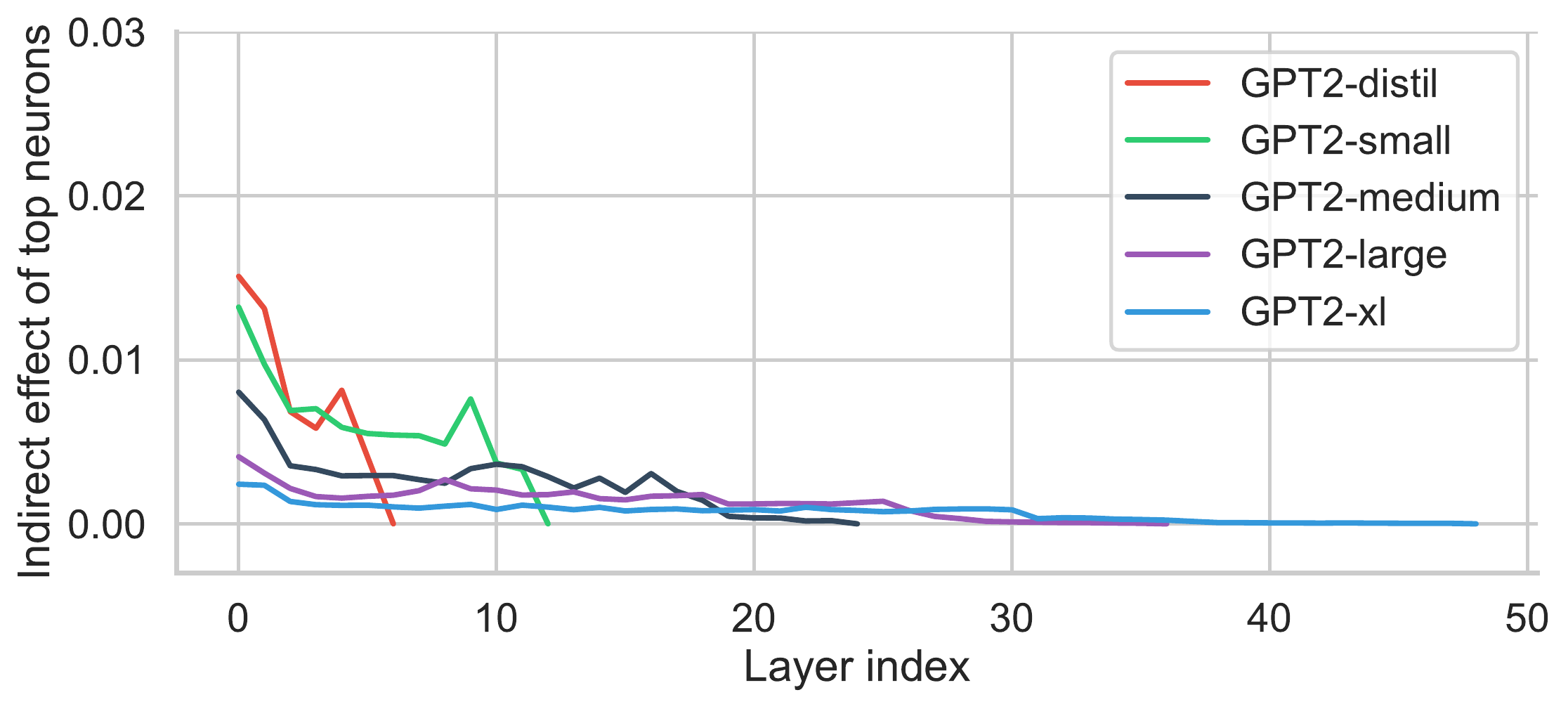}
\vspace{-2em}
\caption{Total Variation distance}
\end{subfigure}
\caption{Indirect effect for the Professions data set on GPT2-small using various alternate metrics.}
\label{fig:professions_heatmaps_metrics}
\end{figure*}

Figures \ref{fig:winobias_heatmaps_metrics}, \ref{fig:winogender_heatmaps_metrics}, and \ref{fig:professions_heatmaps_metrics} evaluate all the metrics on the GPT2-small model using the filtered Winobias Dev, filtered Winogender Bergsma, and Professions data sets respectively.

\begin{figure*}[t]
\centering
\begin{subfigure}[t]{.45\textwidth}
    \includegraphics[width=1\linewidth]{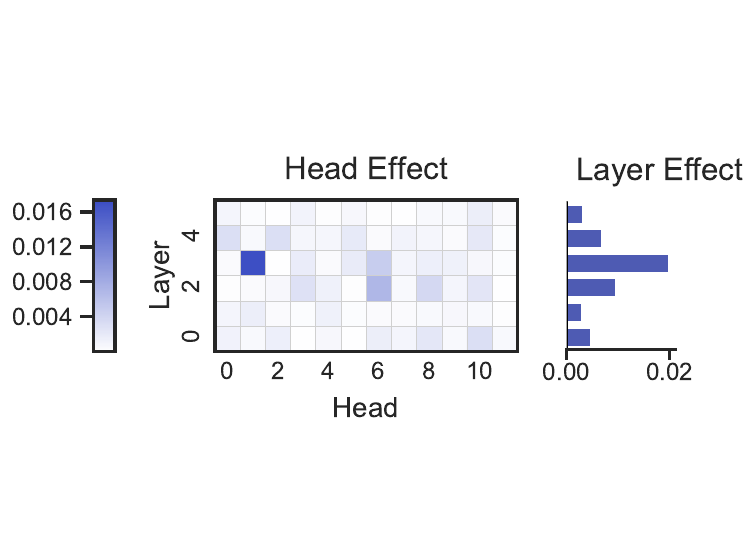}
\caption{GPT2-distil}
\vspace{1em}

\end{subfigure}\hfill
\begin{subfigure}[t]{.45\textwidth}
    \centering
    \includegraphics[width=1\linewidth]{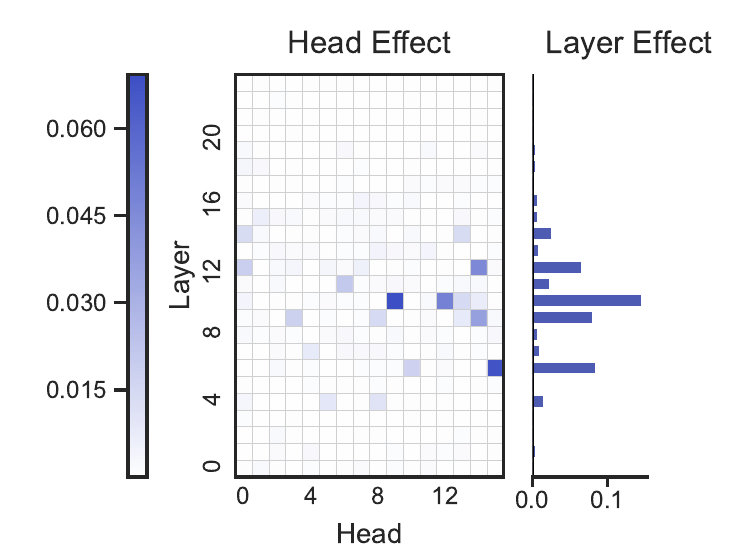}
\caption{GPT2-medium }
\vspace{1em}

\end{subfigure}
\begin{subfigure}[t]{.45\textwidth}
    \centering
    \includegraphics[width=1\linewidth]{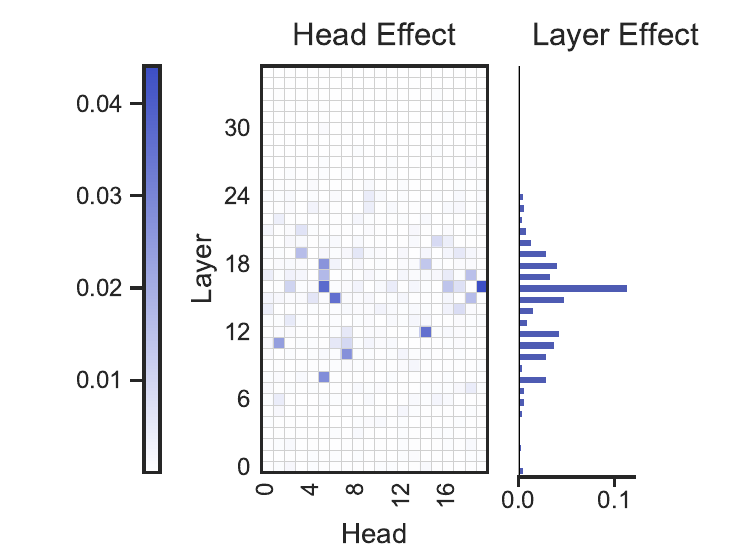}
\caption{GPT2-large}
\end{subfigure}\hfill
\begin{subfigure}[t]{.45\textwidth}
    \centering
    \includegraphics[width=1\linewidth]{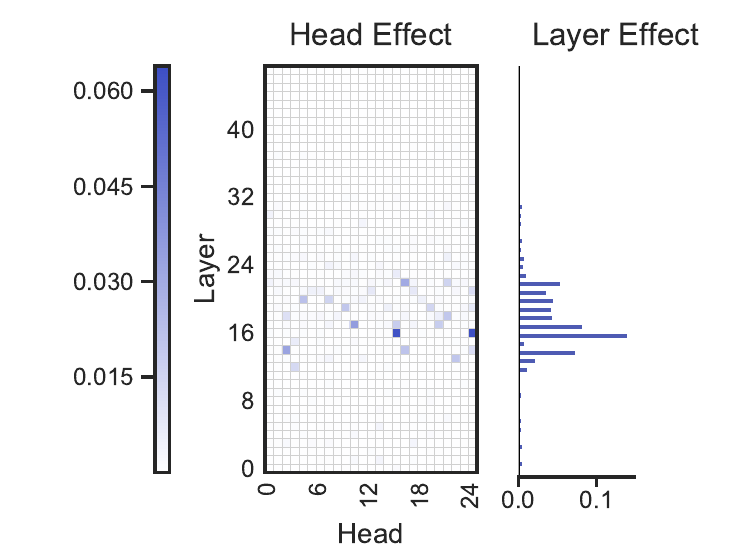}
\caption{GPT2-xl}
\end{subfigure}
\caption{Mean indirect effect on Winobias for heads (the heatmap) and layers (the bar chart) over additional GPT2 variants as measured using the alternate metric, relative $\ell_{\infty}$.}
\label{fig:indirect_heatmap_by_model_linfty}
\end{figure*}

\begin{figure*}[t]
\centering
\begin{subfigure}[t]{.45\textwidth}
    \includegraphics[width=1\linewidth]{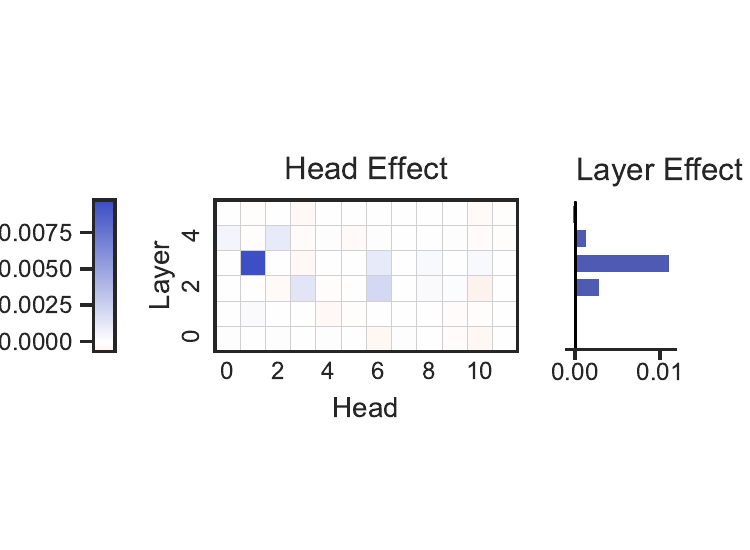}
\caption{GPT2-distil}
\vspace{1em}

\end{subfigure}\hfill
\begin{subfigure}[t]{.45\textwidth}
    \centering
    \includegraphics[width=1\linewidth]{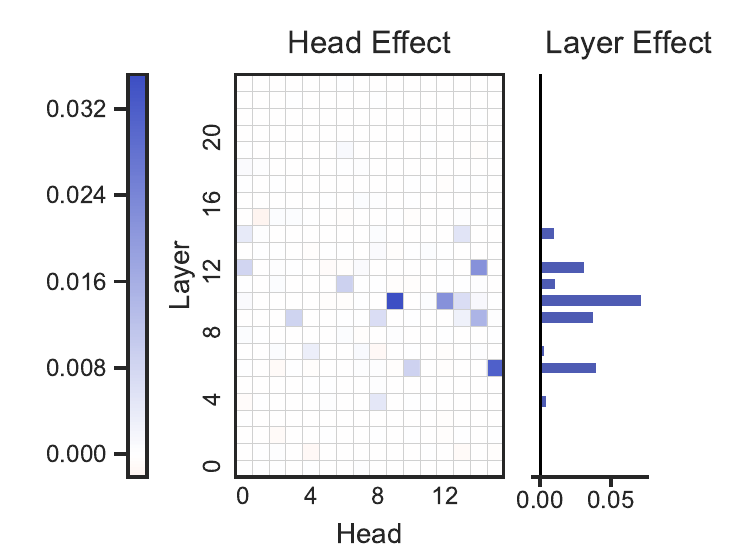}
\caption{GPT2-medium }
\vspace{1em}

\end{subfigure}
\begin{subfigure}[t]{.45\textwidth}
    \centering
    \includegraphics[width=1\linewidth]{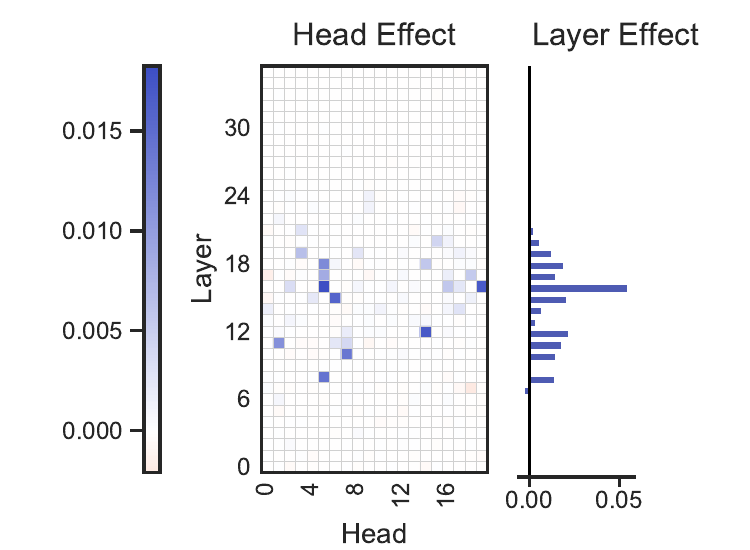}
\caption{GPT2-large}
\end{subfigure}\hfill
\begin{subfigure}[t]{.45\textwidth}
    \centering
    \includegraphics[width=1\linewidth]{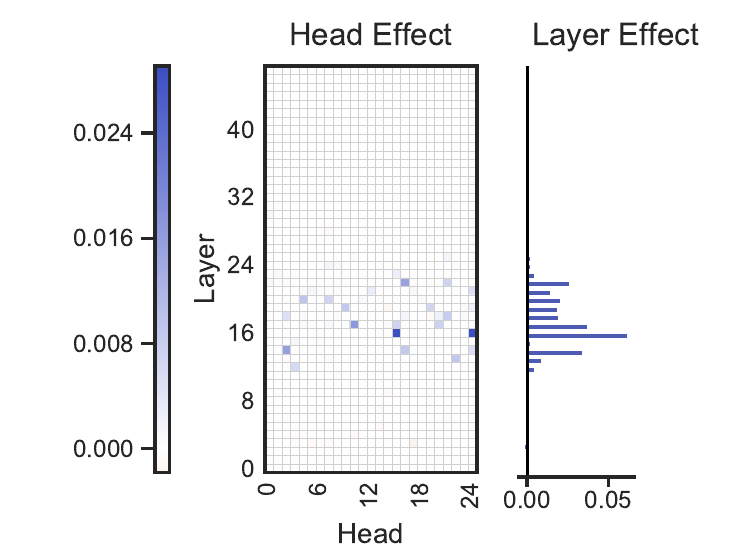}
\caption{GPT2-xl}
\end{subfigure}
\caption{Mean indirect effect on Winobias for heads (the heatmap) and layers (the bar chart) over additional GPT2 variants as measured using the alternate metric, normalized difference.}
\label{fig:indirect_heatmap_by_model_pearl}
\end{figure*}

\begin{figure*}[t]
\centering
\begin{subfigure}[t]{.45\textwidth}
    \includegraphics[width=1\linewidth]{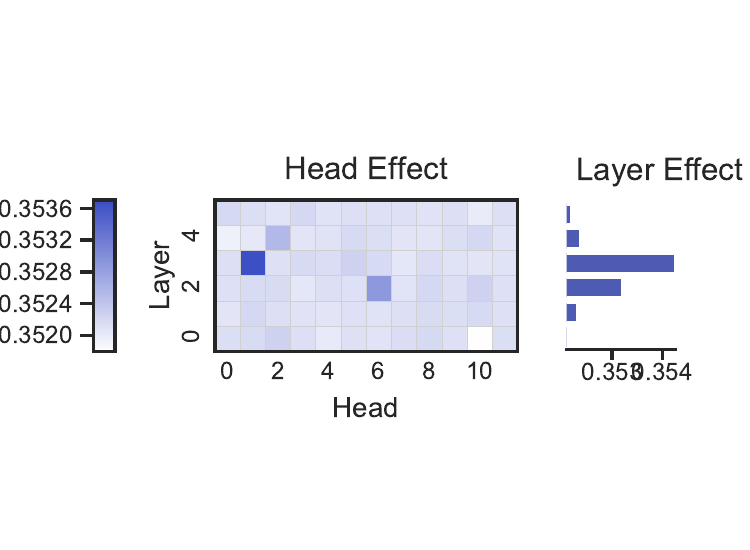}
\caption{GPT2-distil}
\vspace{1em}

\end{subfigure}\hfill
\begin{subfigure}[t]{.45\textwidth}
    \centering
    \includegraphics[width=1\linewidth]{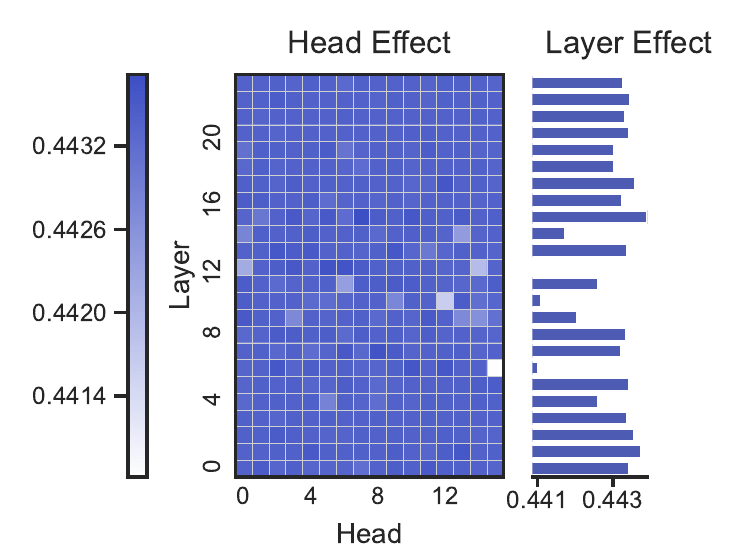}
\caption{GPT2-medium }
\vspace{1em}

\end{subfigure}
\begin{subfigure}[t]{.45\textwidth}
    \centering
    \includegraphics[width=1\linewidth]{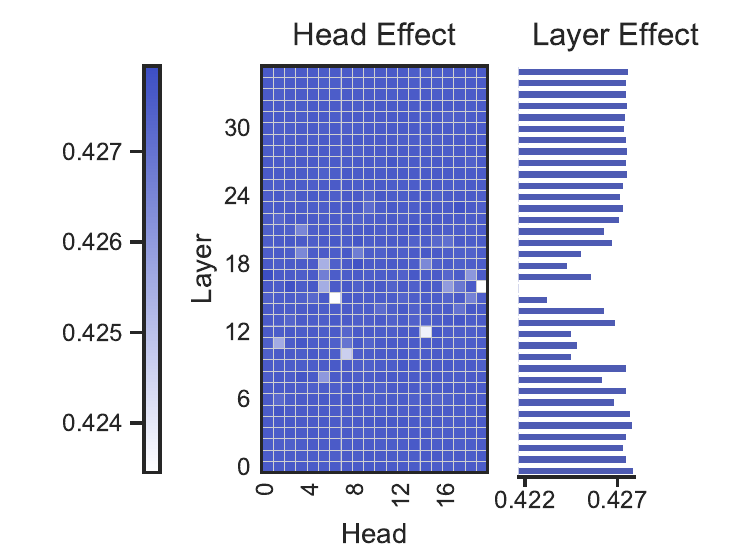}
\caption{GPT2-large}
\end{subfigure}\hfill
\begin{subfigure}[t]{.45\textwidth}
    \centering
    \includegraphics[width=1\linewidth]{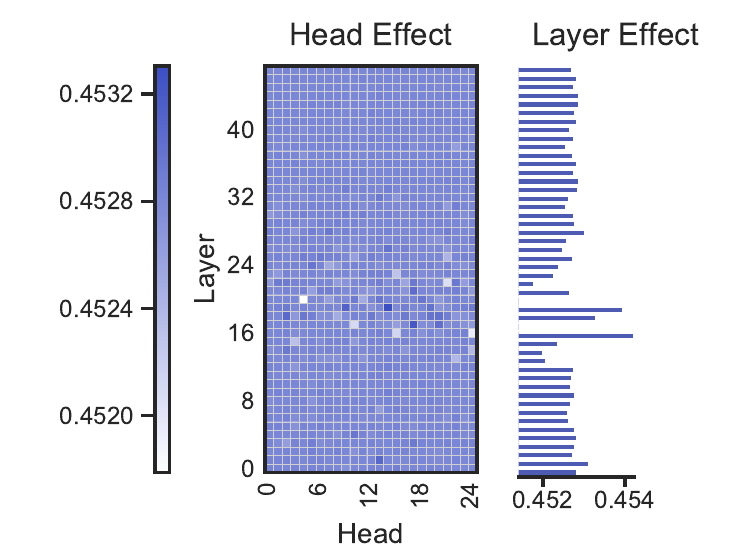}
\caption{GPT2-xl}
\end{subfigure}
\caption{Mean indirect effect on Winobias for heads (the heatmap) and layers (the bar chart) over additional GPT2 variants as measured using the alternate metric, TV-distance.}
\label{fig:indirect_heatmap_by_model_tv}
\end{figure*}

Figures \ref{fig:indirect_heatmap_by_model_linfty}, \ref{fig:indirect_heatmap_by_model_pearl}, and \ref{fig:indirect_heatmap_by_model_tv} evaluate the three alternate metrics -- the relative $\ell_{\infty}$ metric, normalized difference, and TV-distance, respectively -- on the filtered Winobias Dev data set for different GPT2 model variants. These can be contrasted with the comparable results of the original metric in Figure \ref{fig:indirect_heatmap_by_model}.

Tables \ref{tab:metrics-te-winobias} and \ref{tab:metrics-te-winogender} reporting total effects are also included. Overall, the main takeaways are that the results of our analyses are quite robust to the metric that is being used to compute the effects. There are minor visible differences, but the relative behavior overall is consistent, which is not necessarily an intuitive result.

For instance, one notable difference is that there appears to be less absolute sparsity, but relative sparsity holds, which suggests that these alternate metrics are more sensitive to the effects of causal mediation analysis while demonstrating that the conclusions drawn by our article continue to hold in all of these situations. Similarly, larger models appear to exhibit a more pronounced effect than smaller models even under alternate metrics, which is another consistent result.

It is worth noting that because of the way that TV-distance is constructed in terms of difference between magnitudes rather than ratios, the effect values for each model and each data set are on different levels. However, when normalizing relative to a constant (i.e., the minimum effect across layers for the neuron line plots or across heads for the attention heatmaps), then the emerging behavior is typically consistent with the broader patterns that have been identified.

\clearpage

\clearpage

\begin{table*}[h]
\centering
\begin{tabular}{l r r r r r r r r r}
\toprule 
& \multicolumn{4}{c}{Winobias Dev Filtered} \\
\cmidrule(lr){2-5} %
Model & Original & Relative $\ell_{\infty}$ & Norm. Diff. & TV-distance\\
\midrule 
GPT2-distil & 0.118 & 0.105 & 0.045 & 0.358 \\
GPT2-small & 0.249 & 0.183 & 0.079 & 0.405 \\
GPT2-medium & 0.774 & 0.409 & 0.164 & 0.438 \\
GPT2-large & 0.751 & 0.383 & 0.166 & 0.425 \\
GPT2-xl & 1.049 & 0.446 & 0.205 & 0.456 \\
\bottomrule 
\end{tabular}
\caption{Total effects on Winobias Dev filtered for all metrics using different GPT2 variants.}
\vspace{-.5em}
\label{tab:metrics-te-winobias} 
\end{table*}

\begin{table*}[h]
\centering
\begin{tabular}{l r r r r r r r r r}
\toprule 
& \multicolumn{4}{c}{Winogender Bergsma Filtered} \\
\cmidrule(lr){2-5}
Model & Original & Relative $\ell_{\infty}$ & Norm. Diff. & TV-distance\\
\midrule 
GPT2-distil & 0.075 & 0.068 & 0.029 & 0.341 \\
GPT2-small & 0.135 & 0.105 & 0.050 & 0.318 \\
GPT2-medium & 0.384 & 0.226 & 0.126 & 0.360 \\
GPT2-large & 0.350 & 0.222 & 0.120 & 0.322 \\
GPT2-xl & 0.362 & 0.224 & 0.110 & 0.364 \\
\bottomrule 
\end{tabular}
\caption{Total effects on Winogender Bergsma filtered for all metrics using different GPT2 variants.}
\vspace{-.5em}
\label{tab:metrics-te-winogender} 
\end{table*}

\clearpage

\section{Attention intervention experiments with masked language models} \label{app:other-models}

\begin{figure*}[t]
    \centering
    \begin{subfigure}[t]{.45\textwidth}
        \includegraphics[width=1\linewidth]{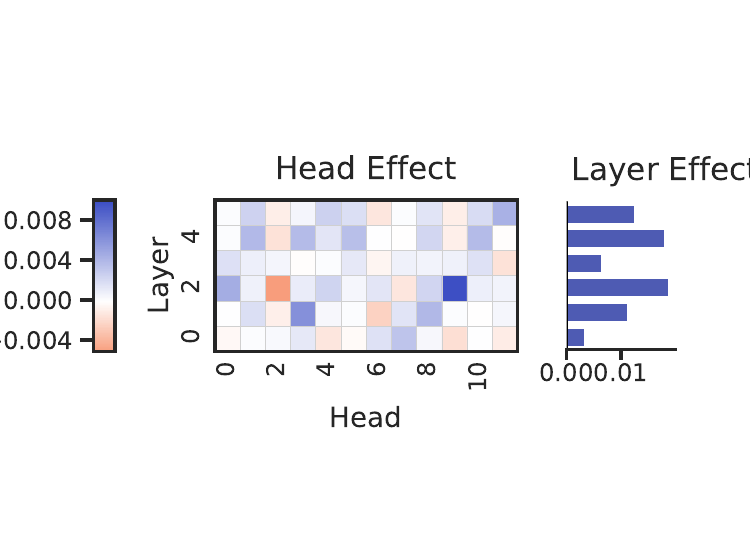}
        \caption{Scoring scheme 1}
        \vspace{1em}
    \end{subfigure}\hfill
    \begin{subfigure}[t]{.45\textwidth}
        \centering
        \includegraphics[width=1\linewidth]{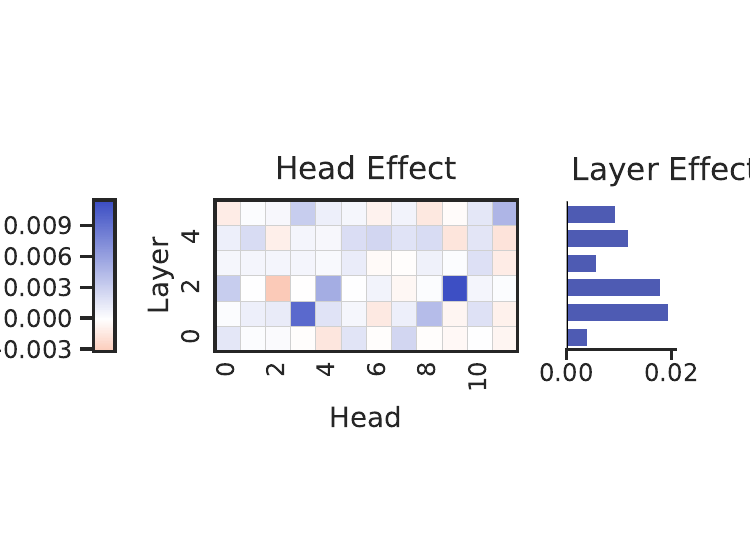}
        \caption{Scoring scheme 2}
        \vspace{1em}
    \end{subfigure}
    \begin{subfigure}[t]{.45\textwidth}
        \centering
        \includegraphics[width=1\linewidth]{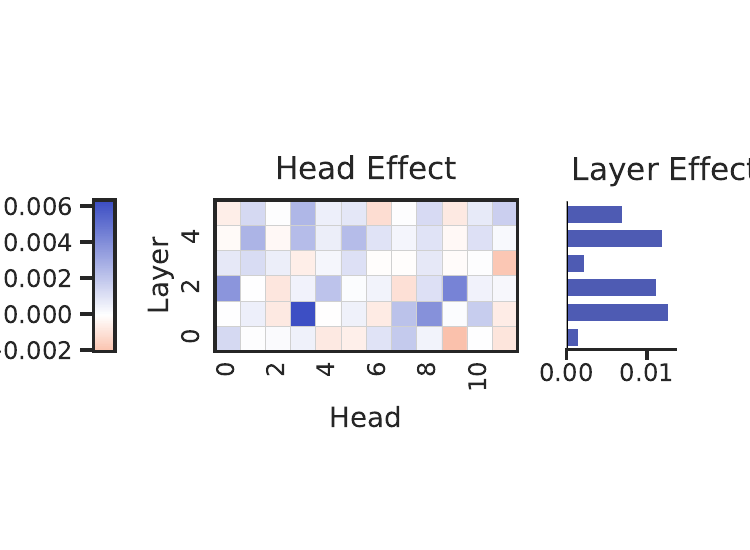}
        \caption{Scoring scheme 3}
        \vspace{1em}
    \end{subfigure}\hfill
    \begin{subfigure}[t]{.45\textwidth}
        \centering
        \includegraphics[width=1\linewidth]{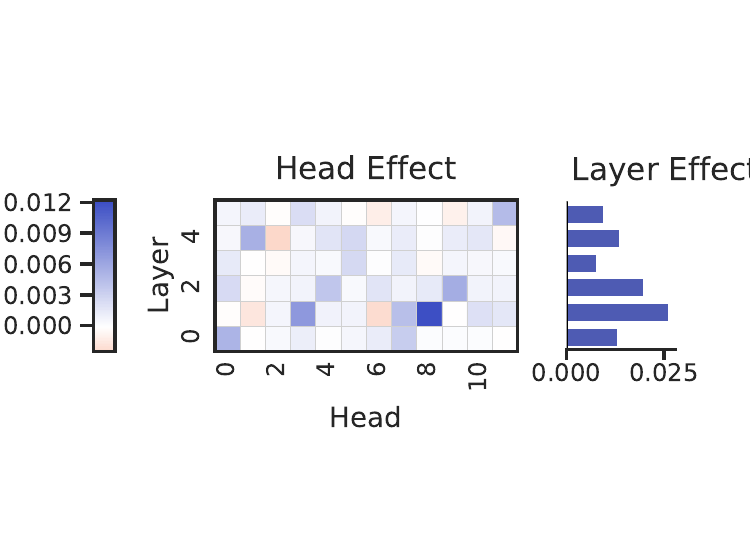}
        \caption{Scoring scheme 1'}
        \vspace{1em}
    \end{subfigure}
    \begin{subfigure}[t]{.45\textwidth}
        \centering
        \includegraphics[width=1\linewidth]{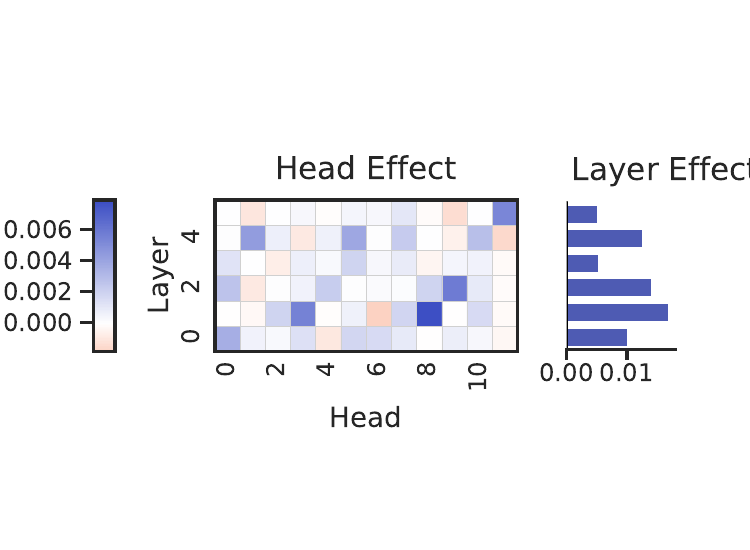}
        \caption{Scoring scheme 2'}
    \end{subfigure}\hfill
    \begin{subfigure}[t]{.45\textwidth}
        \centering
        \includegraphics[width=1\linewidth]{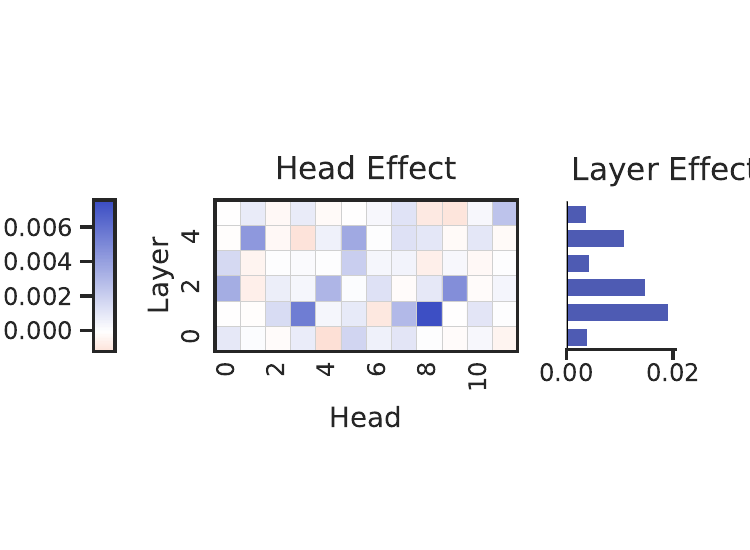}
        \caption{Scoring scheme 3'}
    \end{subfigure}
    \caption{Mean indirect effect on Winobias for heads (the heatmap) and layers (the bar chart) in DistilBERT over different scoring schemes.}
    \label{fig:distilbert-scoring}
\end{figure*}

\begin{figure*}[t]
    \centering
    \begin{subfigure}[t]{.45\textwidth}
        \includegraphics[width=1\linewidth]{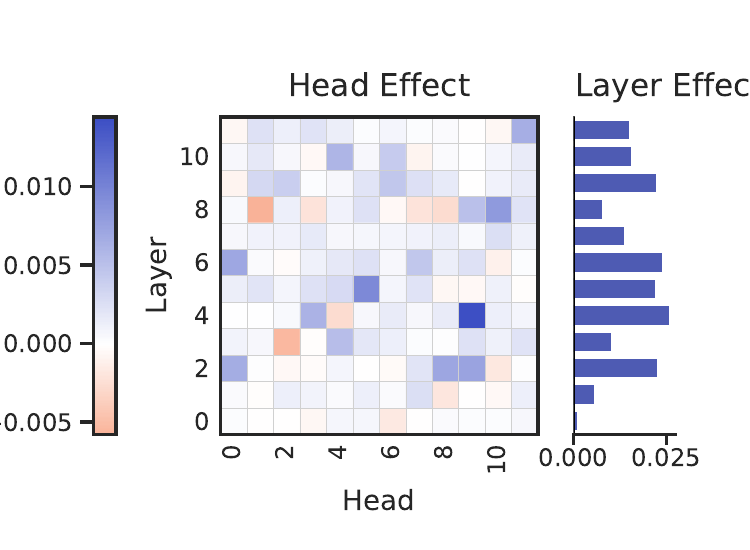}
        \caption{Scoring scheme 1}
        \vspace{1em}
    \end{subfigure}\hfill
    \begin{subfigure}[t]{.45\textwidth}
        \centering
        \includegraphics[width=1\linewidth]{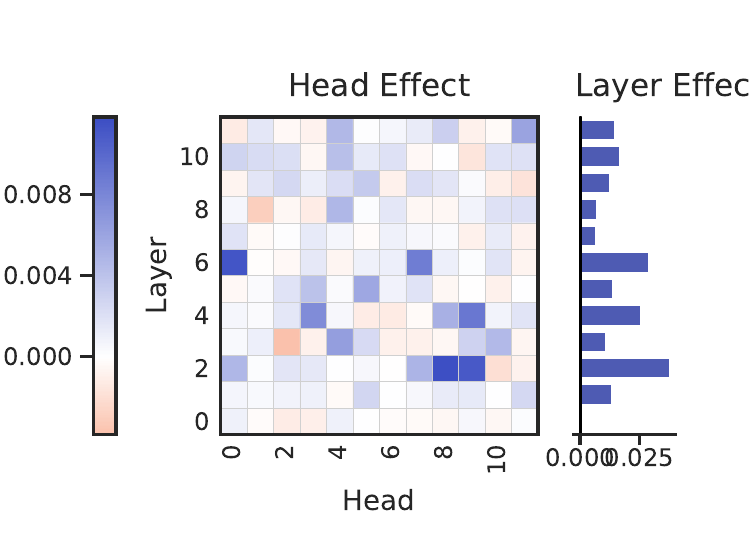}
        \caption{Scoring scheme 2}
        \vspace{1em}
    \end{subfigure}
    \begin{subfigure}[t]{.45\textwidth}
        \centering
        \includegraphics[width=1\linewidth]{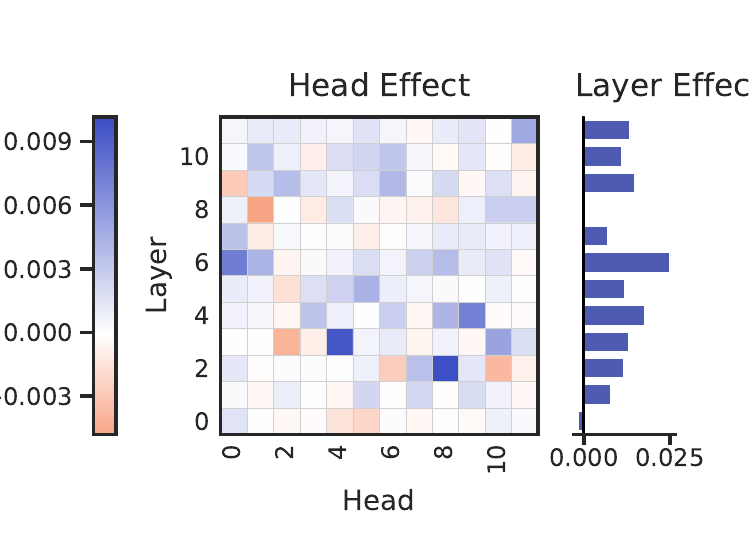}
        \caption{Scoring scheme 3}
        \vspace{1em}
    \end{subfigure}\hfill
    \begin{subfigure}[t]{.45\textwidth}
        \centering
        \includegraphics[width=1\linewidth]{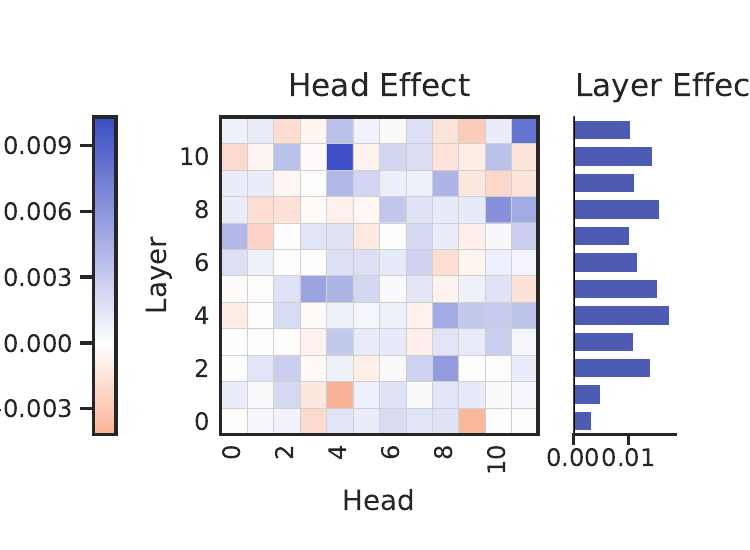}
        \caption{Scoring scheme 1'}
        \vspace{1em}
    \end{subfigure}
    \begin{subfigure}[t]{.45\textwidth}
        \centering
        \includegraphics[width=1\linewidth]{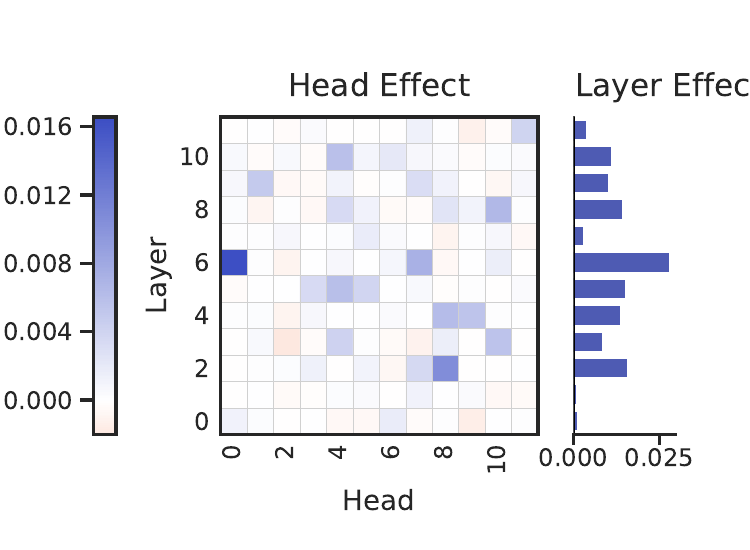}
        \caption{Scoring scheme 2'}
    \end{subfigure}\hfill
    \begin{subfigure}[t]{.45\textwidth}
        \centering
        \includegraphics[width=1\linewidth]{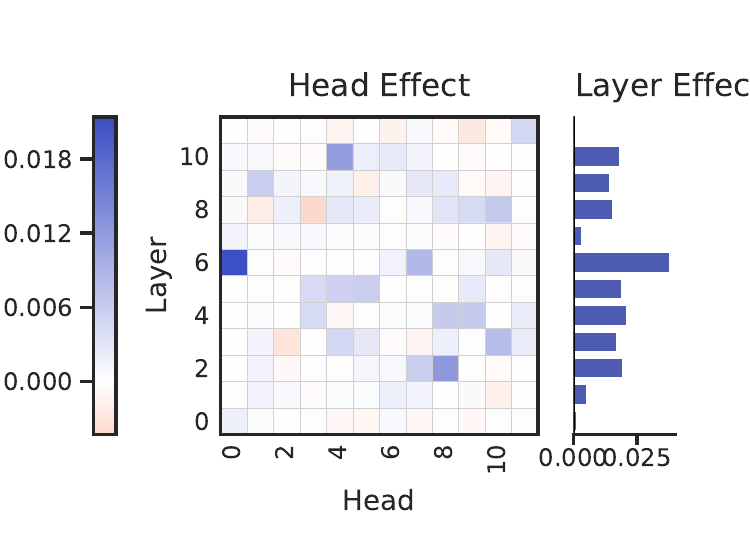}
        \caption{Scoring scheme 3'}
    \end{subfigure}
    \caption{Mean indirect effect on Winobias for heads (the heatmap) and layers (the bar chart) in BERT-base-uncased over different scoring schemes.}
    \label{fig:bert-base-scoring}
\end{figure*}

\begin{figure*}[t]
    \centering
    \begin{subfigure}[t]{.45\textwidth}
        \includegraphics[width=1\linewidth]{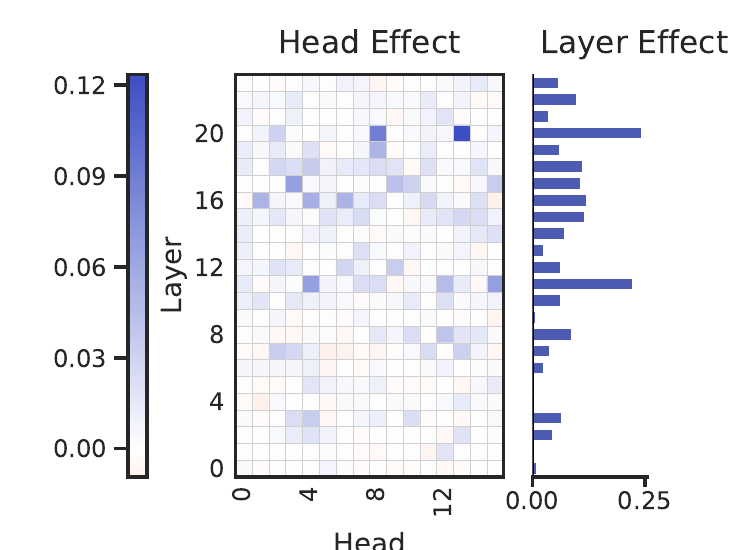}
        \caption{Scoring scheme 1}
        \vspace{1em}
    \end{subfigure}\hfill
    \begin{subfigure}[t]{.45\textwidth}
        \centering
        \includegraphics[width=1\linewidth]{images/winobias_bert-large-uncased_filtered_dev_2.pdf}
        \caption{Scoring scheme 2}
        \vspace{1em}
    \end{subfigure}
    \begin{subfigure}[t]{.45\textwidth}
        \centering
        \includegraphics[width=1\linewidth]{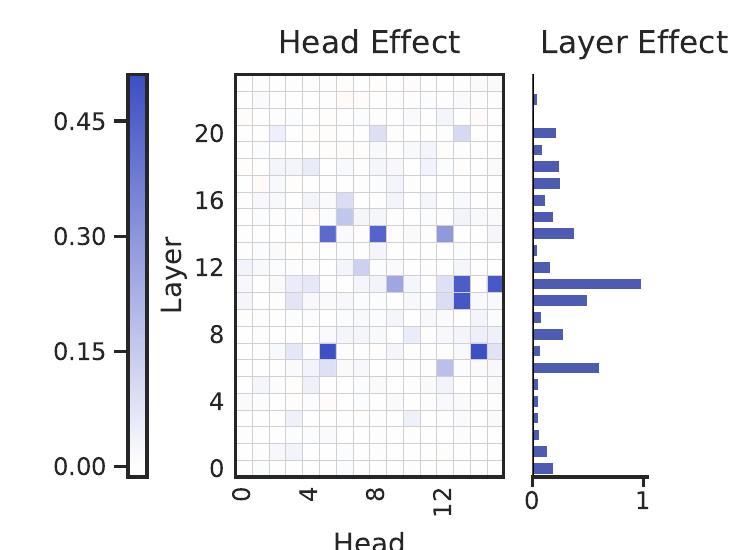}
        \caption{Scoring scheme 3}
        \vspace{1em}
    \end{subfigure}\hfill
    \begin{subfigure}[t]{.45\textwidth}
        \centering
        \includegraphics[width=1\linewidth]{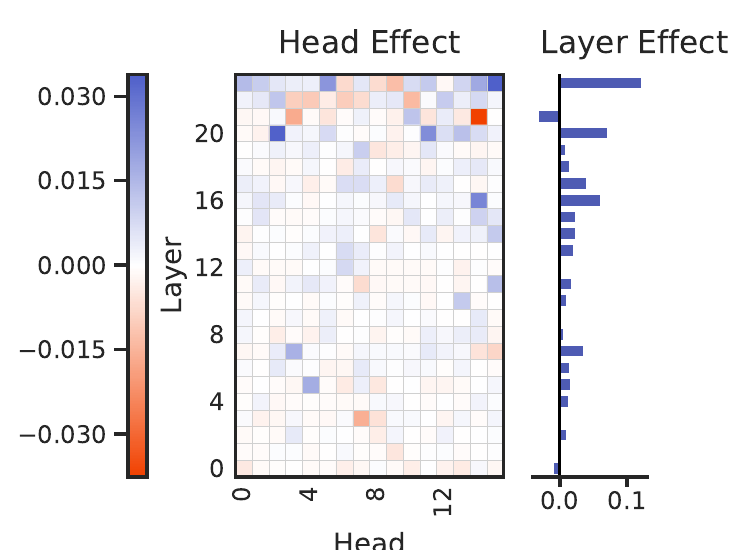}
        \caption{Scoring scheme 1'}
        \vspace{1em}
    \end{subfigure}
    \begin{subfigure}[t]{.45\textwidth}
        \centering
        \includegraphics[width=1\linewidth]{images/winobias_bert-large-uncased_filtered_dev_5.pdf}
        \caption{Scoring scheme 2'}
    \end{subfigure}\hfill
    \begin{subfigure}[t]{.45\textwidth}
        \centering
        \includegraphics[width=1\linewidth]{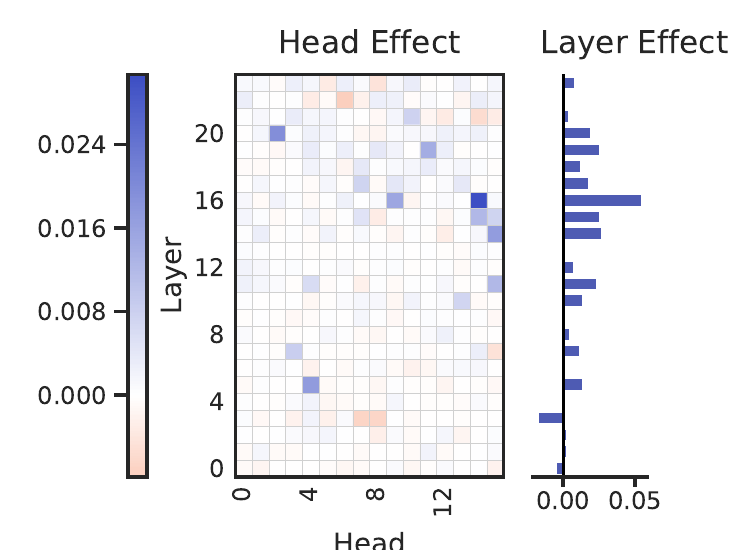}
        \caption{Scoring scheme 3'}
    \end{subfigure}
    \caption{Mean indirect effect on Winobias for heads (the heatmap) and layers (the bar chart) in BERT-large-uncased over different scoring schemes.}
    \label{fig:bert-large-scoring}
\end{figure*}

\begin{figure*}[t]
    \centering
    \begin{subfigure}[t]{.45\textwidth}
        \includegraphics[width=1\linewidth]{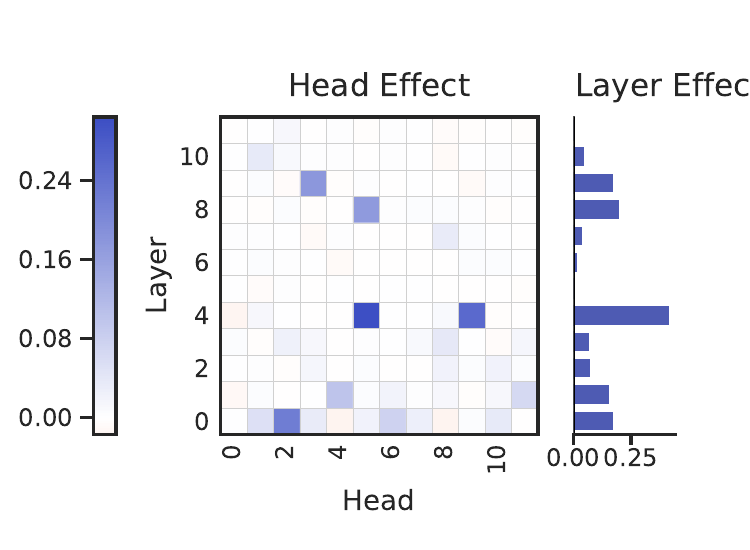}
        \caption{Scoring scheme 1}
        \vspace{1em}
    \end{subfigure}\hfill
    \begin{subfigure}[t]{.45\textwidth}
        \centering
        \includegraphics[width=1\linewidth]{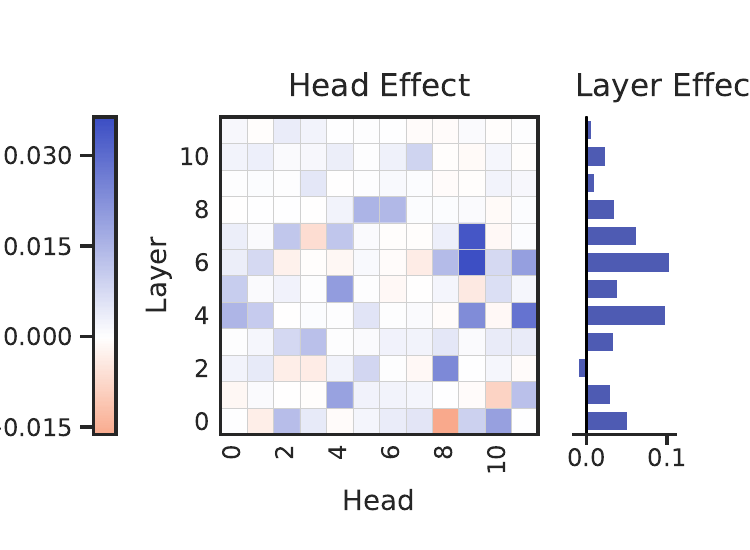}
        \caption{Scoring scheme 2}
        \vspace{1em}
    \end{subfigure}
    \begin{subfigure}[t]{.45\textwidth}
        \centering
        \includegraphics[width=1\linewidth]{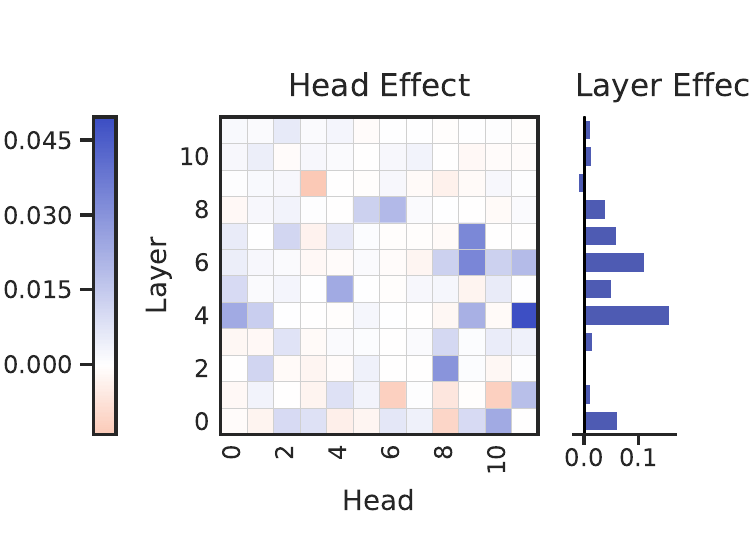}
        \caption{Scoring scheme 3}
        \vspace{1em}
    \end{subfigure}\hfill
    \begin{subfigure}[t]{.45\textwidth}
        \centering
        \includegraphics[width=1\linewidth]{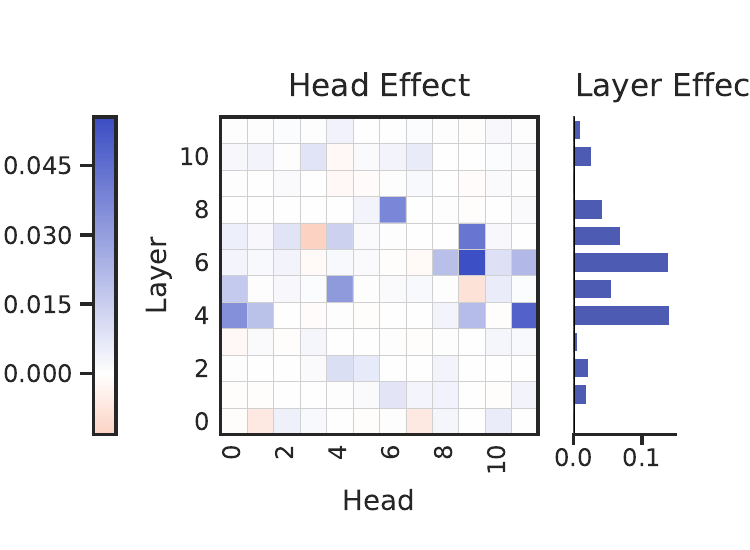}
        \caption{Scoring scheme 1'}
        \vspace{1em}
    \end{subfigure}
    \begin{subfigure}[t]{.45\textwidth}
        \centering
        \includegraphics[width=1\linewidth]{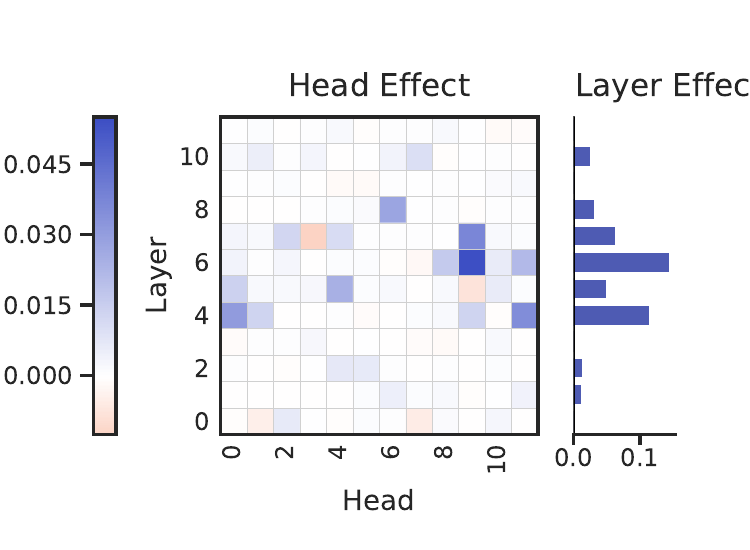}
        \caption{Scoring scheme 2'}
    \end{subfigure}\hfill
    \begin{subfigure}[t]{.45\textwidth}
        \centering
        \includegraphics[width=1\linewidth]{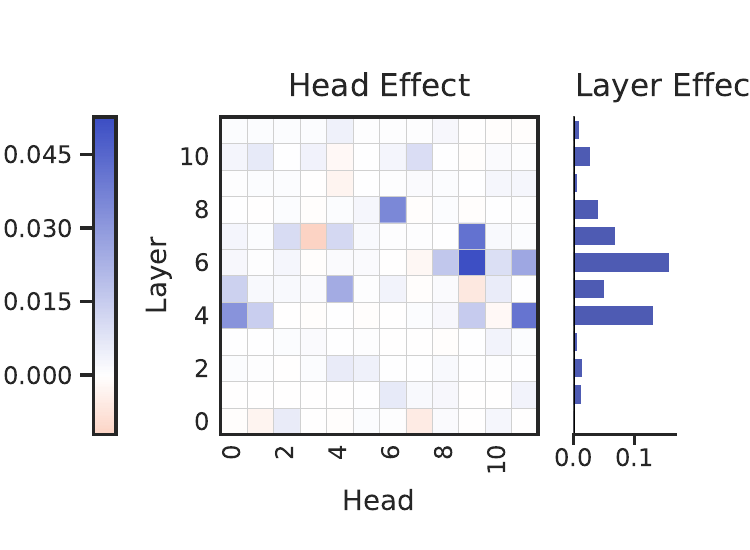}
        \caption{Scoring scheme 3'}
    \end{subfigure}
    \caption{Mean indirect effect on Winobias for heads (the heatmap) and layers (the bar chart) in RoBERTa-base over different scoring schemes.}
    \label{fig:robert-base-scoring}
\end{figure*}

\begin{figure*}[t]
    \centering
    \begin{subfigure}[t]{.45\textwidth}
        \includegraphics[width=1\linewidth]{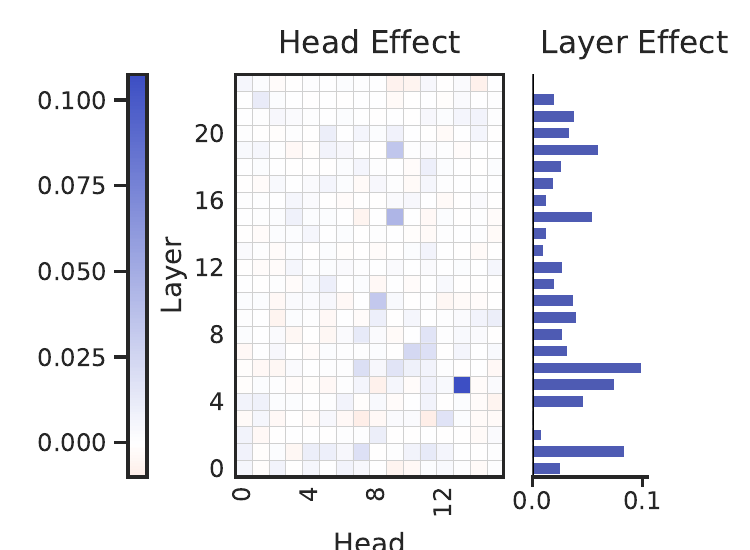}
        \caption{Scoring scheme 1}
        \vspace{1em}
    \end{subfigure}\hfill
    \begin{subfigure}[t]{.45\textwidth}
        \centering
        \includegraphics[width=1\linewidth]{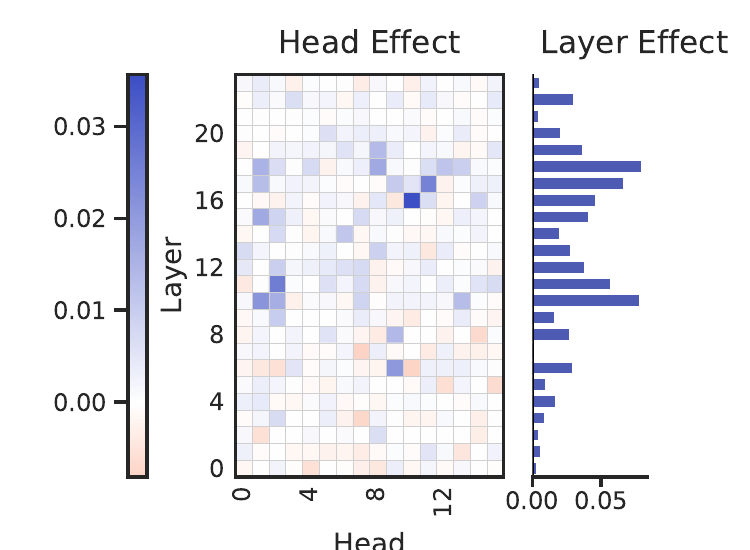}
        \caption{Scoring scheme 2}
        \vspace{1em}
    \end{subfigure}
    \begin{subfigure}[t]{.45\textwidth}
        \centering
        \includegraphics[width=1\linewidth]{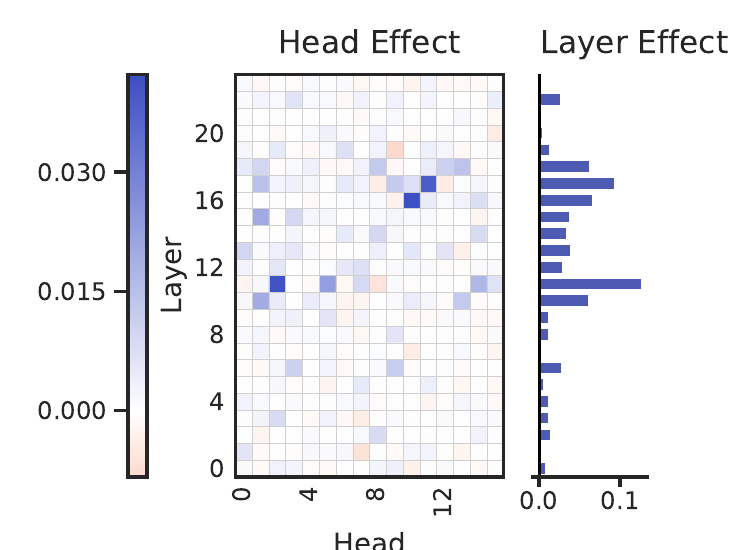}
        \caption{Scoring scheme 3}
        \vspace{1em}
    \end{subfigure}\hfill
    \begin{subfigure}[t]{.45\textwidth}
        \centering
        \includegraphics[width=1\linewidth]{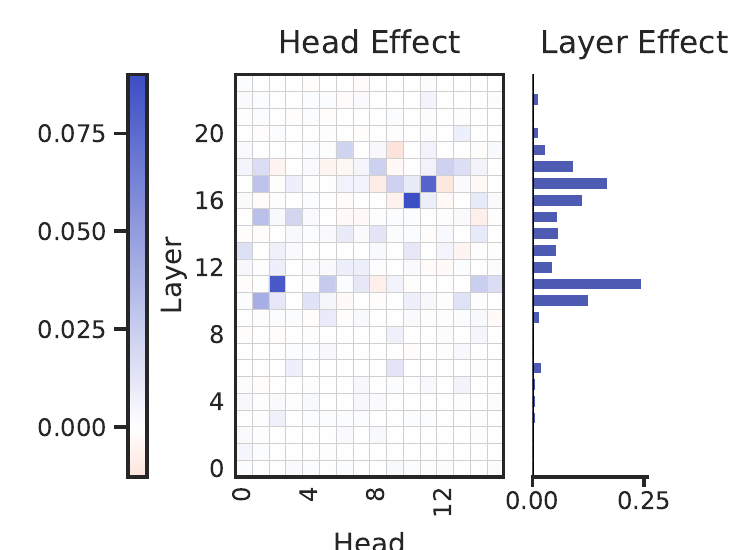}
        \caption{Scoring scheme 1'}
        \vspace{1em}
    \end{subfigure}
    \begin{subfigure}[t]{.45\textwidth}
        \centering
        \includegraphics[width=1\linewidth]{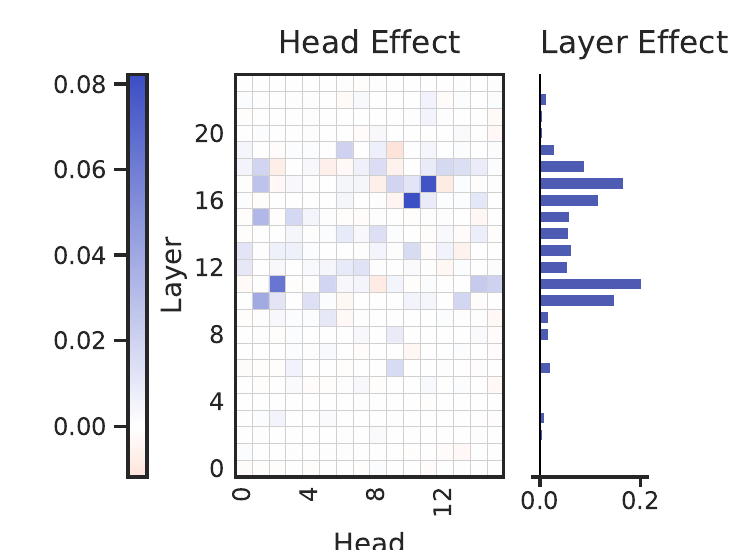}
        \caption{Scoring scheme 2'}
    \end{subfigure}\hfill
    \begin{subfigure}[t]{.45\textwidth}
        \centering
        \includegraphics[width=1\linewidth]{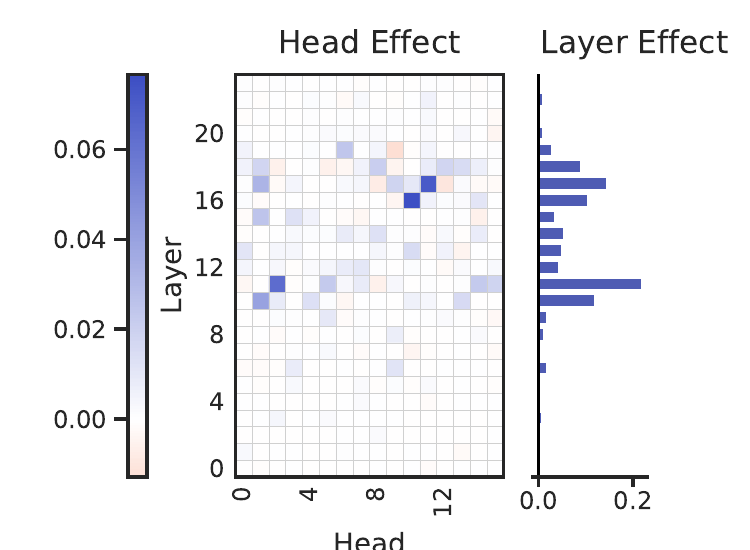}
        \caption{Scoring scheme 3'}
    \end{subfigure}
    \caption{Mean indirect effect on Winobias for heads (the heatmap) and layers (the bar chart) in RoBERTa-large over different scoring schemes.}
    \label{fig:robert-large-scoring}
\end{figure*}

Scoring a multi-token continuation, say $[x_3, x_4]$, given a prefix, say $[x_1, x_2]$, is straightforward with autoregressive models: we simply need to combine the individual token-level probabilities $p_\theta(x_3 \mid x_1, x_2)$ and $p_\theta(x_4 \mid x_1, x_2, x_3)$ (which we do by taking a geometric mean). The problem becomes less trivial when we are faced with masked LMs, as there is not a single obvious way to compute token-level probabilities anymore. In this work, we try out three different ways of doing so, which in our running example would correspond to defining token-level probabilities as:
\begin{itemize}
    \item[1.] $p_\theta(x_3 \mid x_1, x_2, \texttt{mask})$ and $p_\theta(x_4 \mid x_1, x_2, x_3, \texttt{mask})$;
    \item[2.] $p_\theta(x_3 \mid x_1, x_2, \texttt{mask}, x_4)$ and $p_\theta(x_4\mid x_1, x_2, x_3, \texttt{mask})$;
    \item[3.] $p_\theta(x_3 \mid x_1, x_2, \texttt{mask},\texttt{mask})$ and $p_\theta(x_4 \mid x_1, x_2, x_3, \texttt{mask})$.
\end{itemize}
We propose scheme 1 because it seems to give the closest formulation to the autoregressive setting, while schemes 2 and 3 are derived from existing literature \cite{pmlr-v101-shin19a,wang-cho-2019-bert,salazar-etal-2020-masked}. We also have a choice of whether or not to include the special \texttt{cls} and \texttt{sep} tokens (used during the pre-training of the models; e.g., \texttt{[CLS]}, and \texttt{[SEP]} in the case of BERT) when feeding examples to the masked LMs. Consequently, we define three additional scoring schemes which are exact copies of the original ones but with the special tokens included:
\begin{itemize}
    \item[1'.] $p_\theta(x_3 \mid \texttt{cls}, x_1, x_2, \texttt{mask}, \texttt{sep})$ and $p_\theta(x_4 \mid \texttt{cls}, x_1, x_2, x_3, \texttt{mask}, \texttt{sep})$;
    \item[2'.] $p_\theta(x_3 \mid \texttt{cls}, x_1, x_2, \texttt{mask}, x_4, \texttt{sep})$ and $p_\theta(x_4 \mid \texttt{cls}, x_1, x_2, x_3, \texttt{mask},\texttt{sep})$;
    \item[3'.] $p_\theta(x_3 \mid \texttt{cls}, x_1, x_2, \texttt{mask},\texttt{mask}, \texttt{sep})$ and $p_\theta(x_4 \mid \texttt{cls}, x_1, x_2, x_3, \texttt{mask}, \texttt{sep})$.
\end{itemize}
As mentioned in Section \ref{sec:other-models}, we find considerable variation in the results with different masked LMs and scoring schemes. Figures \ref{fig:distilbert-scoring}, \ref{fig:bert-base-scoring}, \ref{fig:bert-large-scoring}, \ref{fig:robert-base-scoring}, and \ref{fig:robert-large-scoring} show the indirect effects for each head and layer in DistilBERT, BERT-base-uncased, BERT-large-uncased, RoBERTa-base, and RoBERTa-large, respectively, using all six of the scoring schemes, on the filtered Winobias Dev dataset. We do observe the general trend of total effects being larger for larger variants of the same model, however. This is illustrated in tables \ref{tab:mlm-te-winobias} and \ref{tab:mlm-te-winogender}, which list the total effects for different models and scoring schemes on the filtered Winobias Dev, Winogender Bergsma, and Professions datasets.

\clearpage

\begin{table*}[h]
    \centering
    \begin{tabular}{l r r r r r r r r r r}
        \toprule 
        & \multicolumn{6}{c}{Winobias Dev Filtered} \\
        \cmidrule(lr){2-7}
        Model/Scoring scheme & 1 & 2 & 3 & 1' & 2' & 3' \\
        \midrule 
        DistilBERT & 0.224 & 0.191 & 0.158 & 0.215 & 0.189 & 0.161 \\
        BERT-base-uncased & 0.213 & 0.179 & 0.178 & 0.262 & 0.240 & 0.245 \\
        BERT-large-uncased & 0.935 & 0.744 & 0.950 & 0.430 & 0.303 & 0.304 \\
        RoBERTa-base & 0.812 & 0.478 & 0.469 & 0.399 & 0.457 & 0.436 \\
        RoBERTa-large & 0.402 & 0.610 & 0.614 & 0.994 & 1.033 & 0.859 \\
        \bottomrule 
    \end{tabular}
    \caption{Total effects on Winobias Dev filtered for all masked LMs using different scoring schemes.}
    \vspace{-.5em}
    \label{tab:mlm-te-winobias} 
\end{table*}

\begin{table*}[h]
    \centering
    \begin{tabular}{l r r r r r r r}
        \toprule
        Model & WB & WG & Prof. \\
        \midrule
        Transformer-XL & 0.356 & 0.353 & 36.168 \\
        XLNet-base-cased & 0.327 & 0.201 & 0.729 \\
        XLNet-large-cased & 1.140 & 0.405 & \\
        DistilBERT & 0.189 & 0.082 & 16.278 \\
        BERT-base-uncased & 0.240 & 0.076 & 3.675 \\
        BERT-large-uncased & 0.303 & 0.165 & 1.775 \\
        RoBERTa-base & 0.457 & 0.191 & 29.625 \\
        RoBERTa-large & 1.033 & 0.230 & 1.789 \\
        \bottomrule 
    \end{tabular}
    \caption{Total effects (TE) of gender bias in models other than GPT2 evaluated on Winobias (WB),  Winogender (WG), and the professions dataset (Prof.). Results for masked LMs on WB and WG are with scoring scheme 2'.}
    \vspace{-.5em}
    \label{tab:mlm-te-winogender} 
\end{table*}

\clearpage

\vskip 0.2in
\bibliography{acl2020,anthology}
\bibliographystyle{theapa}

\end{document}

%% file: math_commands.tex
\usepackage{amsmath,amsfonts,bm}

\def\eqref#1{equation~\ref{#1}}

\def\1{\bm{1}}

\def\vh{{\bm{h}}}

\def\vu{{\bm{u}}}

\def\vx{{\bm{x}}}
\def\vy{{\bm{y}}}
\def\vz{{\bm{z}}}

\DeclareMathAlphabet{\mathsfit}{\encodingdefault}{\sfdefault}{m}{sl}
\SetMathAlphabet{\mathsfit}{bold}{\encodingdefault}{\sfdefault}{bx}{n}

\newcommand{\E}{\mathbb{E}}

\newcommand{\R}{\mathbb{R}}